\documentclass[10pt,twocolumn,letterpaper]{article}

\usepackage[accsupp]{axessibility} %
\usepackage{cvpr}              %

\usepackage[dvipsnames]{xcolor}

\definecolor{cvprblue}{rgb}{0.21,0.49,0.74}
\usepackage[pagebackref,breaklinks,colorlinks,citecolor=cvprblue]{hyperref}
\usepackage{url}
\usepackage{booktabs}       %
\usepackage{amsfonts}       %
\usepackage{nicefrac}       %
\usepackage{microtype}      %
\usepackage{xcolor}         %
\usepackage{graphicx} %
\usepackage{multirow}
\usepackage{amsmath}
\usepackage{float}
\usepackage{enumitem}
\usepackage{placeins} %
\usepackage{subcaption} %
\usepackage{algorithm}
\usepackage{algpseudocode}
\usepackage{pifont}
\usepackage{colortbl}
\usepackage{mdframed}
\usepackage{indentfirst}
\usepackage{nicematrix}
\definecolor{mygray}{gray}{0.85}
\definecolor{softgray}{gray}{0.96}

\usepackage{tabularx}
\usepackage{graphicx}
\usepackage{array}

\usepackage{pgfplots}
\usepgfplotslibrary{groupplots}

\usepackage{color}
\usepackage{ifthen}

\definecolor{zhiqiu_color}{rgb}{0,0.5,1}
\definecolor{deva_color}{rgb}{0.2,.64,0}
\definecolor{deepak_color}{rgb}{1,0,1}
\definecolor{shihong_color}{rgb}{1,0,1}
\definecolor{sam_color}{rgb}{1,0,1}

\newif\ifsubmit
\submitfalse
\ifsubmit
    \newcommand{\zhiqiu}[1]{}
    \newcommand{\deva}[1]{}
    \newcommand{\deepak}[1]{}
    \newcommand{\shihong}[1]{}
    \newcommand{\sam}[1]{}
\else
    \newcommand{\zhiqiu}[1]{\textsf{\textcolor{zhiqiu_color}{[{\bf Zhiqiu}: #1]}}}
    
    \newcommand{\deepak}[1]{\textsf{\textcolor{deepak_color}{[{\bf Deepak}: #1]}}}
    \newcommand{\shihong}[1]{\textsf{\textcolor{shihong_color}{[{\bf shihong}: #1]}}}
    \newcommand{\sam}[1]{\textsf{\textcolor{sam_color}{[{\bf sam}: #1]}}}
\fi

\usepackage{amsmath}
\usepackage{bbm}

\newcommand{\deva}[1]{{\leavevmode\color[rgb]{1,0,1}[Deva: #1]}}

\newcolumntype{P}[1]{>{\raggedright\arraybackslash}p{#1}}
\newcolumntype{M}[1]{>{\centering\raggedright\arraybackslash}m{#1}}
\newcolumntype{B}[1]{>{\raggedright\arraybackslash}b{#1}}

\def\paperID{8674} %
\def\confName{CVPR}
\def\confYear{2024}

\title{Language Models as Black-Box Optimizers for Vision-Language Models}

\author{Shihong Liu\thanks{Co-first authors. Published at CVPR 2024.}\quad Zhiqiu Lin$^*$\quad Samuel Yu$^*$\quad Ryan Lee \quad Tiffany Ling \\ Deepak Pathak\quad Deva Ramanan \\ \\
  Carnegie Mellon University \\
}

\begin{document}
\maketitle

\begin{figure*}[h]
  \centering
  \includegraphics[width=0.8\textwidth]{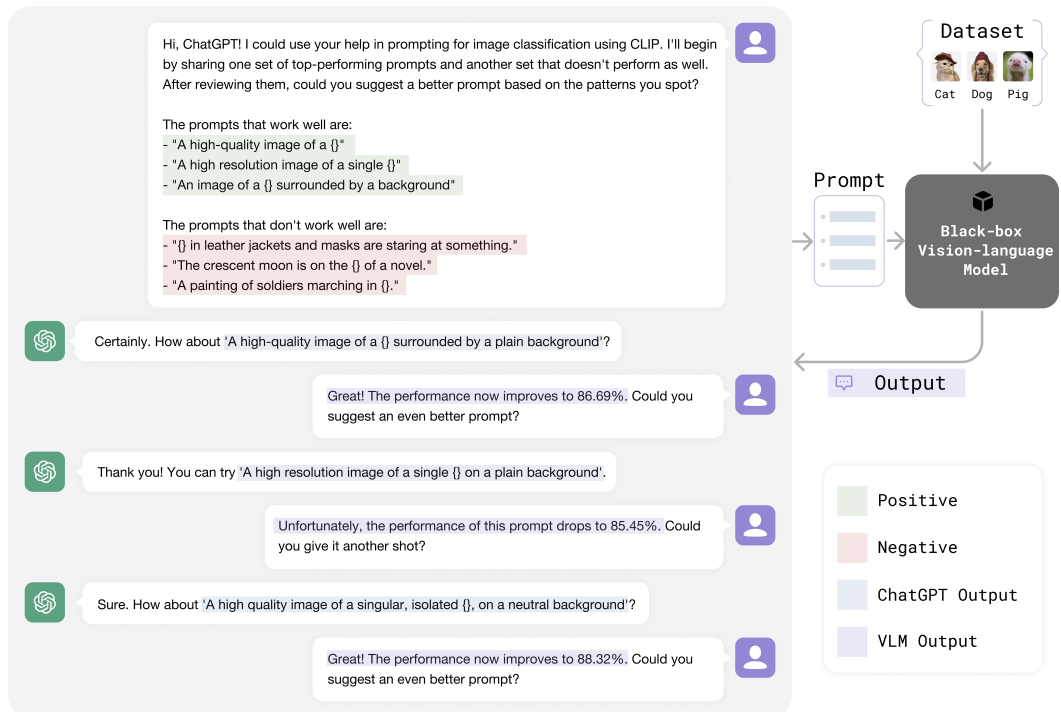}
  \caption{{\bf Prompting VLMs using chat-based LLMs.} Similar to how human prompt engineers iteratively test and refine prompts, we employ ChatGPT~\cite{gpt4, chatgpt} to continuously optimize prompts for vision-language models (VLMs). Our iterative approach assesses the performance of ChatGPT-generated prompts on a few-shot dataset (highlighted in \textcolor{blue}{blue}) and provides feedback (marked in \textcolor{violet}{violet}) to ChatGPT through simple conversations, as depicted in the illustrative figure. This straightforward method delivers state-of-the-art results for one-shot image classification across 11 datasets using CLIP, operated in a black-box manner without accessing model weights, feature embeddings, or output logits. We show that providing both positive (in \textcolor{green}{green}) and negative prompts (in \textcolor{red}{red}) enhances efficiency. Remarkably, our approach outperforms both white-box methods such as gradient-based continuous prompting (CoOp~\cite{coop}) and human-engineered prompts~\cite{clip} in this extremely low-shot scenario. This figure only shows a typical conversation using ChatGPT's web user interface. Our code implementation follows this pattern using the ChatGPT API. We detail and ablate the prompts in \autoref{sec:appendix_setup}.}
  \label{fig:teaser}
  \vspace{-1mm}
\end{figure*}

\begin{abstract}
Vision-language models (VLMs) pre-trained on web-scale datasets have demonstrated remarkable capabilities on downstream tasks when fine-tuned with minimal data. However, many VLMs rely on proprietary data and are not open-source, which restricts the use of white-box approaches for fine-tuning. As such, we aim to develop a black-box approach to optimize VLMs through {\bf natural language prompts}, thereby avoiding the need to access model parameters, feature embeddings, or even output logits. We propose employing chat-based LLMs to search for the best text prompt for VLMs. Specifically, we adopt an automatic ``hill-climbing'' procedure that converges to an effective prompt by evaluating the performance of current prompts and asking LLMs to refine them based on textual feedback, all within a conversational process without human-in-the-loop. In a challenging 1-shot image classification setup, our simple approach surpasses the white-box continuous prompting method (CoOp) by an average of $1.5\%$ across 11 datasets including ImageNet. Our approach also outperforms both human-engineered and LLM-generated prompts. We highlight the advantage of conversational feedback that incorporates both positive and negative prompts, suggesting that LLMs can utilize the implicit ``gradient'' direction in textual feedback for a more efficient search. In addition, we find that the text prompts generated through our strategy are not only more interpretable but also transfer well across different VLM architectures in a black-box manner. Lastly, we apply our framework to optimize the state-of-the-art black-box VLM (DALL-E 3) for text-to-image generation, prompt inversion, and personalization.
\end{abstract}

\section{Introduction}
\label{sec:intro}

Vision-language models~\cite{clip, flamingo, wang2022git, blip2} (VLMs) excel at a wide range of classic vision and multimodal~\cite{deng2009imagenet, coco, flickr30k, vqa, vqa2} tasks, surpassing the performance of their fully-supervised counterparts on downstream tasks even when fine-tuned with minimal data~\cite{lin2023multimodality, coop}. However, fine-tuning VLMs typically requires transparent {\em white-box} access to the model weights, such as gradient-based approaches that rely on backpropagation.

{\bf VLMs as black-box services.} Despite community efforts to collect web-scale public datasets~\cite{laion400m, laion5b} and to replicate proprietary VLMs~\cite{ilharco_gabriel_2021_5143773, openflamingo}, an increasing number of models~\cite{flamingo, yu2022coca, gpt4, wang2022git, driess2023palm, dalle3} are not releasing their weights due to privacy and legal concerns~\cite{li2020federated, madry2017towards}. Therefore, one cannot use popular {\em white-box} fine-tuning strategies (such as LoRA~\cite{hu2021lora} and Adapter~\cite{houlsby2019parameter}) that rely on model weights, feature embeddings, and output logits. Given that contemporary black-box VLMs~\cite{chatgpt,gpt4} like DALL-E~\cite{dalle3, ramesh2022dalle2} still offer a language-based user interface and may be accessed through APIs that facilitate input and output in {\em natural language}, this allows users to customize these models through optimizing textual prompts.

{\bf Manual prompting.} Manual prompt engineering has been proven successful in adapting black-box LLMs to language tasks~\cite{wei2022chain, kojima2022large}. Similarly, carefully crafted prompts can enhance the performance of VLMs. For instance, CLIP has demonstrated improved zero-shot recognition performance using specifically tailored prompts, such as \texttt{"a photo of a \{class\}"} for Internet photos and \texttt{"a satellite image of a \{class\}"} for satellite imagery. Despite its effectiveness, manual prompting can be a laborious process, inspiring efforts to explore automated prompt creation and thereby remove the need for human involvement. These strategies typically leverage an LLM as a knowledge base to create rich visual descriptors that augment the prompts for each class~\cite{menon2022visual, pratt2022does} in a zero-shot fashion.

{\bf Human-free prompting with conversational LLMs (our approach).} We show how to effectively leverage chat-based LLMs~\cite{gpt4} to emulate human-level prompt engineering {\em without} any human input. We first address an illustrative low-shot image classification task, aiming to find the best class-agnostic prompt (or ``template'') for image classification with CLIP. We start with a random set of prompts and evaluate the one-shot training accuracy of each. Then, akin to human prompt engineering, our method repeatedly presents ChatGPT with the best and worst prompts, asking it to review the results and suggest an improvement (see \autoref{fig:teaser}). %

{\bf Learning with implicit ``gradients'' provided through conversational feedback.} One of our key findings is that LLMs can learn the difference between effective and ineffective prompts, and can use this implicit ``gradient'' direction provided through language to perform more efficient searches. Compared to previous automatic prompting methods that only use LLMs as a knowledge base~\cite{menon2022visual, pratt2022does} or paraphrasing tool~\cite{ape}, we show a novel use of LLMs as an {\em optimizer} that can utilize the patterns hidden in textual feedback. In our experiments, we find that the inclusion of such feedback greatly improves the efficiency and accuracy of our method, sometimes surpassing existing white-box methods~\cite{coop, wortsman2021wiseft} on challenging one-shot scenarios.

{\bf Optimizing text-to-image generation with DALL-E 3.} We further demonstrate our optimization framework on a state-of-the-art black-box VLM, DALL-E~\cite{dalle3}, for two illustrative one-shot generative tasks: (1) Text-to-image (T2I) generation (see \autoref{fig:dalle3}), where we sample challenging text queries from Winoground~\cite{winoground} that involve reasoning over compositions of objects, attributes, and relations. Examples include ``{\tt an animal watches a person}'' and ``{\tt there is less milk than orange juice}'', which DALL-E 3 might initially fail to generate. (2) Prompt inversion (see \autoref{fig:dalle3_inversion}), which attempts to reverse-engineer the textual prompt to generate a specific image for later customization (personalization)~\cite{dreambooth} (see \autoref{tab:customization}). To achieve this, we leverage conversational feedback from a multimodal LLM (GPT4-V~\cite{gpt4}) to iteratively refine the prompts based on the current generated images. We present qualitative results in \autoref{tab:opt_dalle} and conduct a user study to demonstrate that our framework can be more efficient than manual prompting, even for graphical designers experienced with AI content-generation tools. 

{\bf Our contributions.} In this work, we introduce a novel prompting method for VLMs, utilizing an LLM as an {\em optimizer}. Our black-box approach can surprisingly compete with various white-box methods in a low-shot setting. Additionally, we extensively explore various strategies for conversing with ChatGPT, uncovering several key factors that significantly enhance the efficiency of this tool. We also show that our discovered natural language prompts are not only {\em interpretable} but also {\em transfer} better across CLIP architectures, eg., from RN50 to ViT/B-16, than continuous prompts discovered by previous white-box prompting method~\cite{coop}. Finally, we show practical applications of our framework on text-to-image generation using black-box DALL-E 3. We release our code for future research on prompt optimization and AI-driven content creation~\footnote{Project site: \url{llm-can-optimize-vlm.github.io}}.

\section{Related Works}
\label{sec:relate_dwork}

{\bf LLMs for multimodal tasks.} Cutting-edge LLMs like GPTs~\cite{chatgpt, gpt4} have been successfully applied to multimodal tasks, either through zero-shot composition with pre-trained multimodal models~\cite{li2022composing, zeng2022socratic} or by jointly fine-tuning with modality-specific encoders~\cite{blip2, flamingo} on large-scale multimodal datasets~\cite{laion5b}. LLMs are also utilized as neuro-symbolic reasoners~\cite{gupta2022visual, shen2023hugginggpt, lu2023chameleon, zheng2023can}, translating natural language instructions into modular programs (like Python code) that invoke APIs of multimodal models. In this work, we show the potential of LLMs as a {\em black-box optimizer} for multimodal foundation models with language interfaces, and more specifically vision-language models (VLMs).

{\bf Prompt optimization of foundation models. } 
Following the success of in-context learning~\cite{gpt3}, which appends user-generated natural language instruction and few-shot samples to text inputs, prompting~\cite{liu2023pre} has emerged as the preferred fine-tuning paradigm for LLMs due to its superior performance and parameter-efficiency. However, recent prompt optimization methods, including continuous prefix-tuning~\cite{li2021prefix, sun2022blackbox, xu2022gps, chai2022clip, sun2022bbtv2} and discrete token-searching~\cite{shin2020autoprompt, deng2022rlprompt, diao2022blackbox}, still operate in a white-box manner, requiring access to either the tokenizer or output logits. Moreover, black-box prompting methods, such as heuristic-based editing~\cite{prasad2022grips, mishra2021reframing}, are tailored towards language-only tasks and are thus not applicable in VLM settings.

{\bf LLMs for prompt optimization.} APE~\cite{ape} leverages an LLM to automatically write prompts using few-shot samples based on instruction induction~\cite{honovich2022instruction} and paraphrasing~\cite{russell2010artificial, mitchell1993will}. However, it is only designed to address language tasks, while we focus on multimodal tasks using black-box VLMs. LLMs have also proven to be an effective external knowledge base~\cite{shen2022k, menon2022visual, pratt2022does} for generating prompts in a zero-shot setting for multimodal models. For example, DCLIP~\cite{menon2022visual} uses GPT3 to come up with rich visual descriptions to improve zero-shot classification with CLIP~\cite{clip}. We extend this line of work to show that LLMs can {\em iteratively} optimize prompts for VLMs in a black-box fashion given few-shot samples. We further illustrate that prompt optimization with LLMs can be made more efficient by leveraging {\em conversational} feedback, such as providing ChatGPT with explicit language feedback on how well the most recent prompt performs. Our findings align with the perspective~\cite{dai2022can} of LLMs as meta-optimizers that can implicitly perform gradient search through in-context learning.

{\bf Few-shot adaptation of VLMs.} Prompting has also been successfully adopted in VLMs~\cite{gan2022vision}, as demonstrated by methods like CoOp~\cite{coop} that fine-tune an ensemble of continuous prefix tokens using cross-entropy loss. \cite{lin2023multimodality} achieves state-of-the-art few-shot performance with a cross-modal (image and text) cross-entropy loss. However, these methods all require access to model parameters for gradient backpropagation. We also note that while some concurrent works, such as BlackVIP~\cite{oh2023blackvip} and LFA~\cite{ouali2023black}, claim to operate in a ``black-box'' setting, they still require access to {\em privileged} information including output logits and embeddings. 
In this work, we introduce a truly black-box and gradient-free approach that yields competitive results to white-box approaches in extremely low-shot scenarios.

\section{Prompting VLMs Using Chat-Based LLMs}
\label{sec:conversational_prompting}

We now present our approach for prompting VLMs using chat-based LLMs as optimizers. 

{\bf Preliminaries.} Motivated by recent proprietary VLMs~\cite{gpt4, dalle3}, we adopt a stricter yet practical black-box setting compared to prior works~\cite{oh2023blackvip, ouali2023black}, requiring {\em minimal} knowledge about the model's inner workings. This is crucial since releasing output logits or embeddings can potentially facilitate unauthorized knowledge extraction through distillation methods~\cite{hinton2015distilling}. Our objective is to enhance the performance of a VLM equipped with a language interface capable of processing a textual prompt $p \in T$. We assume that the targeted task is accompanied by a training dataset denoted as $D_{train} \subset D$, and its performance can be evaluated with respect to the prompt, represented as a function $F : D \times T \to \mathbb{R}$. For example, in a classification task, $D_{train} = \{x, y\}_n$ where $x$ is an image and $y$ is its class label. The black-box VLM takes the image as input and returns a predicted label. We measure the performance of the textual prompt by calculating the average classification accuracy as $F(D_{train}, p)$. Our goal in prompt engineering is to search for the optimal prompt $p^{\ast}$ without accessing or modifying the black-box VLM.

{\bf Background: human prompt engineering.} Our method draws inspiration from the typical workflow of human prompt engineers. Prompt engineering is often an iterative process that involves: (a) creating an initial prompt $U = \{p_1\}$ based on the understanding of a task, (b) evaluating the performance of prompts in $U$, (c) refining prompts based on the outcomes, (d) repeating the last two steps until convergence, and (e) returning the prompt $p^{\ast}$ with the highest $F(D_{train}, p^\ast)$. This hands-on approach helps optimize the model's performance, but it can be tedious and labor-intensive. Algorithm~\autoref{alg:method_general} formally illustrates this process.

{\bf Example: prompting for image classification with CLIP~\cite{clip}.} CLIP is one of the most popular VLM that takes a set of class-specific prompts when performing ``zero-shot'' image classification. \cite{clip} details the laborious prompting procedure over the course of a year. Interestingly, they find that a default class-agnostic prompt (or so-called ``template''), ``\texttt{a photo of a \{class\}}'' can provide a decent boost in accuracy for most datasets compared to using vanilla class labels. In this scenario, the evaluation function $F$ is the classification accuracy on the test set, and the prompt $p = \{\text{``}\texttt{a photo of a \{class\}}\text{''} | c \in C\}$, where $C$ is the set of class names for a given dataset.

\begin{algorithm}[t]
\caption{We formalize {\em human} prompt engineering with the following algorithm, which motivates our LLM-based algorithm \eqref{alg:method}.}
\begin{algorithmic}[1]
\Require $D_{\text{train}} = \{{x, y}\}_n$: training samples, $F: D \times T \to \mathbb{R}$: evaluation function
\State Create initial prompts: $\mathcal{U} \leftarrow \{p_1\}$  %
\State Evaluate prompts on training set: $S \leftarrow \{F(D_{\text{train}}, p_1)\}$
\While {not converged}
    \State Generate a new prompt $p'$ based on $S$
    \State Evaluate the new prompt: $s' = F(D_{\text{train}}, p')$
    \State $\mathcal{U} \leftarrow \mathcal{U} \cup \{p'\}$
    \State $S \leftarrow S \cup \{s'\}$
\EndWhile
\State \Return optimal prompt $p^* \leftarrow \arg\max_{p\in \mathcal{U}} F(D_{\text{train}}, p)$
\end{algorithmic}
\label{alg:method_general}
\end{algorithm}

\begin{algorithm}[t]
\caption{LLM-based prompt engineering on the illustrative classification task. Our algorithm requires a chat-based LLM and a (black-box) evaluation function, such as accuracy. We highlight mechanisms for ``exploration'' (\textcolor{blue}{restart} and \textcolor{blue}{reset}) in \textcolor{blue}{blue} and ``exploitation'' (\textcolor{red}{iter}) in \textcolor{red}{red}. We mark the key component of ``\textcolor{violet}{conversational feedback}'' of our approach in \textcolor{violet}{violet}. The actual prompts are attached in \autoref{sec:appendix_setup}. }
\begin{algorithmic}[1]
\Require $D_{\text{train}} = \{{x, y}\}_n$: training samples, $F: D \times T \to \mathbb{R}$: evaluation function.
\Require $n_{\text{restart}}$: number of initial sampled prompt sets, $n_{\text{reset}}$: number of resets for a prompt set, $n_{\text{iter}}$: number of hill-climbing iterations, $m$: size of one initial prompt set, $k$: number of prompts send to ChatGPT.

\State $p^* \leftarrow \emptyset$ 

\For{\textcolor{blue}{{1::$n_{\text{restart}}$}}}
    \State Sample a new prompt set, $\mathcal{U}_{\text{init}} \leftarrow \{p_1, ..., p_m\}$  %
    \For{\textcolor{blue}{{1::$n_{\text{reset}}$}}}
        \State Reset to initial prompt set: $\mathcal{U} \leftarrow \mathcal{U}_{\text{init}}$

        \For{\textcolor{red}{1::$n_{\text{iter}}$}}

            \State Sort $\mathcal{U}$ by score outcomes $\{F(D_{\text{train}}, p)\}_{p \in U}$
            \State $\mathcal{U}_{\text{top}} \leftarrow$ top-k prompts in $\mathcal{U}$
            \State $\mathcal{U}_{\text{bot}} \leftarrow$ bottom-k prompts in $\mathcal{U}$
            \State \textcolor{violet}{Get a new prompt $p_{\text{new}} \leftarrow \text{LLM}(\mathcal{U}_{\text{top}}, \mathcal{U}_{\text{bot}})$}
            
            \State $\mathcal{U} \leftarrow \mathcal{U} \cup \{p_{\text{new}}\}$
        \EndFor
        \State $p^* \leftarrow \arg\max_{p\in \mathcal{U} \cup \{p^*\}} F(D_{\text{train}}, p)$
    \EndFor
\EndFor
\State \Return prompt with highest score $p^*$
\end{algorithmic}
\label{alg:method}
\end{algorithm}

{\bf Prompting with chat-based LLMs (our approach).} Given the strong in-context reasoning capabilities of LLMs, we envision them as a {\em black-box optimizers} that can improve prompts based on their performance outcomes, 
akin to how human prompt engineers iteratively refine prompts. Specifically, we maintain a pool of prompts $U$ and their corresponding performance outcomes $S$. In each iteration, we provide the LLM with both {\em positive} and {\em negative} prompts, such as the highest and lowest-performing candidates. Such textual feedback through in-context prompts offers LLMs an implied "gradient" direction~\cite{dai2022can}, making optimization more efficient than taking random local steps. We facilitate this feedback mechanism through {\em conversations} with state-of-the-art chat-based LLMs like ChatGPT~\cite{chatgpt} as illustrated in \autoref{fig:teaser}. We note that such a multi-turn conversation is not the only way of conversing with ChatGPT, and ablate different in-context feedback mechanisms in \autoref{sec:appendix_setup}.

\section{Illustrative Few-Shot Classification Task}
\label{sec:image_classification}

We illustrate our approach using a few-shot image classification task. Specifically, a prompt $p \in T$ consists of a set of class-specific prompts -- that is, one textual description per class. The evaluation function $F$ takes the prompt $p$, along with an image dataset $D_{train}$, and returns the accuracy using the black-box VLM. To prevent overfitting and simplify our search space, we restrict our search to finding a single class-agnostic template, e.g., {\tt a photo of a \{\}}, filling in the blank with label names provided with the dataset.

\begin{figure}[h]
  \centering
    \begin{subfigure}{0.5\columnwidth}
    \includegraphics[width=\columnwidth]{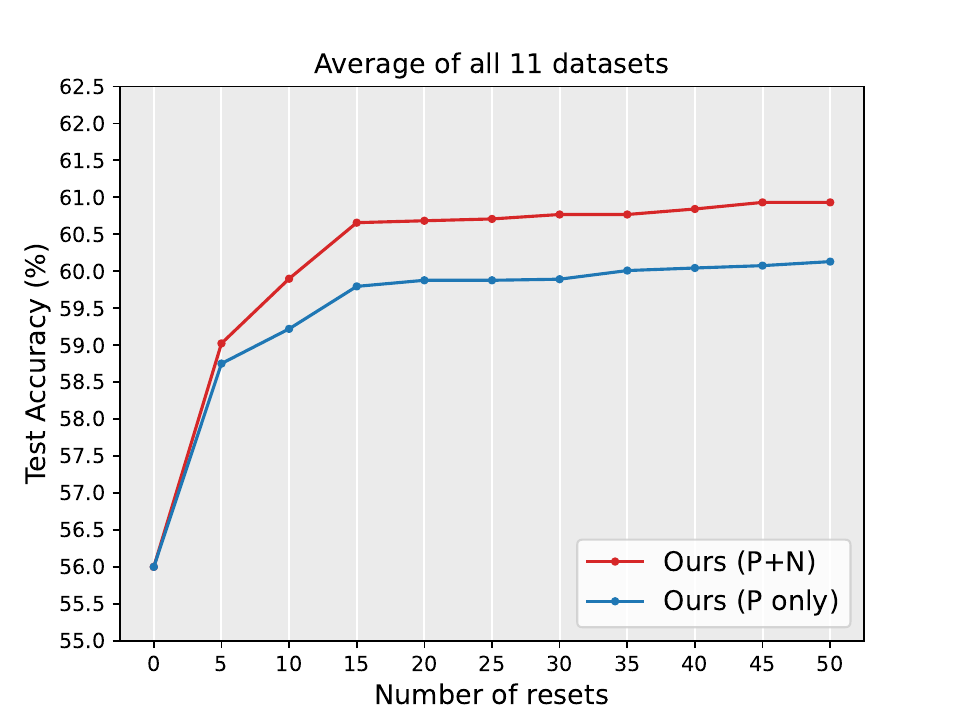}
  \end{subfigure}\hfill
  \begin{subfigure}{0.5\columnwidth}
    \includegraphics[width=\columnwidth]{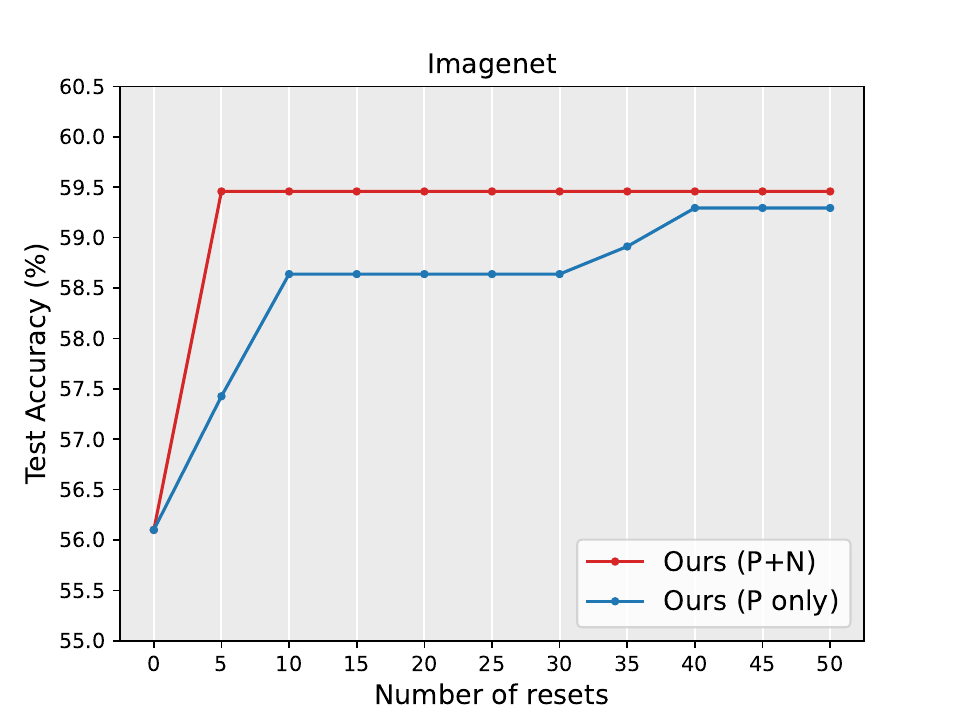}
  \end{subfigure}
  \caption{\textbf{Conversational feedback incorporating both positive and negative prompts leads to improved efficiency.} We fix the number of restarts to 20 and iterations to 10, and ablate different numbers of resets on all 11 datasets (left) and ImageNet (right). Notably, our approach using ``P+N'' (both top-15 and bottom-15 prompts) can optimize faster within a much fewer number of resets than using ``P-Only'' (top-30 prompts), resulting in the highest overall performance.}
  \label{fig:pncomparison}
\end{figure}

{\bf Outline of our approach (Alg.~\autoref{alg:method}).} 
To start, 
we sample entirely random initial prompts from a text corpus such as LAION-COCO~\cite{laion400m} captions. %
Our approach follows the classical {\em stochastic hill-climbing framework with random-restart}~\cite{russell2010artificial}, which prevents ChatGPT from being trapped in local optima by balancing ``exploration'' and ``exploitation''. Our {\bf restart} mechanism is implemented by sampling $n_{\text{restart}}$ initial prompt sets to encourage exploration. Because ChatGPT performs stochastic top-k sampling for text generation (as we adopt the default temperature of 1.0), we also implement a {\bf reset} mechanism to foster additional exploration by retrying a given prompt set $n_{\text{reset}}$ times. For exploitation, we converse with ChatGPT for $n_{\text{iter}}$ iterations. We find that it is critical to balance exploration and exploitation for optimal performance, and thoroughly examine this trade-off in \autoref{sec:appendix_experiments}. Lastly, we present ChatGPT both the top and bottom-performing prompts, denoted as $(\mathcal{U}_{top}, \mathcal{U}_{bot})$. We show that this simple adjustment can improve the efficiency of our approach in \autoref{fig:pncomparison}. 

{\bf Experimental setup.} We apply our approach to the few-shot image classification benchmark introduced in CoOp~\cite{coop}, which is the most commonly studied setup for fine-tuning VLMs. This benchmark involves a collection of 11 datasets covering diverse image domains including ImageNet~\cite{deng2009imagenet} and more niche datasets such as FGVC-Aircraft~\cite{maji13aircraft}. For each dataset, we adhere to the same three-fold k-shot train sets in~\cite{lin2023multimodality}, reporting the average accuracy across all folds. Importantly, our method only utilizes the train set to compute the score and does not require the few-shot validation set. We use CLIP following prior work~\cite{lin2023multimodality, coop} to emulate a black-box VLM, and we employ ChatGPT (GPT3.5) as the chat-based LLM.

{\bf Implementation details.} To start, we sample entirely random 1M initial prompts from a text corpus (LAION-COCO~\cite{laion400m} captions). For each caption, we extract all the noun phrases using spaCy part-of-speech tagging~\cite{honnibal2017spacy}. Subsequently, we replace one noun phrase in the caption with \texttt{``\{\}''} (a placeholder where the class name will be inserted) to create a template. Given that each caption contains an average of 2 noun phrases, our initial prompt pool consists of approximately 2M templates.  We run our algorithm with $n_{\text{restart}}=20$ restarts, $n_{\text{resets}}=50$ resets, and $n_{\text{iter}}=10$ iterations. We opt to sample $m=100$ prompts per restart and present the top and bottom $k=15$ prompts to ChatGPT. We ablate different sets of hyperparameters and explain how we balance the tradeoff between exploration and exploitation in \autoref{sec:appendix_experiments}. We adopt \texttt{gpt-3.5-turbo-0301} model for ChatGPT using OpenAI's official API and keep the default sampling temperature of $1.0$. We also ablate \texttt{gpt-4} in \autoref{tab:gpt4} and find it achieves similar performance. The exact prompts used to converse with ChatGPT are documented in~\autoref{sec:appendix_setup}. For a fair comparison, we use CLIP-RN50 for our experiments following prior work~\cite{lin2023multimodality, coop}. We will open-source our code
and release the initial prompt pool (LAIONCOCO-1M) to the public.

{\bf Oracle white-box baselines.} Our black-box setup substantially differs from, and is more constrained than, the scenarios considered in previous white-box baselines. Specifically, we do {\bf not} expose the pre-trained weights, model architectures, feature embeddings, or even output logits of VLMs. These constraints render many established {\em gradient-based fine-tuning} baselines inapplicable. Among the {\em oracle} white-box approaches we later compare to, {\bf CoOp}~\cite{coop} performs continuous prompting and requires backpropagation across all layers. {\bf WiSE-FT}~\cite{wortsman2021wiseft} ensembles fine-tuned weights with the original CLIP weights. {\bf Cross-Modal Adaptation}~\cite{lin2023multimodality} fine-tunes a linear classifier leveraging both image and text embeddings from CLIP. {\bf BlackVIP}~\cite{oh2023blackvip} and {\bf LFA}~\cite{ouali2023black} are two most recent baselines that apply CLIP logits or embeddings for gradient back-propagation. Finally, while {\bf DCLIP~\cite{menon2022visual}} queries GPT3 for rich visual descriptors for each class and does not require gradient-based fine-tuning, it performs {\em prompt ensembling} using 4-6 class-specific prompts, which breaches our black-box assumption for accessing the output logits. 

{\bf Black-box methods.} We additionally benchmark our method against truly black-box solutions, including the vanilla class-agnostic templates ``\texttt{\{class\}}'' and ``\texttt{a photo of a \{class\}}''. Also, we compare our approach to the best {\bf Hand-Engineered} templates released by OpenAI, searched using {\em test set} performance to represent the theoretical upper bound of human performance, eg., ``\texttt{a centered satellite photo of \{class\}.}'' for EuroSAT~\cite{helber2017eurosat}. 
Finally, we present two versions of conversational feedback of our approach: (a) using 30 positive ({\bf P only}) or (b) using 15 positive and 15 negative prompts ({\bf P+N}) in each iteration. For a fair comparison, both of our approaches start with the same initial sampled prompts, referred to as {\bf LAIONCOCO-1M}. We also show the performance of the best initial sampled prompt searched using trainset performance.

\begin{table*}[h]
\centering
\renewcommand{\arraystretch}{1.3}
\scalebox{0.7}{
\begin{NiceTabular}{ccccccccccccc}
            \CodeBefore
              \rectanglecolor{softgray}{3-1}{9-13}
            \Body
\toprule[1.5pt]
\multirow{2}{*}[-1mm]{\textbf{Method}} & \multicolumn{11}{c}{\textbf{Dataset}} & \multirow{2}{*}[-1mm]{\bf Avg}
\\ 
\cmidrule(l){2-12}
          &  Caltech & ImageNet & Aircraft & Food & Pets  & Cars  & SUN & UCF & DTD & EuroSAT & Flowers &  \\ \hline
\Block{1-13}{\textbf{Oracle white-box approaches}} \\
\textcolor{gray}{Cross-Modal}~\cite{lin2023multimodality}  & \textcolor{gray}{89.1} & \textcolor{gray}{61.6} & \textcolor{gray}{20.6} & \textcolor{gray}{77.1} & \textcolor{gray}{85.7} & \textcolor{gray}{59.0}  & \textcolor{gray}{63.4}  & \textcolor{gray}{64.7} & \textcolor{gray}{49.9}  & \textcolor{gray}{61.8} & \textcolor{gray}{76.3} & \textcolor{gray}{64.7}\\

\textcolor{gray}{WiSE-FT}~\cite{wortsman2021wiseft}  & \textcolor{gray}{85.5} & \textcolor{gray}{58.3} & \textcolor{gray}{{18.6}} & \textcolor{gray}{71.9} & \textcolor{gray}{81.7} & \textcolor{gray}{{55.7}} & \textcolor{gray}{56.6} & \textcolor{gray}{59.4}  & \textcolor{gray}{44.2} & \textcolor{gray}{{52.3}} & \textcolor{gray}{65.8} & \textcolor{gray}{59.1} \\

\textcolor{gray}{CoOp}~\cite{coop}    & \textcolor{gray}{87.5} & \textcolor{gray}{57.2} & \textcolor{gray}{9.6} & \textcolor{gray}{74.3} & \textcolor{gray}{{85.9}} & \textcolor{gray}{55.6} & \textcolor{gray}{60.3} & \textcolor{gray}{{61.9}} & \textcolor{gray}{{44.4}} & \textcolor{gray}{{50.6}} & \textcolor{gray}{{68.1}} & \textcolor{gray}{59.6} \\ 

\textcolor{gray}{LFA}~\cite{ouali2023black}    & \textcolor{gray}{81.6} & \textcolor{gray}{52.4} & \textcolor{gray}{17.0} & \textcolor{gray}{63.1} &  \textcolor{gray}{75.3} &  \textcolor{gray}{41.4} &  \textcolor{gray}{58.4} & \textcolor{gray}{56.7} & \textcolor{gray}{38.4} &  \textcolor{gray}{60.7} &  \textcolor{gray}{74.9} &  \textcolor{gray}{56.4}  \\ 

\textcolor{gray}{BlackVIP}~\cite{oh2023blackvip}    & \textcolor{gray}{85.8} & \textcolor{gray}{58.8} & \textcolor{gray}{15.3} & \textcolor{gray}{76.7} &  \textcolor{gray}{85.2} &  \textcolor{gray}{56.4} &  \textcolor{gray}{57.0} &  \textcolor{gray}{58.8} &  \textcolor{gray}{40.1} &  \textcolor{gray}{30.0} &  \textcolor{gray}{61.1} &  \textcolor{gray}{56.8}  \\ 

\textcolor{gray}{DCLIP}~\cite{menon2022visual}  & \textcolor{gray}{-} & \textcolor{gray}{{59.6}} & \textcolor{gray}{-} & \textcolor{gray}{76.4} & \textcolor{gray}{83.8} & \textcolor{gray}{-} & \textcolor{gray}{-} & \textcolor{gray}{-} & \textcolor{gray}{41.7} & \textcolor{gray}{34.7} & \textcolor{gray}{-} & \textcolor{gray}{-} \\

\midrule
\Block{1-13}{\textbf{Manual prompting approaches}} \\
\texttt{\{\}}     & 78.5     & 55.3   & 15.5   & 74.0  & 78.9 & 52.2 & 53.4 & 55.5 & 41.4 & 32.1  & 57.3  & 54.0 \\ 

a photo of a \texttt{\{\}}  & 84.5     & 57.9   & 15.9   & 74.0  & 83.2 & 53.9 & 58.0 & 56.9 & 38.8 & 28.6  & 60.2 & 55.6 \\ 

{Hand-Engineered}~\cite{clip}  & {86.3} & {58.2} & {17.3} & {77.3} & \underline{85.8} & {55.6} & {58.5} & {\bf {61.5}} & {42.3} & {37.6} & {66.1} & {58.8} \\

\midrule

\Block{1-13}{\textbf{Our black-box approaches}} \\

{LAIONCOCO-1M}  &  81.4& 56.2& 17.4& 76.5& 79.6& 51.3& 54.9& 55.8& 43.1& 38.6& 61.3& 56.0 \\

{Ours (P only)}  & {\underline{89.0}}     & \underline{59.4}    & {\underline{17.9}}    & {\underline{77.8}}   & {85.7}  & {\underline{55.7}}  & {\underline{60.4}}  & {58.7}  & \underline{43.6} & \underline{46.7}   & \underline{66.6}  & {\underline{60.1}} \\ 

{Ours (P+N)}  & {{\bf 89.1}}  & {{\bf 59.6}}  & {\bf 18.1} & {{\bf 78.3}}  & {{\bf 88.1}}  & {{\bf 56.2}}  & {{\bf 61.0}}  & \underline{60.2}  & {{\bf 44.8}}  & {\bf 49.0}  & {\bf {67.2}} & {{\bf 61.1}} \\ 

\bottomrule[1.5pt]
\end{NiceTabular}
}
\caption{\textbf{Comparison of our method with other baselines on one-shot classification tasks.} We report the average accuracy of each method across three folds, optimized using 1-shot training sets. We \textbf{bold} the best black-box result for each dataset, and \underline{underline} the second best result. First, we note that our approach can effectively improve upon the initial prompts selected from LAIONCOCO-1M from $56\%$ to $61\%$. Our approach is also competitive against the best Human-Engineered prompts released by OpenAI~\cite{clip} searched using {\em test set} performance. Additionally, we show that using both positive and negative prompts improves the overall accuracy by $1\%$. For reference, we report {\em oracle} white-box approaches in \textcolor{gray}{gray}. Remarkably, we also surpass white-box solutions such as WiSE-FT~\cite{wortsman2021wiseft} and CoOp~\cite{coop} by $1.5\%$. These methods require either gradient-based fine-tuning (CoOp/WiSE-FT/Cross-Modal) or prompt ensembling using output logits (DCLIP). While our approach is less effective than the SOTA white-box method (Cross-Modal Adaptation), we stress that our black-box setup is significantly more challenging, because we restrict the optimization space to {\em natural language} and do {\em not} access the pre-trained weights, model architectures, feature embeddings, and output logits of VLMs.}

\label{tab:main}
\end{table*}

{\bf SOTA one-shot performance against existing methods on 11 datasets.} We report the test set performance of our method versus the aforementioned baselines in a challenging 1-shot classification scenario in \autoref{tab:main}. First, compared to the top-performing initial prompts selected from {\bf LAIONCOCO-1M} based on train set performance, our prompt optimization using ChatGPT notably improves the initial prompts by an average of $5\%$ ($56\%$ to $61\%$). Remarkably, our black-box approach surpasses the two white-box gradient-based fine-tuning techniques CoOp and WiSE-FT by at least $1.5\%$. Given that both CoOp and our method optimize a single class-agnostic template, we attribute this gap in performance to {\em reduced overfitting}. More specifically, we posit that our optimization space of natural language effectively acts as a regularizer in extremely low-shot tasks, standing as a more robust alternative to the continuous prompting approach of CoOp. Furthermore, our method benefits from textual feedback and shows improved performance by $1.0\%$ when using both positive and negative prompts. In \autoref{sec:appendix_experiments}, we show that our approach remains effective across different CLIP and ChatGPT variants.

{\bf Incorporating negative prompts leads to more efficient optimization.} In \autoref{fig:pncomparison}, we demonstrate that incorporating both positive and negative prompts fosters better optimization efficiency, achieving higher accuracy within a much fewer number of resets. Specifically, we hypothesize that LLMs can leverage the implicit ``gradient'' direction suggested in textual feedback to achieve faster convergence. For additional analysis, we ablate different ways of providing conversational feedback to ChatGPT in \autoref{sec:appendix_setup} and conclude that iteratively updating both positive and negative prompts is the key for efficient optimization.

\section{More Benefits of Natural Language Prompts}
\label{sec:additional_benefits}

In this section, we delve deeper into the advantages of utilizing natural language prompts compared to the continuous prompts~\cite{coop}. We highlight that the prompts derived through our method are {\em interpretable}; for instance, they often contain descriptions of the targeted image domain. Our prompts can also {\em transfer} across CLIP architectures in a {\em black-box manner}, such as from RN50 to ViT/B-16.

\begin{table*}[]
\centering
\renewcommand{\arraystretch}{1.2}
\scalebox{0.65}{
\begin{tabular}{c|c}
\toprule[1.5pt]
\textbf{Dataset} & \textbf{Example of Top Templates} \\ \hline
Caltech~\cite{li2020caltech101} & \texttt{An image of a \{\} with a blurred background that emphasizes the subject} \\ 
DTD~\cite{cimpoi14dtd} & \texttt{The essential elements of \{\} are amplified with visual simplicity} \\
EuroSAT~\cite{helber2017eurosat} & \texttt{A top-down view of \{\} arranged in a pattern \{\}} \\
Aircraft~\cite{maji13aircraft} & \texttt{A clear, high-quality image of a single \{\} with a white background} \\
Food~\cite{bossard14food} & \texttt{A \{\} featuring diverse cuisine and ingredients} \\
ImageNet~\cite{deng2009imagenet} & \texttt{An image of a \{\} with bright and natural lighting} \\
Flowers~\cite{Nilsback08flowers} & \texttt{A clear and vivid photograph of the \{\} in its natural setting} \\
Pets~\cite{parkhi12pets} & \texttt{A \{\} with distinct and recognizable characteristics} \\
Cars~\cite{Krause2013cars} & \texttt{A \{\} featuring a wide range of color options for easy selection} \\
SUN~\cite{xiao2016sun} & \texttt{A high-resolution photo of a \{\} with clear background and natural lighting} \\
UCF~\cite{soomro2012ucf101} & \texttt{A black and white photo of a \{\} in motion}\\ \bottomrule[1.5pt]
\end{tabular}}
\caption{\textbf{Example templates returned by our algorithm on each dataset.} Although we do not provide ChatGPT with any information regarding the targeted dataset, we observe that the resulting templates are remarkably similar to human-engineered templates, with many domain-specific details such as ``motion'' and ``cuisine'', and stylistic elements such as ``bright and natural lighting''.}
\label{tab:templates}
\end{table*}

{\bf Interpretable natural language prompts.} While CoOp~\cite{coop} concedes that continuous prompts can be difficult to interpret, our method -- without explicitly instructing ChatGPT to generate interpretation -- often yields interpretable results. \autoref{tab:templates} showcases the templates returned by our algorithm for each dataset, frequently including keywords that reflect the targeted image domain. For example, the template for Food101~\cite{bossard14food} mentions ``diverse cuisine and ingredients'', and the template for UCF101~\cite{soomro2012ucf101} (an action recognition dataset) mentions ``in motion''. Likewise, these templates identify general stylistic attributes of the datasets; they refer to ``bright and natural lighting'' for ImageNet~\cite{deng2009imagenet} and note images that ``emphasize the subject'' for Caltech101~\cite{li2020caltech101}. These prompts are particularly intriguing because we do not provide ChatGPT with any information about the downstream task, yet it manages to generate prompts containing domain-specific keywords that are similar to those engineered by human experts.

\begin{table}[]
\centering
\renewcommand{\arraystretch}{1.3}
\scalebox{0.76}{
\begin{tabular}{c|c|ccc}
\toprule[1.5pt]
Method & RN50 & \textrightarrow RN101 & \textrightarrow ViT-B/32 & \textrightarrow ViT-B/16 \\ 
\midrule
\texttt{a photo of a \{\}} & 57.9 & 60.6 & 61.9 & 66.6 \\
CoOp & {\bf 63.0} & 20.6 & 31.7 & 39.5 \\
Ours & 59.9 & {\bf 60.7} & {\bf 62.2} & {\bf 67.0} \\
\bottomrule[1.5pt]
\end{tabular}
}
\vspace{-2mm}
\caption{\textbf{Black-box prompt transfer from ResNet-50 to other CLIP architectures.} We evaluate both our natural language prompts and CoOp's continuous prompts on 16-shot ImageNet, which are trained using the RN50 CLIP backbone. As a reference point, we include the baseline prompt ``\texttt{a photo of a \{\}}'', and show that the prompts derived from our method using RN50 consistently surpass it after transferring to different backbones. In contrast, while CoOp achieves better 16-shot ImageNet performance using RN50, its performance plummets during the transfer, e.g., from $63\%$ to a mere $21\%$ for RN101. }
\label{tab:transfer}
\end{table}

{\bf Black-box prompt transfer.} Our text prompts also maintain consistently high performance across different CLIP backbones. For comparison, since CoOp uses the same tokenizer for all CLIP architectures (including RN50, RN101, ViT/B-32, and ViT/B-16) and optimizes continuous prompts of the same shape (16 x 512), we assess the transferability of these learned continuous prompts from RN50 to other backbones using the official weights on 16-shot ImageNet. \autoref{tab:transfer} showcases the results of this experiment, where we also include the baseline prompt ``\texttt{a photo of a \{\}}'' for reference.  We observe a significant decline in accuracy when transferring CoOp's prompts (up to a $40\%$ decrease despite utilizing more powerful backbones), implying that continuous prompts tend to overfit to the specific CLIP model. In contrast, our natural language prompts maintain their performance and outperform the baseline across all backbones.

\section{Application: Text-to-Image Generation}
\label{sec:text_to_image}

\begin{figure*}[h]
  \centering
  \includegraphics[width=0.8\textwidth]{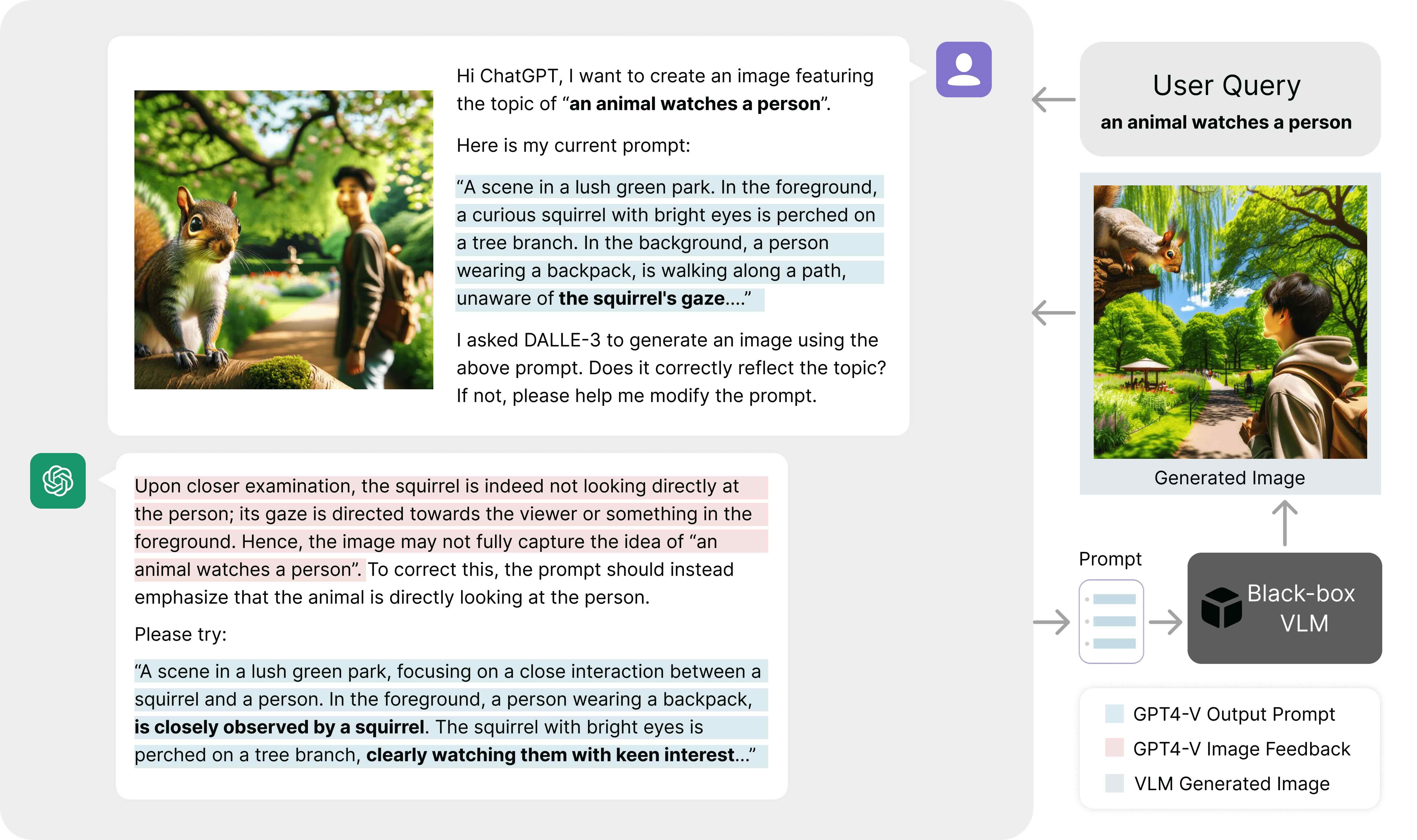}
  \caption{{\bf Improving text-to-image (T2I) generation using chat-based {\em multimodal} LLMs.} We apply our framework to optimize prompts for the state-of-the-art black-box generative VLM, DALL-E 3~\cite{dalle3}, using the multimodal GPT4-V~\cite{gpt4}. For complicated user queries that DALL-E 3 may initially fail to generate, we send the generated image (in \textcolor{violet}{violet}) along with the current prompt to GPT4-V to ask for feedback on improvements (in \textcolor{red}{red}) and then generate a new prompt (in \textcolor{blue}{blue}). We show that such a simple framework is surprisingly effective at correcting DALL-E 3 mistakes on some challenging Winoground~\cite{winoground} text queries that involve action, logical, and spatial reasoning. We conduct a human evaluation on the quality of generated images in \autoref{tab:human_study} and include the actual prompts in \autoref{sec:appendix_setup}. We open-source our code at \href{llm-can-optimize-vlm.github.io}{link} to facilitate future research on AI-driven content generation.}
  \label{fig:dalle3}
\end{figure*}

In this section, we present a direct application of our prompt optimization framework to generative tasks using a truly black-box text-to-image (T2I) VLM, DALL-E 3~\cite{dalle3}.

{\bf Optimizing T2I using a multimodal LLM.} DALL-E 3 can generate high-fidelity images following diverse user queries, but crafting effective prompts is tricky even for designers experienced with AI content generation tools~\cite{liu2023beyond}. Therefore, we are motivated to implement our LLM-based optimization framework to assist with creative visual design. Our framework is shown in \autoref{fig:dalle3} for the illustrative task of text-to-image generation. In this task, the user specifies a query (topic) in text, such as ``{\tt an animal watches a person}'', and the goal is to write a prompt that can generate an image reflecting this topic. We adopt a {\em multimodal} LLM GPT4-V~\cite{gpt4} ({\tt gpt-4-1106-preview}) to provide feedback on the generated image and optimize the prompt. We find that this framework is surprisingly effective due to GPT4-V's strong visual reasoning capabilities, which can often spot subtle errors in generated images and offer more accurate prompts.

\begin{figure*}[t]
  \centering
  \includegraphics[width=0.8\textwidth]{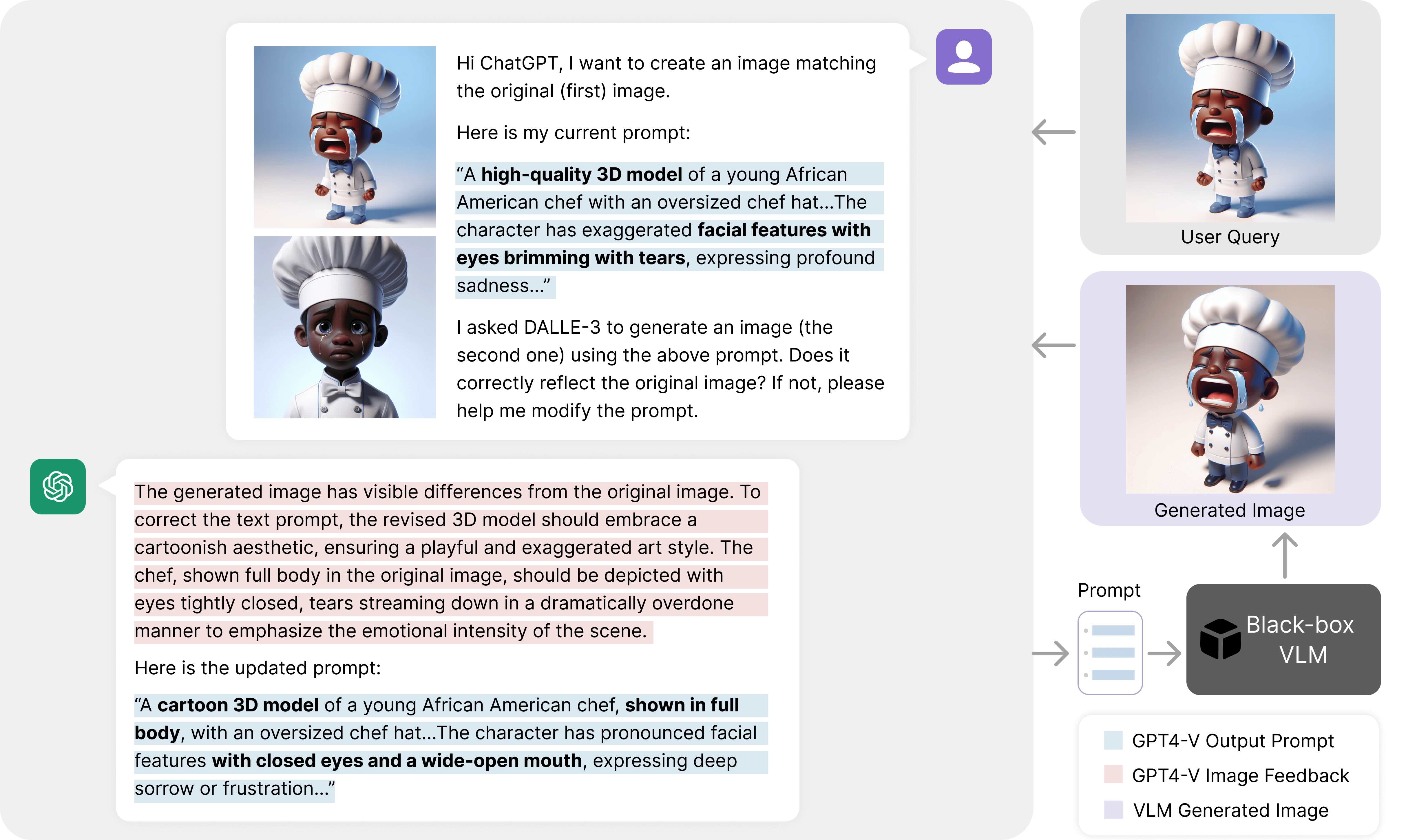}
  \caption{{\bf Prompt inversion using chat-based {\em multimodal} LLMs.} We apply our framework to reverse engineer the text prompt to generate the same user-queried image. We send the generated image (in \textcolor{violet}{violet}) along with the original image to GPT4-V to ask for feedback on improvements (in \textcolor{red}{red}) and then generate a new prompt (in \textcolor{blue}{blue}). The final reversed-engineered text prompt allows users to readily perform personalized (customized) generation (see \autoref{tab:customization}).}
  \label{fig:dalle3_inversion}
\end{figure*}

\begin{table}[]
\centering
\scalebox{0.75}{
\begin{NiceTabular}{M{0.30\linewidth} M{0.25\linewidth} M{0.33\linewidth} M{0.25\linewidth}}
\CodeBefore
    \Body
\toprule[1.5pt]
\textbf{User Query} & \textbf{Init. Image} & \textbf{LLM Feedback} & \textbf{Final Image} \\ \midrule
\Block{1-4}{{\bf Text-to-image generation}} \\
There is less milk than orange juice. & \includegraphics[width=18mm,height=18mm]{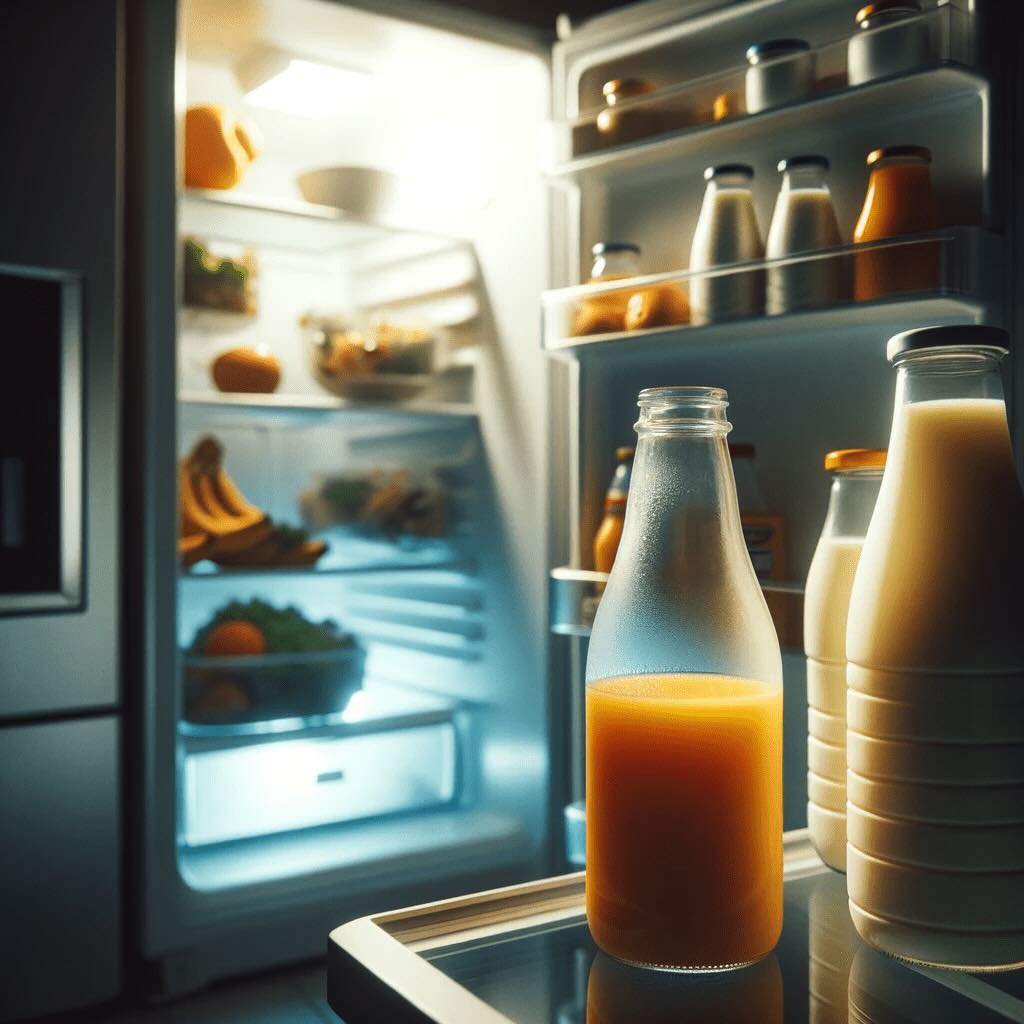} & Incorrect, the milk bottle appears full, more than orange juice... & \includegraphics[width=18mm,height=18mm]{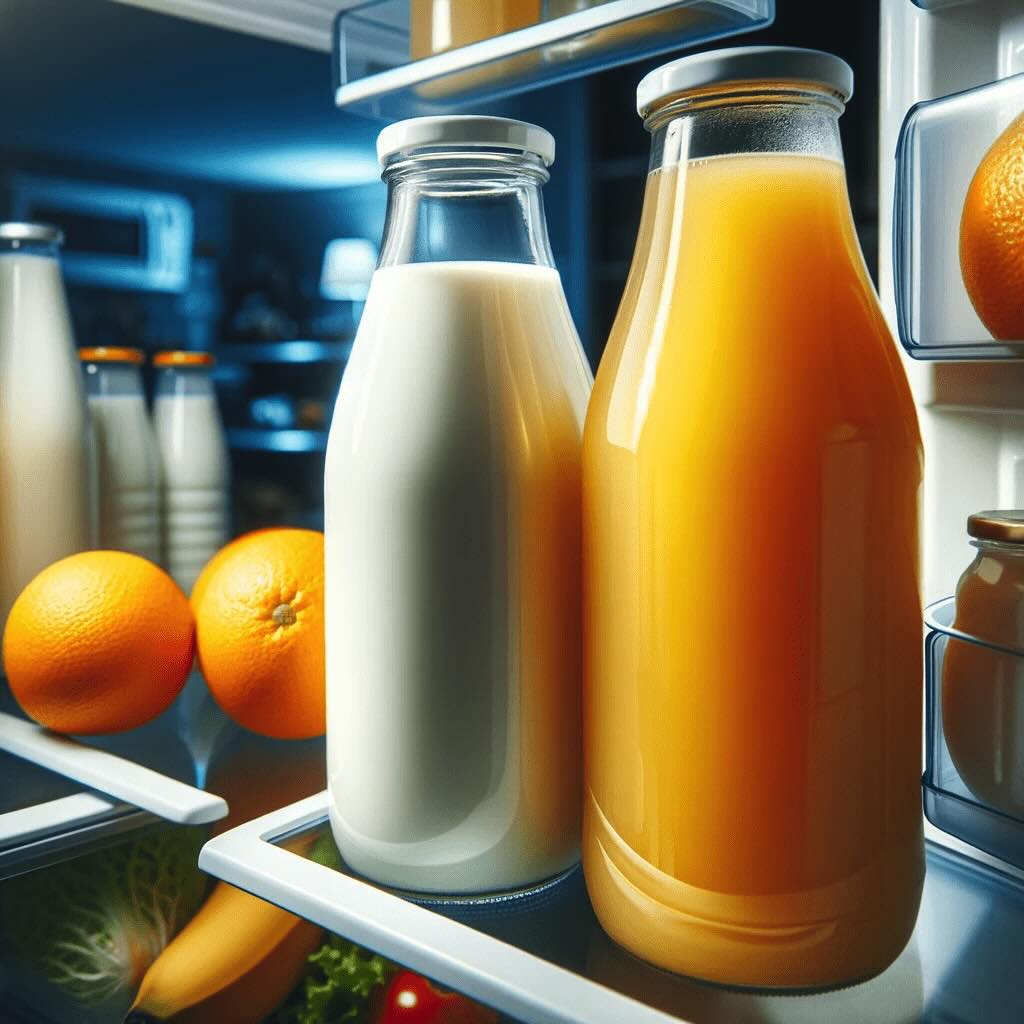} \\
A shorter person is covering the eyes of a taller person. & \includegraphics[width=18mm,height=18mm]{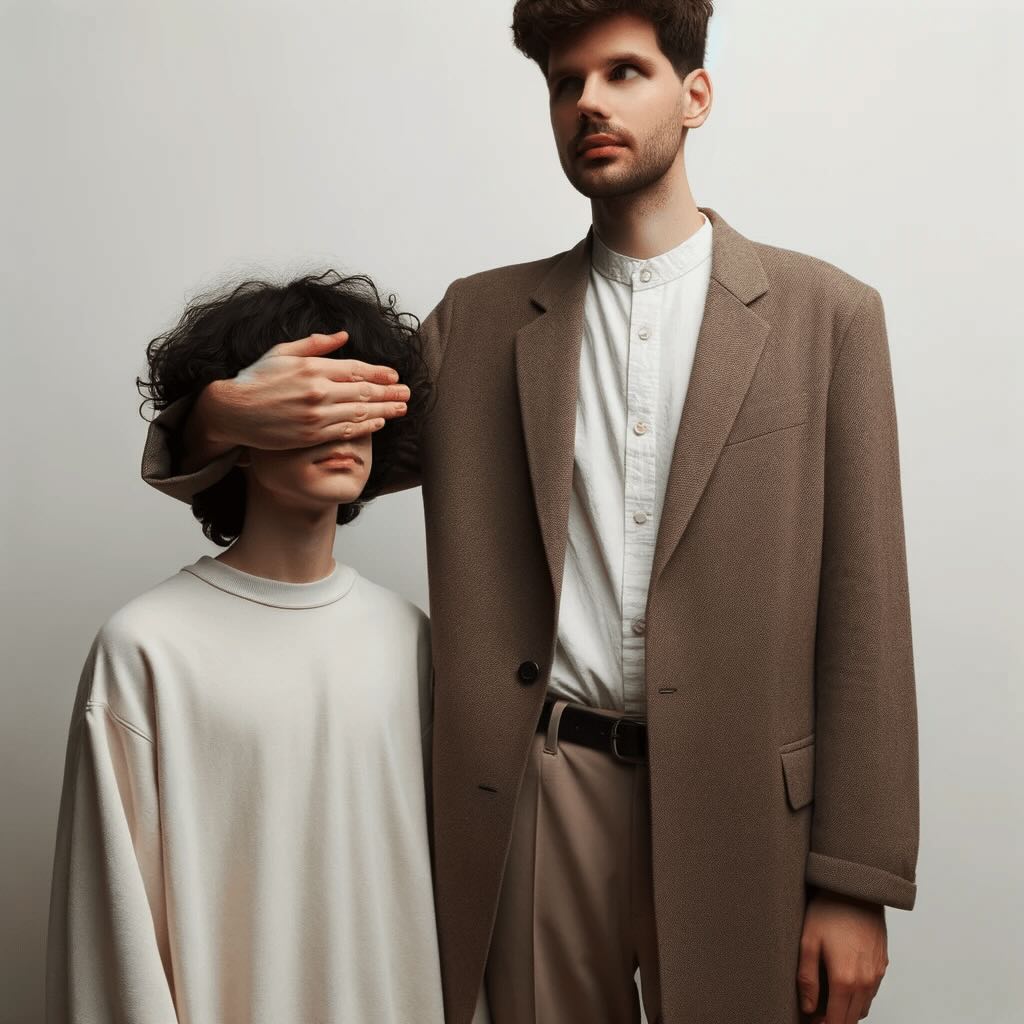} & Incorrect, the taller person is covering the shorter person's eyes. Instead, ... & \includegraphics[width=18mm,height=18mm]{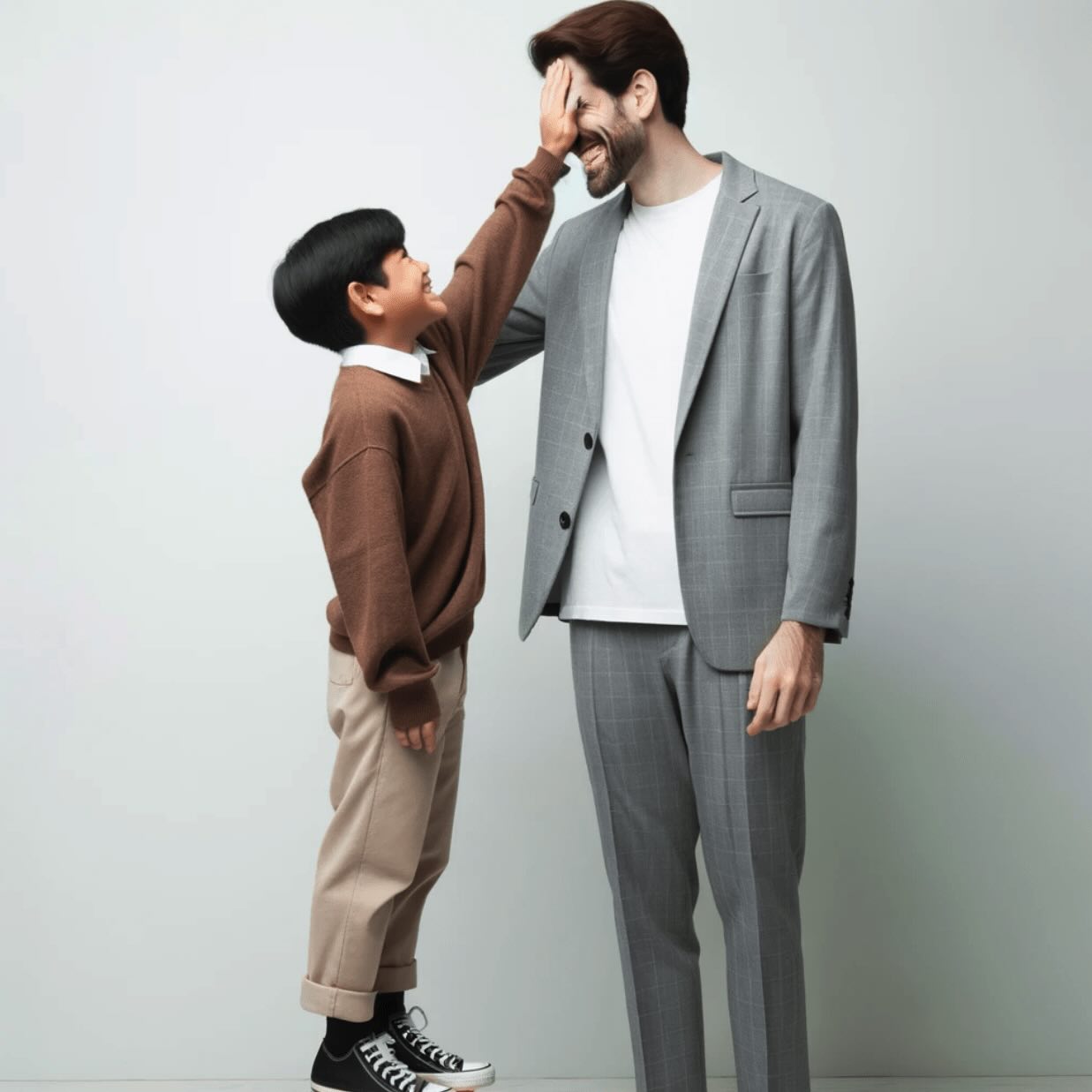} \\ \midrule
\Block{1-4}{{\bf Prompt inversion}} \\
\includegraphics[width=18mm,height=18mm]{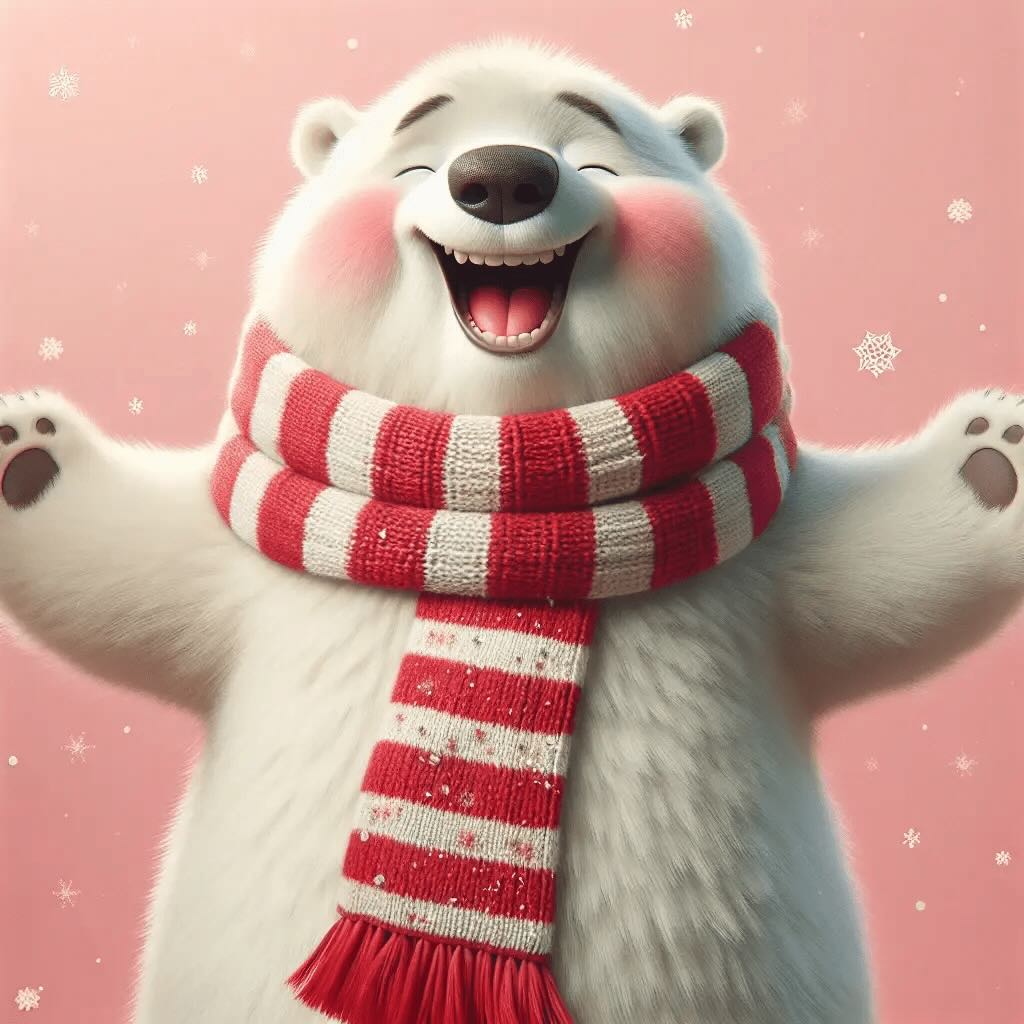} & \includegraphics[width=18mm,height=18mm]{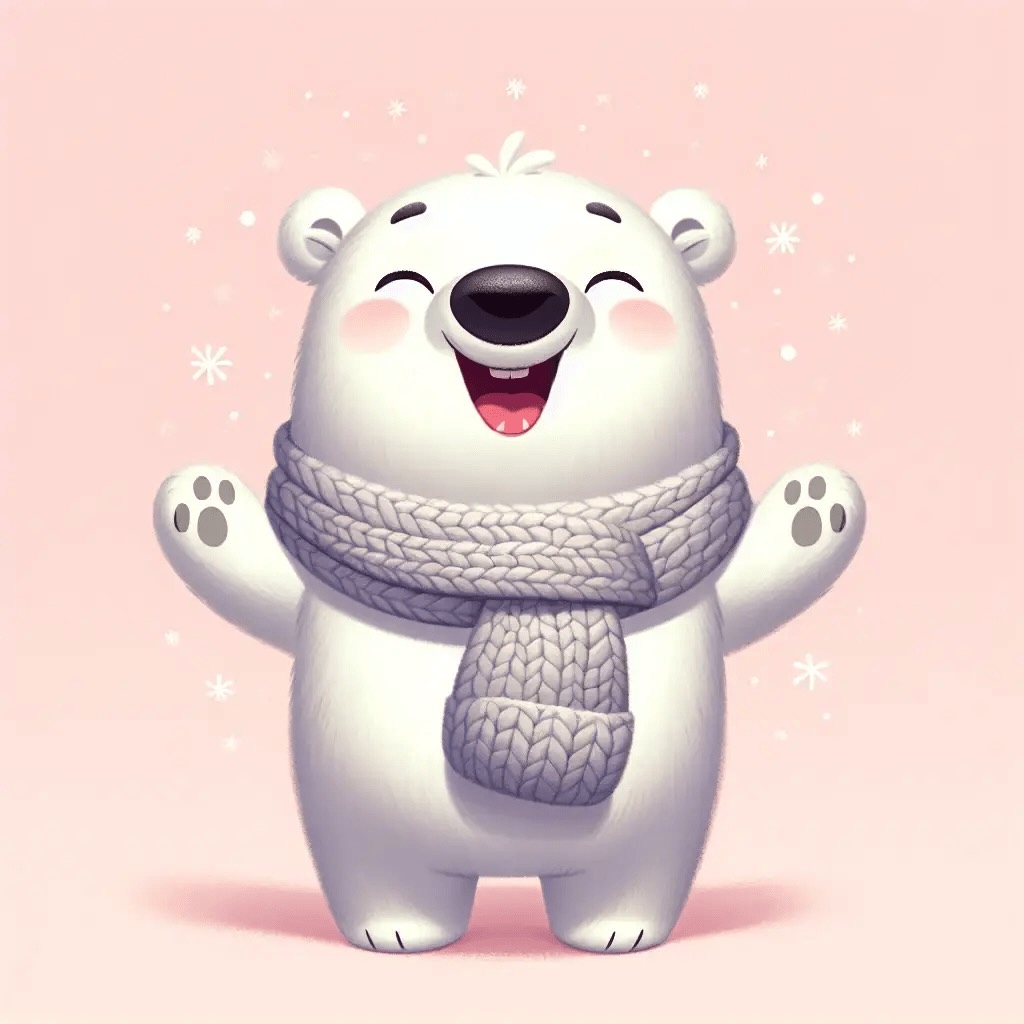} & The scarf should feature red and white stripes, and the fur is fluffy... & \includegraphics[width=18mm,height=18mm]{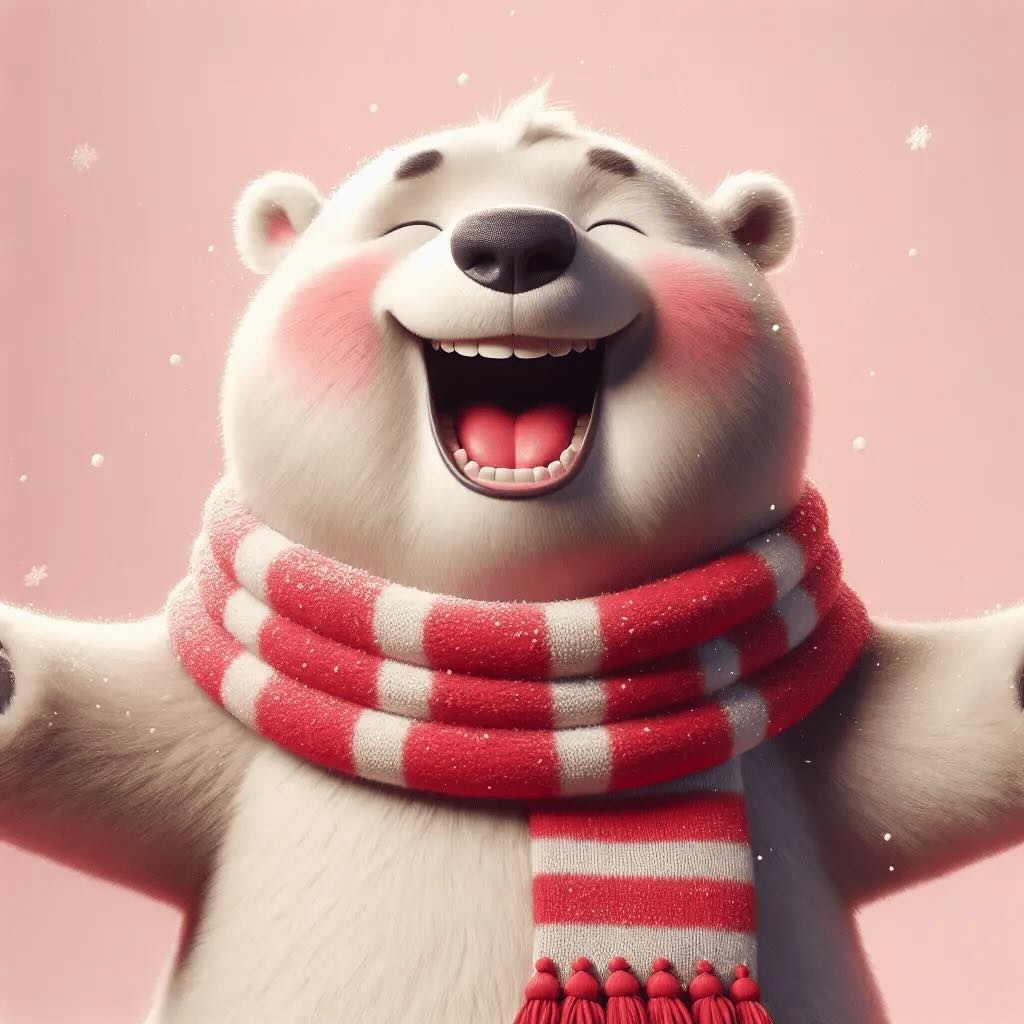} \\ 
\includegraphics[width=18mm,height=18mm]{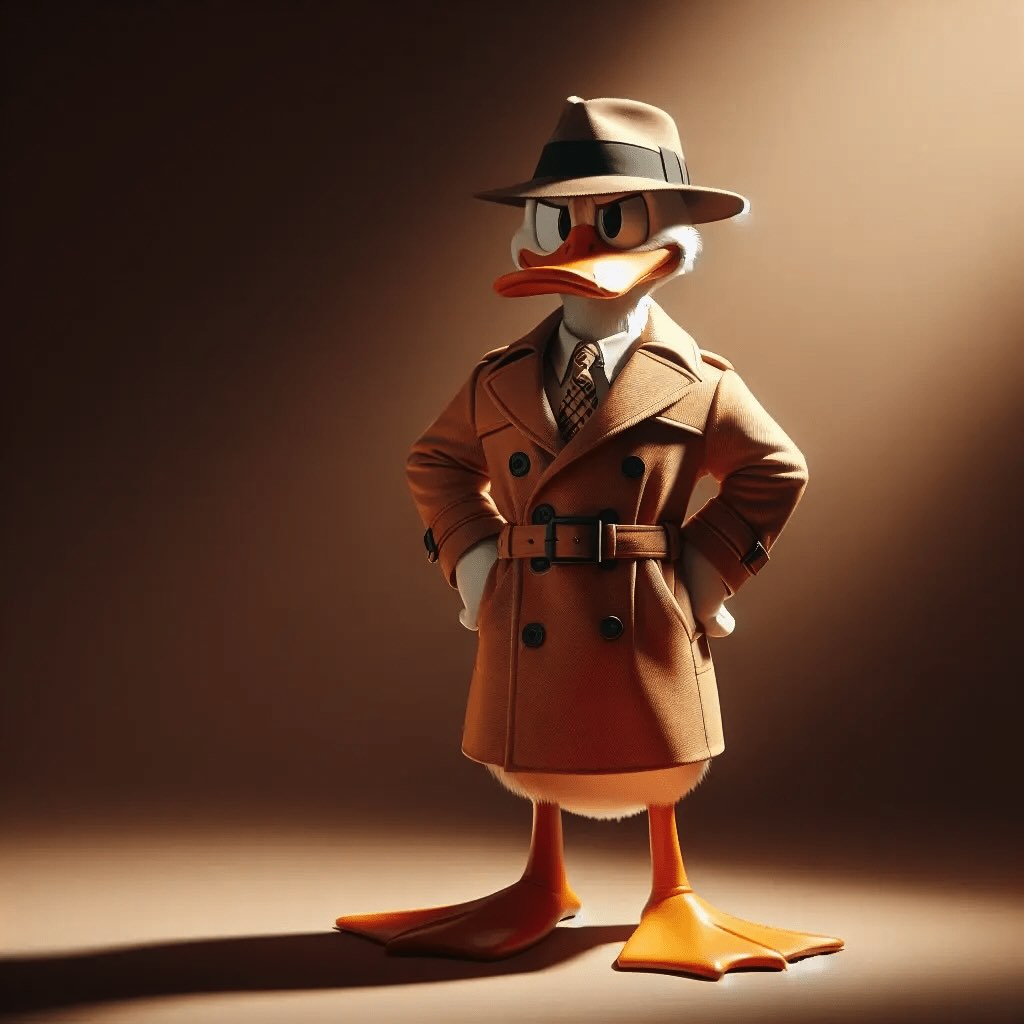} & \includegraphics[width=18mm,height=18mm]{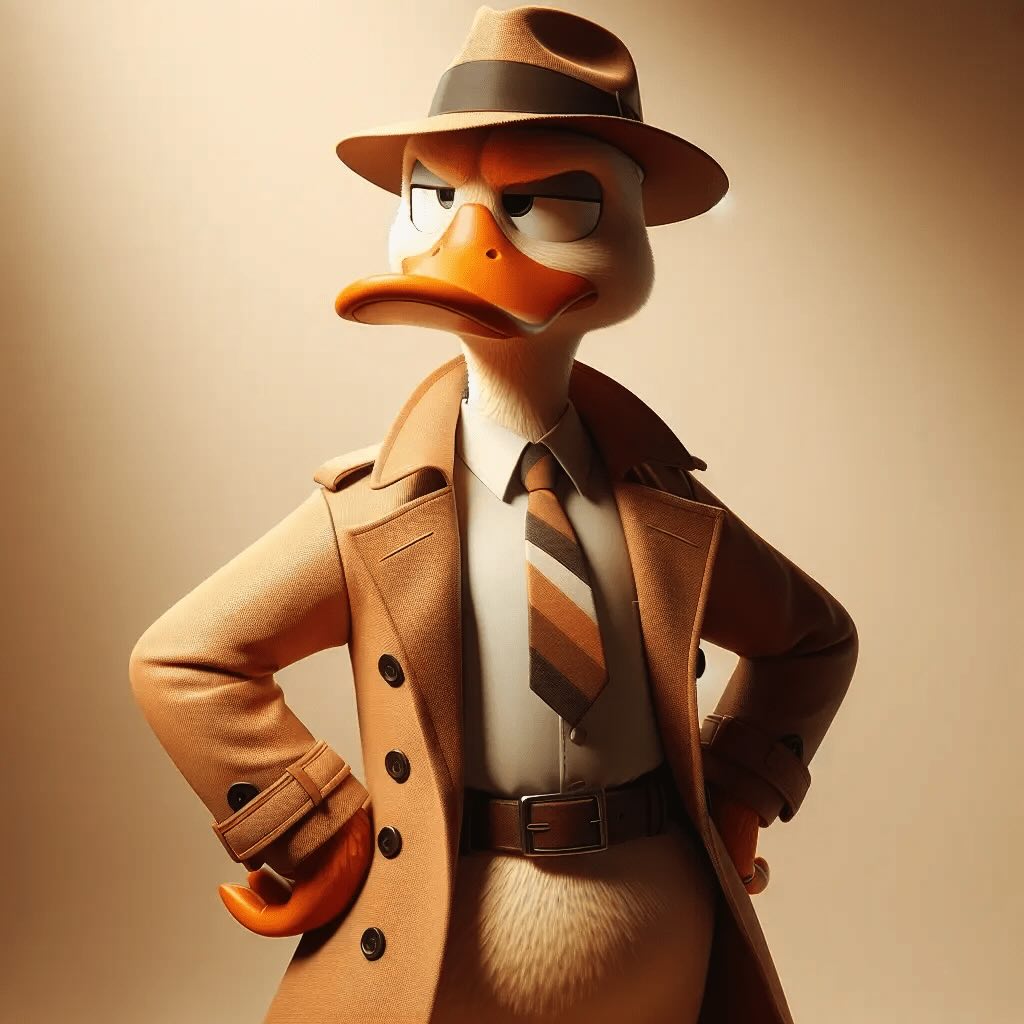} & The coat should be buttoned and the lighting exhibits a stronger contrast...  & \includegraphics[width=18mm,height=18mm]{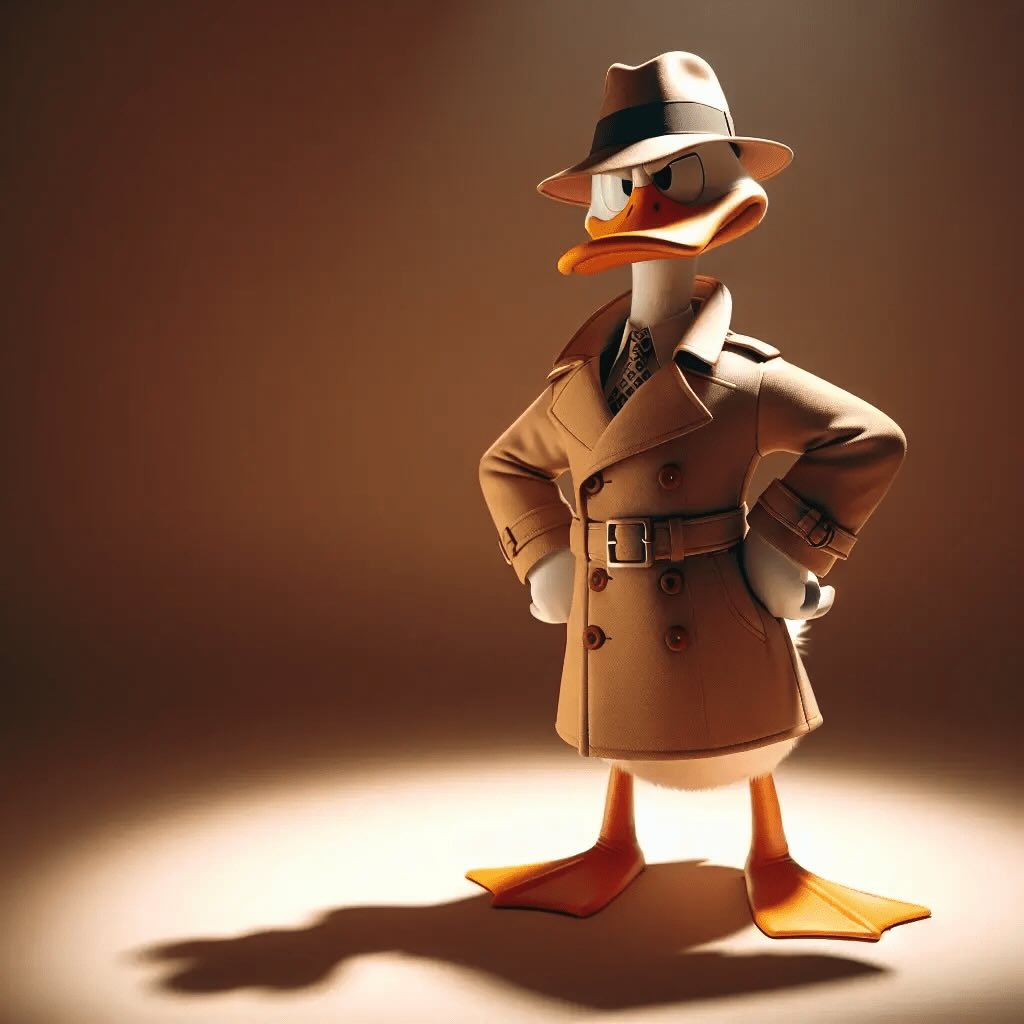} \\ 
\bottomrule[1.5pt]
\end{NiceTabular}
}
\caption{\textbf{Examples of T2I optimization.} We show that our framework (\autoref{fig:dalle3}) can automatically improve the faithfulness of images generated by DALL-E 3, with respect to user-specified textual topics (for T2I generation) or reference images (for prompt inversion). This is achieved through three rounds of prompt optimization, using feedback from the multimodal LLM (GPT4-V). \autoref{tab:more_t2i} and \autoref{tab:more_inversion} shows more examples with actual prompts.}
\vspace{-2mm}
\label{tab:opt_dalle}
\end{table}

\begin{table*}[h]
\centering
\scalebox{0.75}{
\begin{NiceTabular}{M{0.17\linewidth} | M{0.17\linewidth} M{0.17\linewidth} M{0.17\linewidth} M{0.17\linewidth} M{0.17\linewidth} M{0.17\linewidth}}
\CodeBefore
    \Body
\toprule[1.5pt]
\textbf{User Query} & \textbf{Inverted Image} & \textbf{Example 1} & \textbf{Example 2} & \textbf{Example 3} & \textbf{Example 4} & \textbf{Example 5} \\ \midrule
\includegraphics[width=30mm,height=30mm]{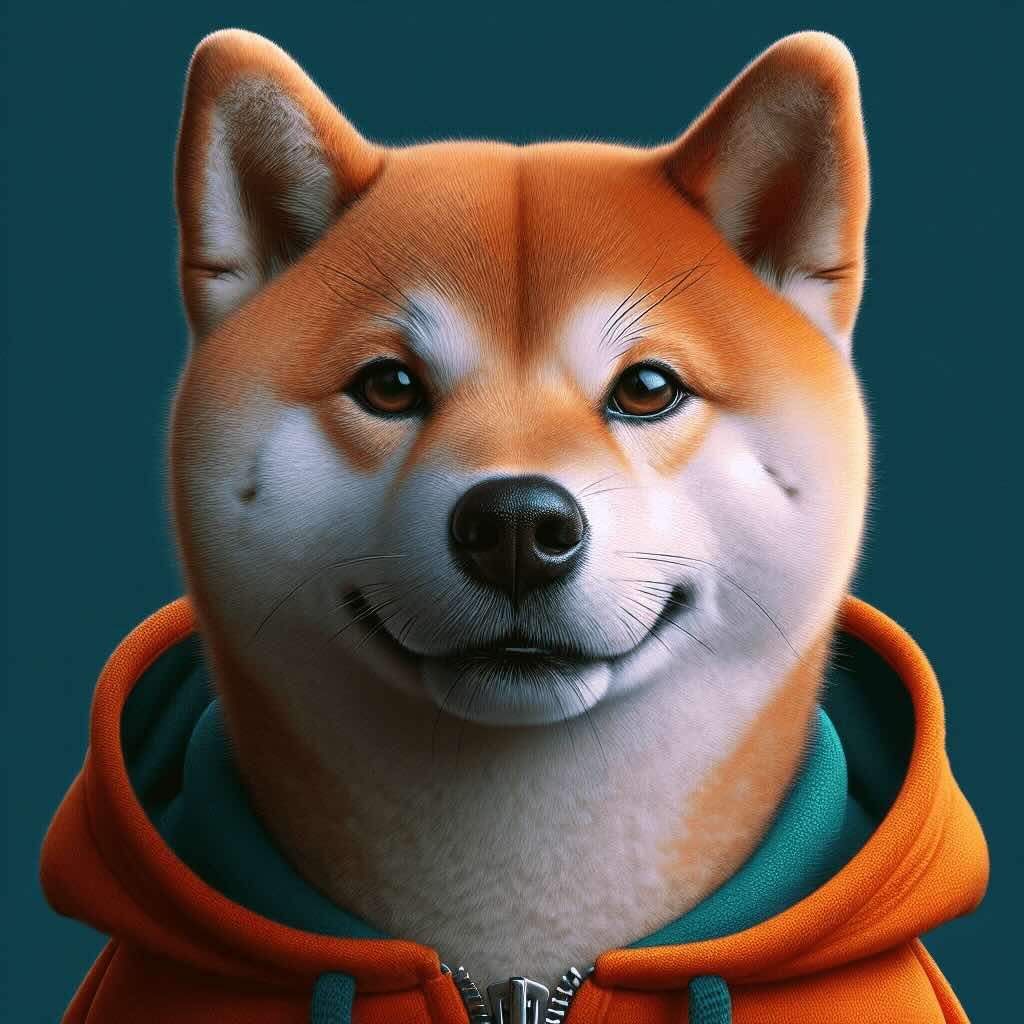} & \includegraphics[width=30mm,height=30mm]{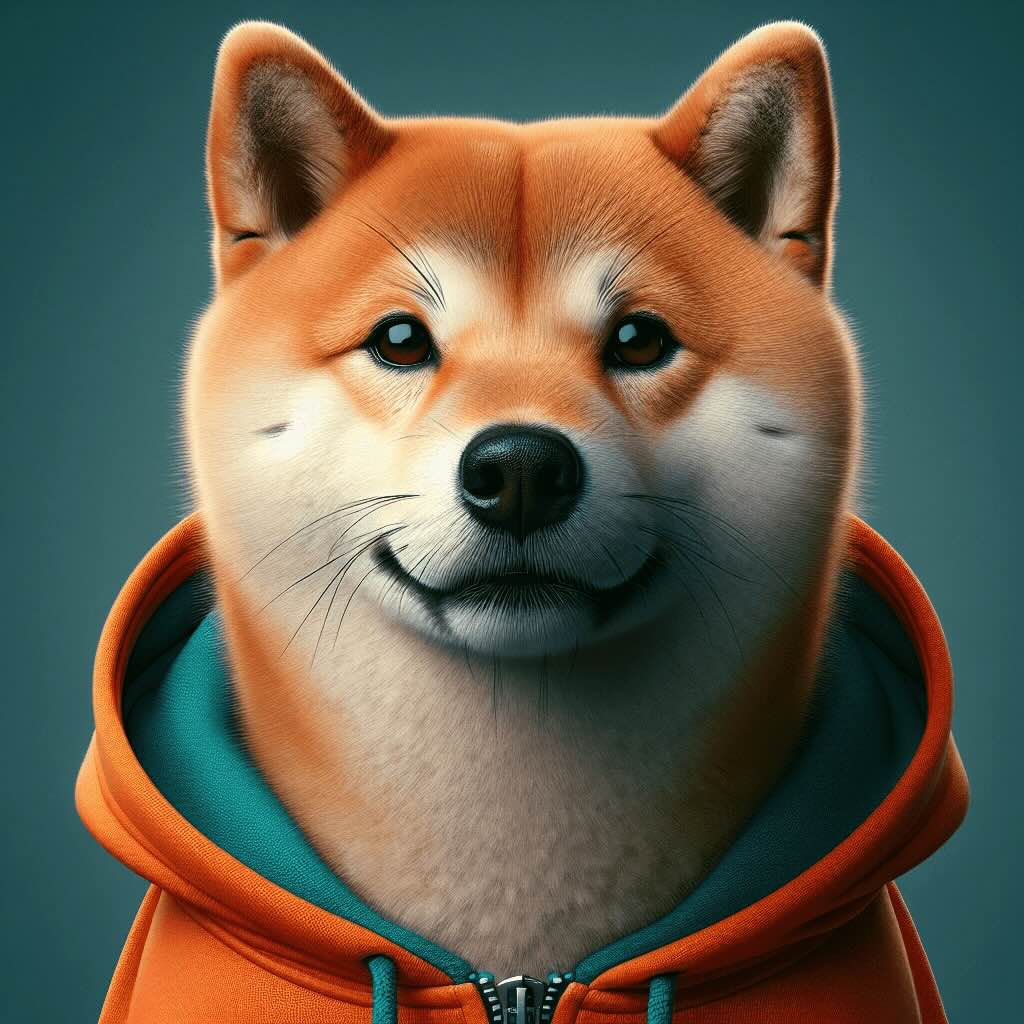} & \includegraphics[width=30mm,height=30mm]{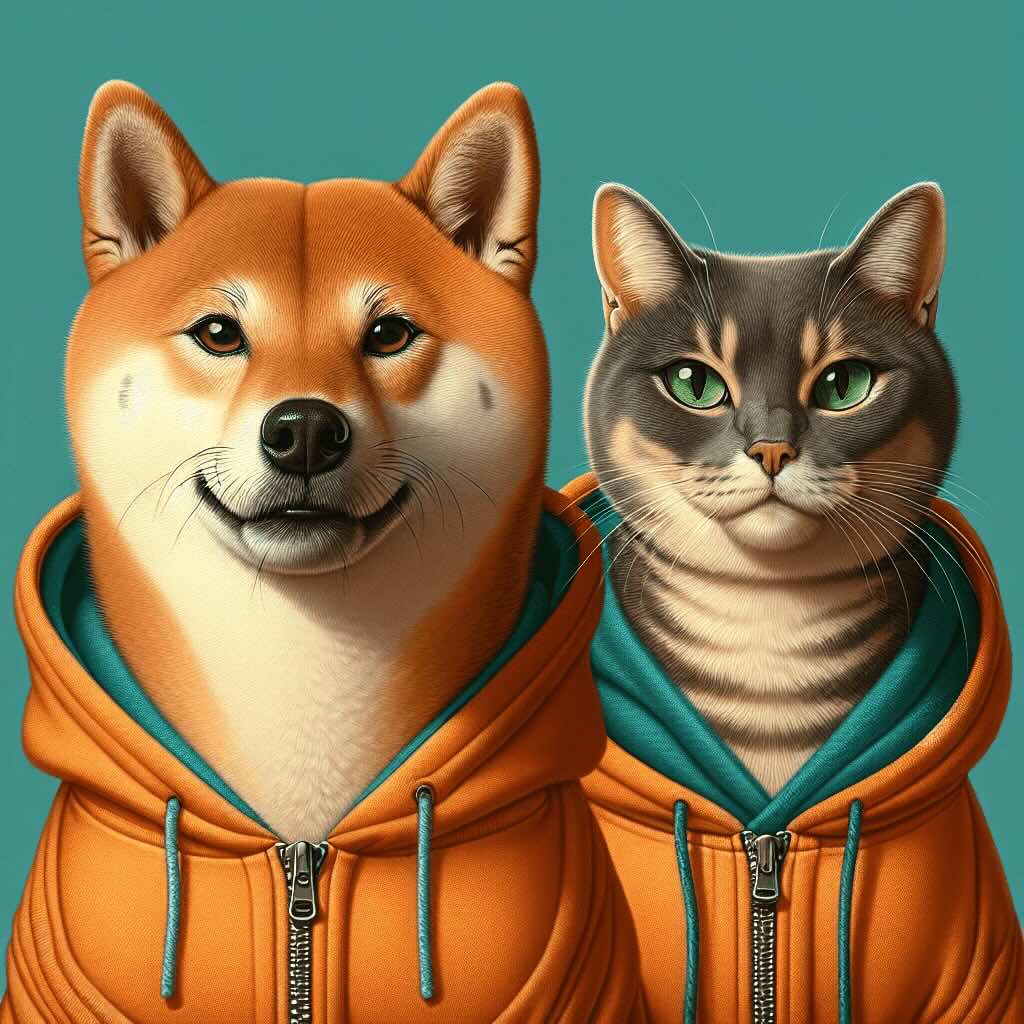} & \includegraphics[width=30mm,height=30mm]{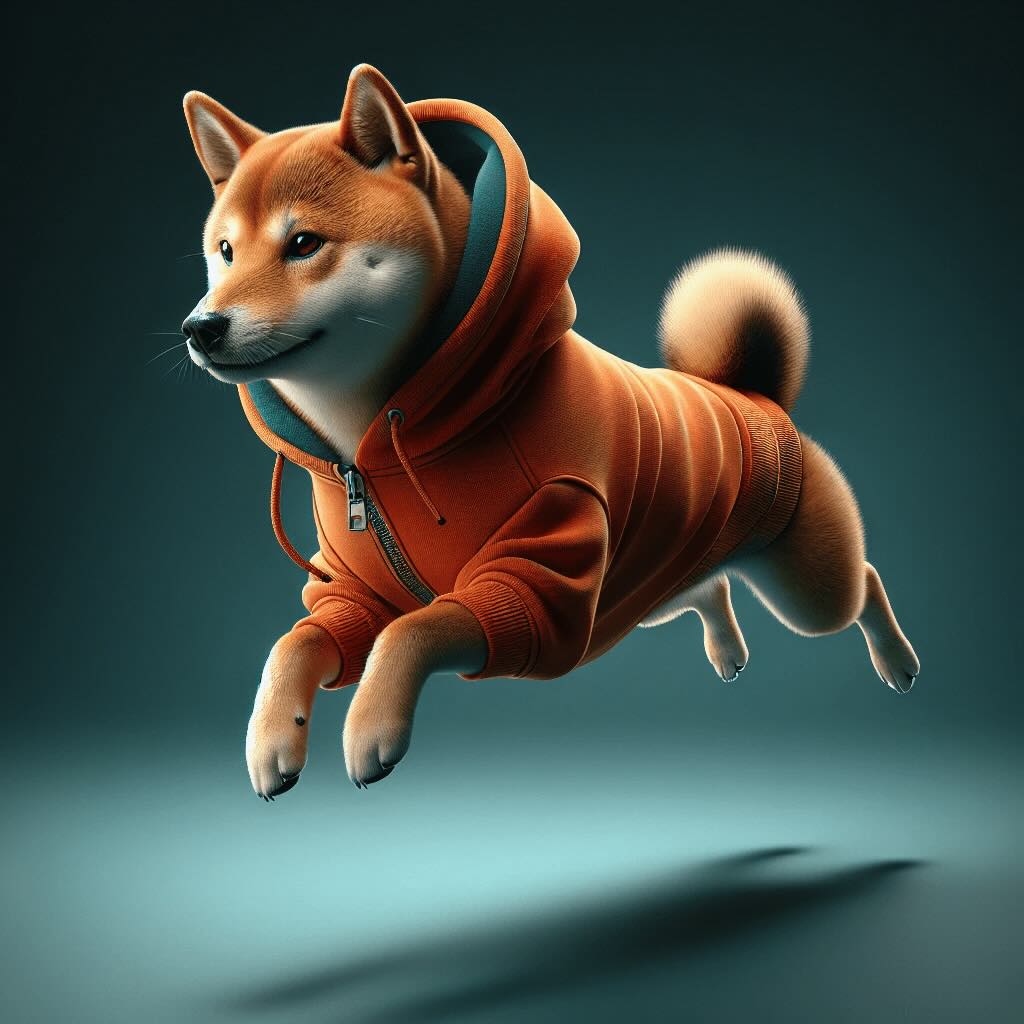} & \includegraphics[width=30mm,height=30mm]{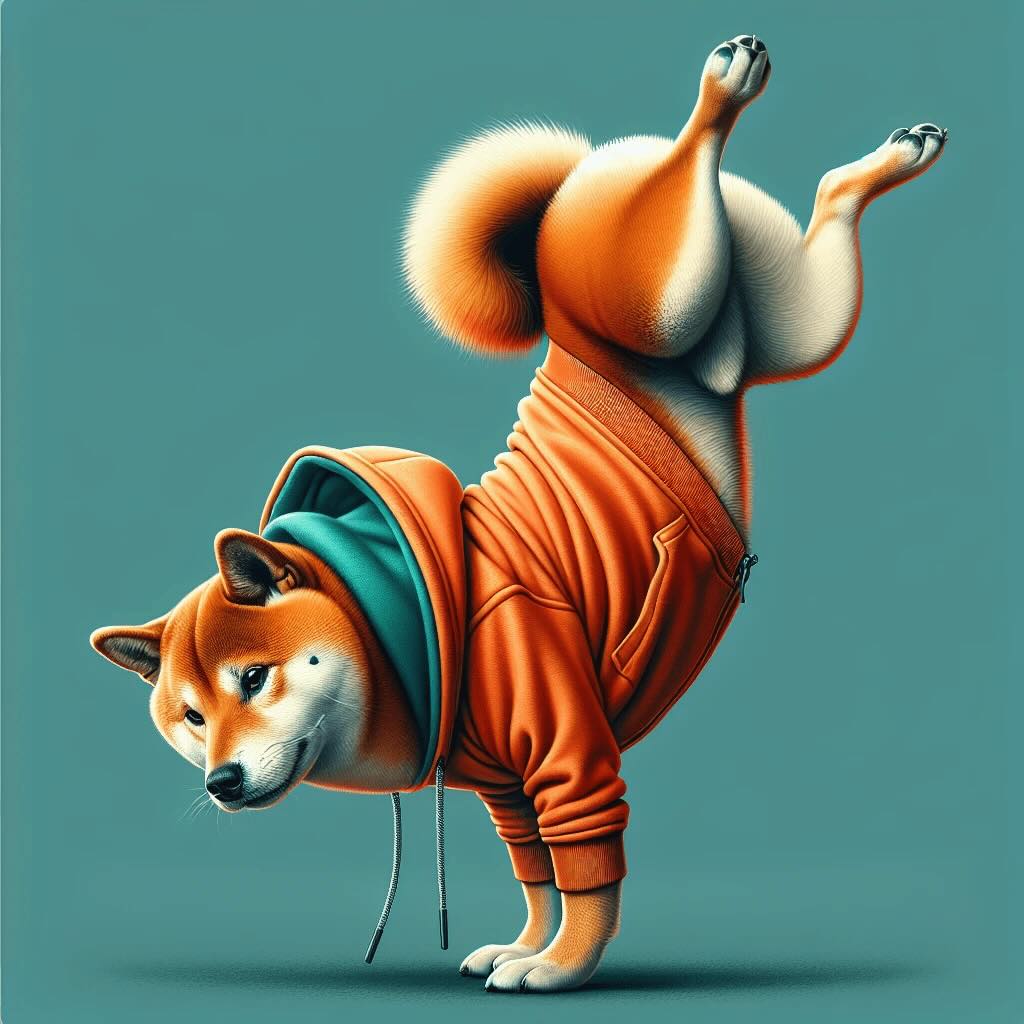} & \includegraphics[width=30mm,height=30mm]{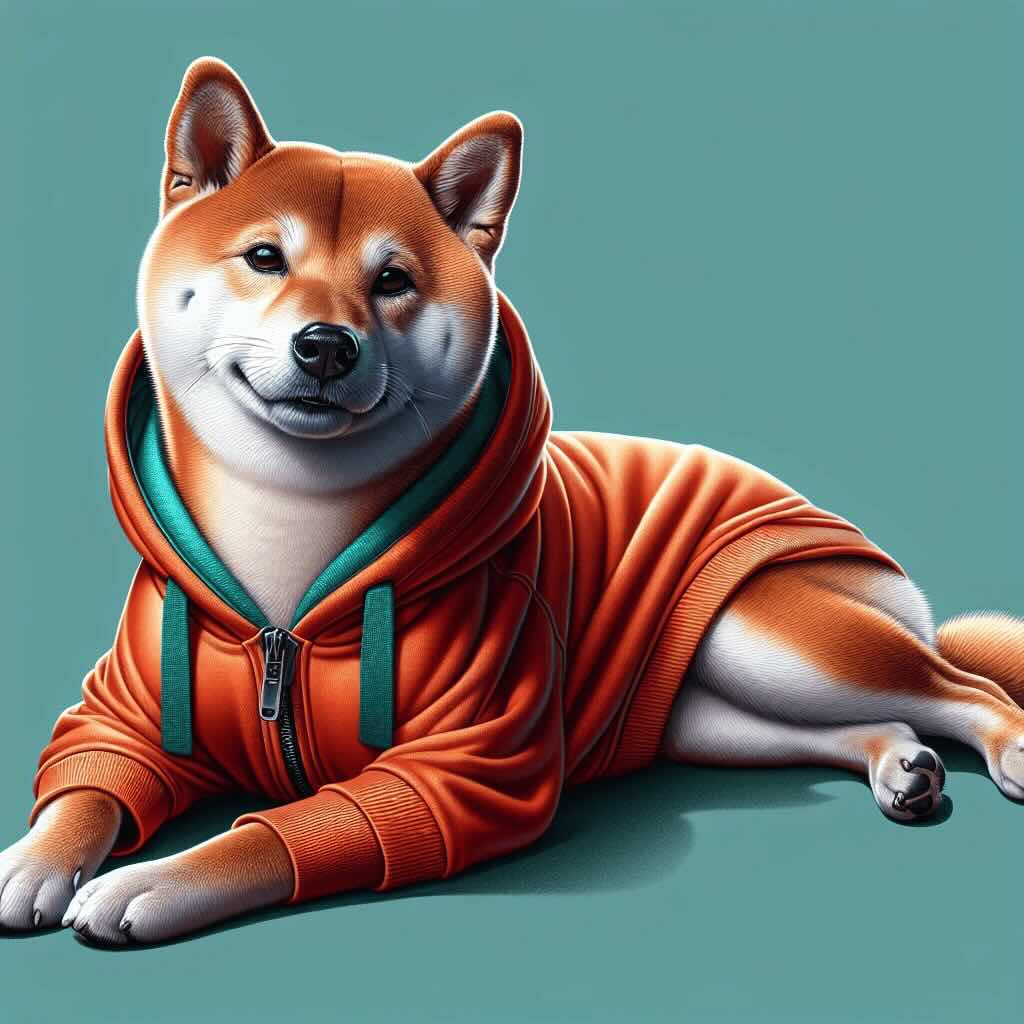} & \includegraphics[width=30mm,height=30mm]{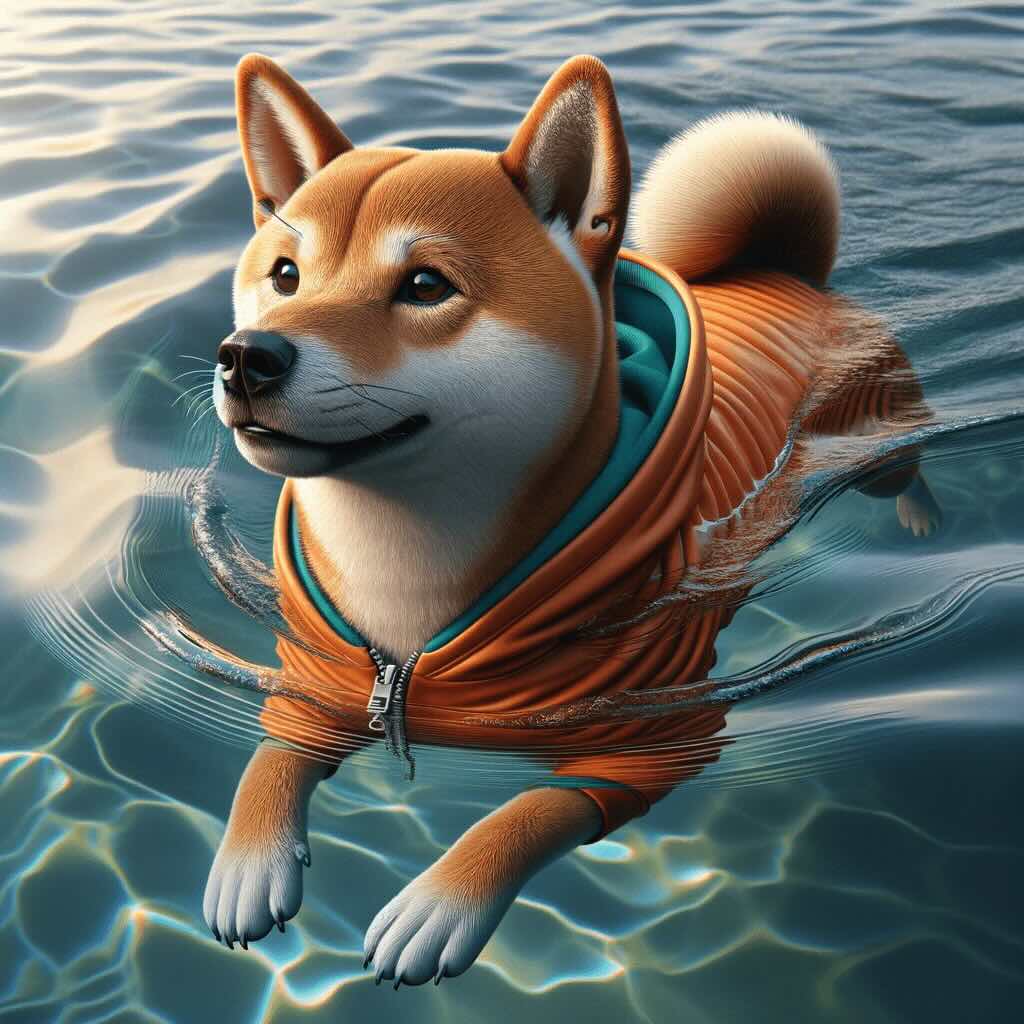} \\
&  & Give the dog a cat friend. & Make the dog be in the middle of a jump. & Make the dog do a handstand. & Make the dog lie down on its side. & Make the dog swim in water.  \\ 
\midrule
\includegraphics[width=30mm,height=30mm]{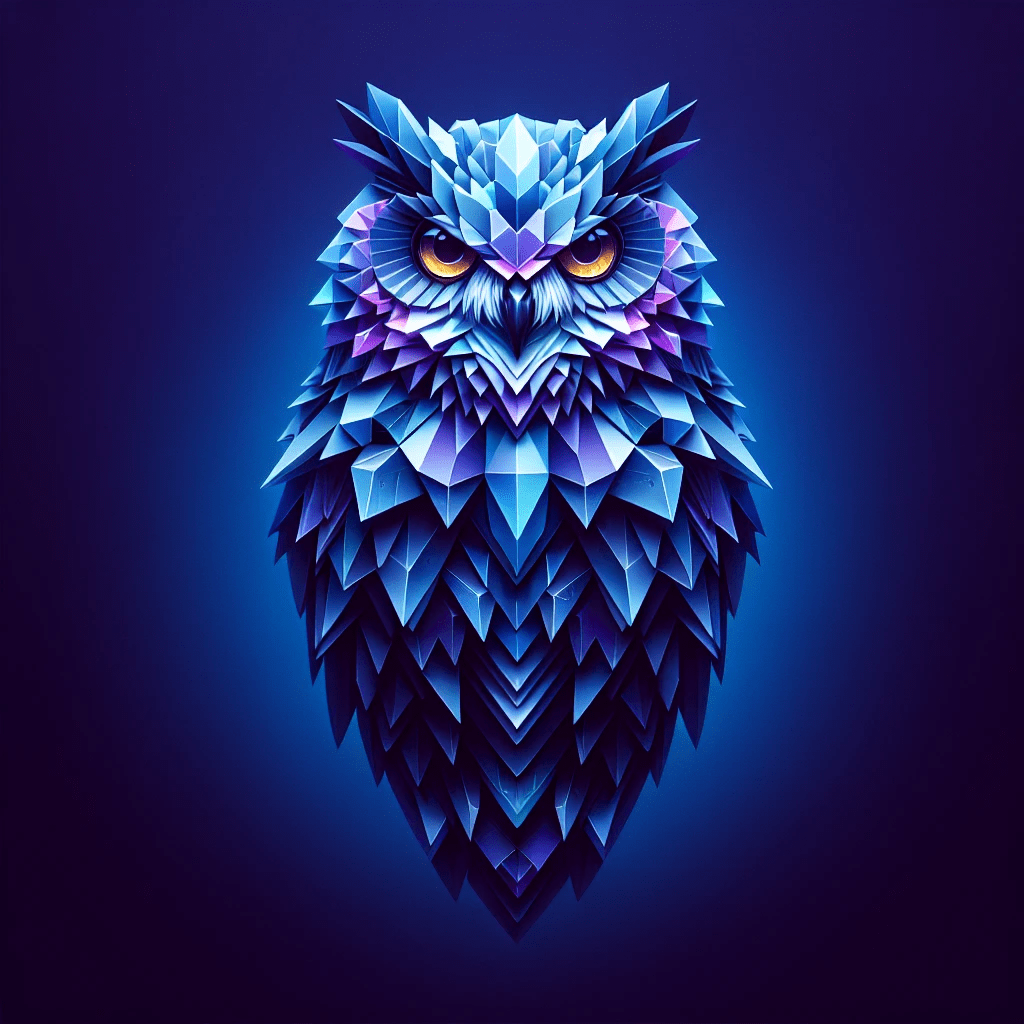} & \includegraphics[width=30mm,height=30mm]{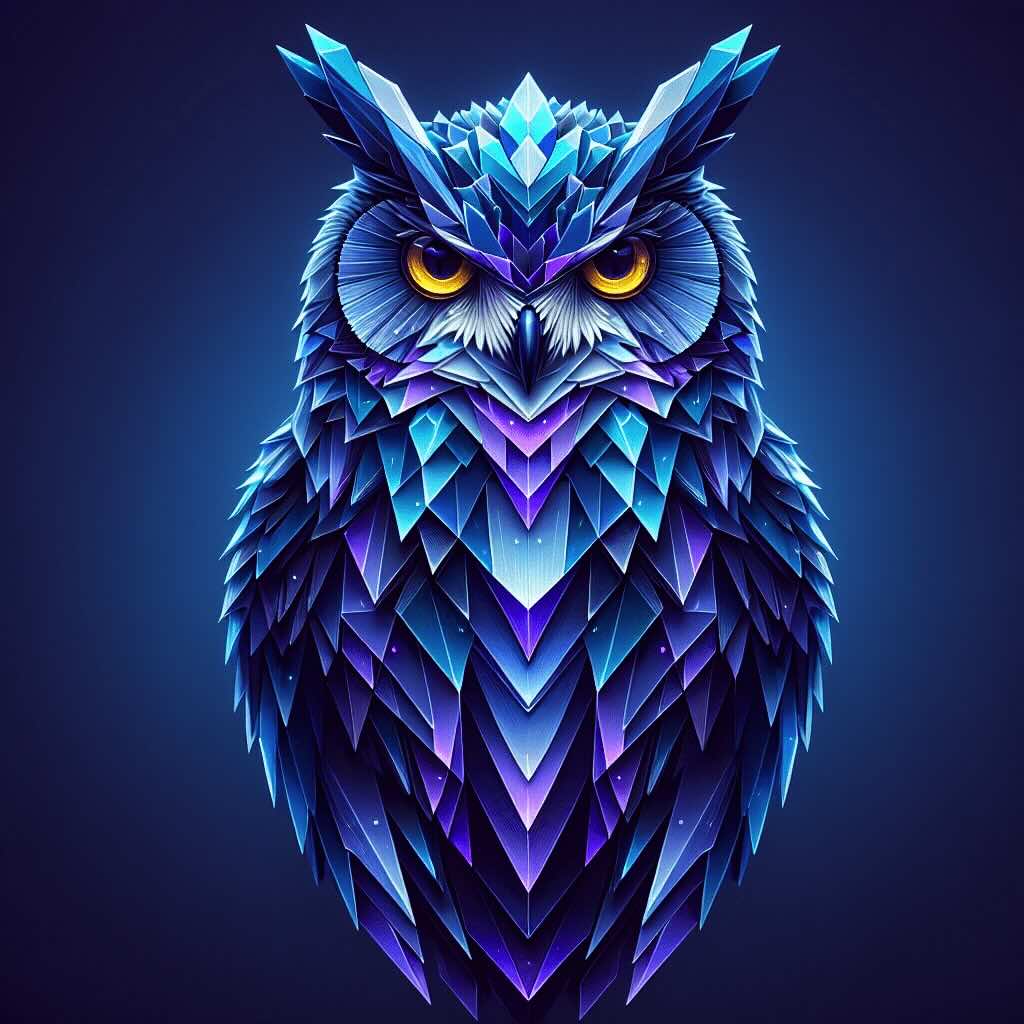} & \includegraphics[width=30mm,height=30mm]{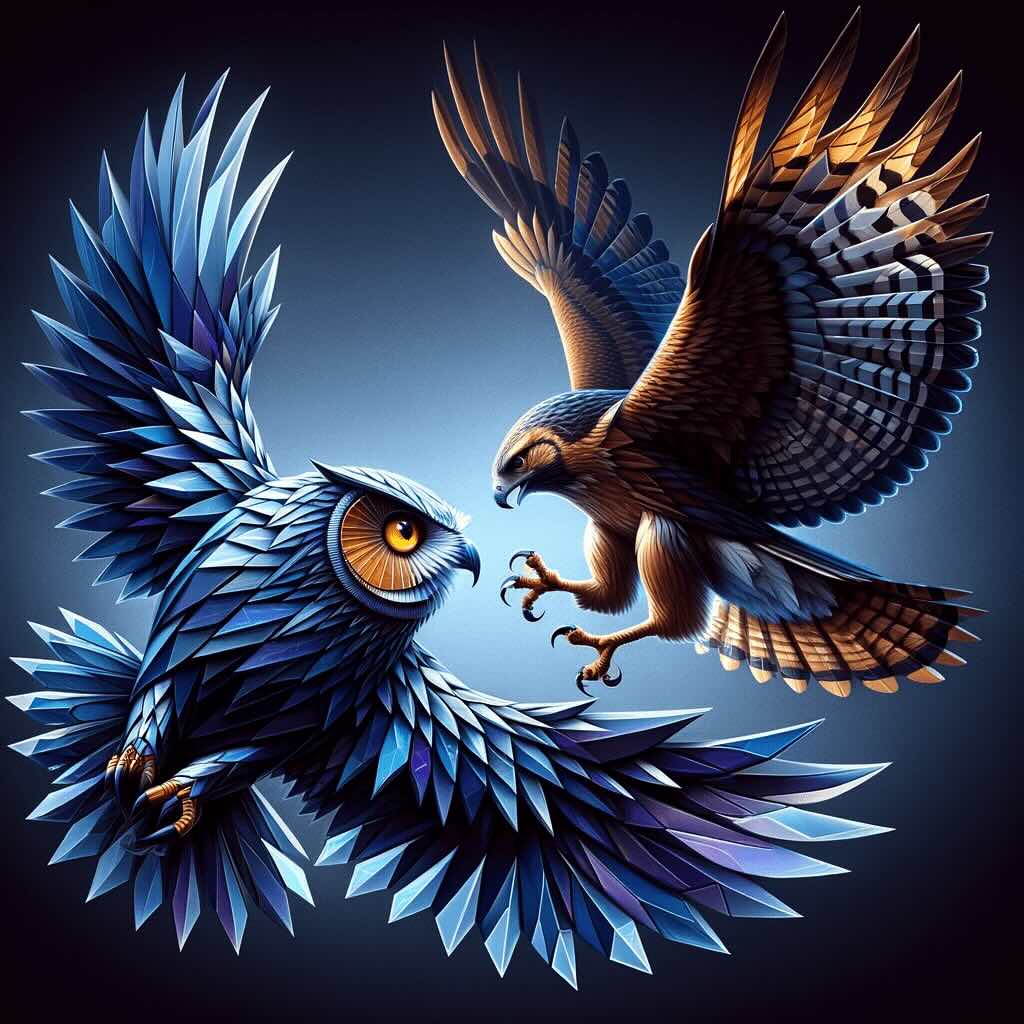} & \includegraphics[width=30mm,height=30mm]{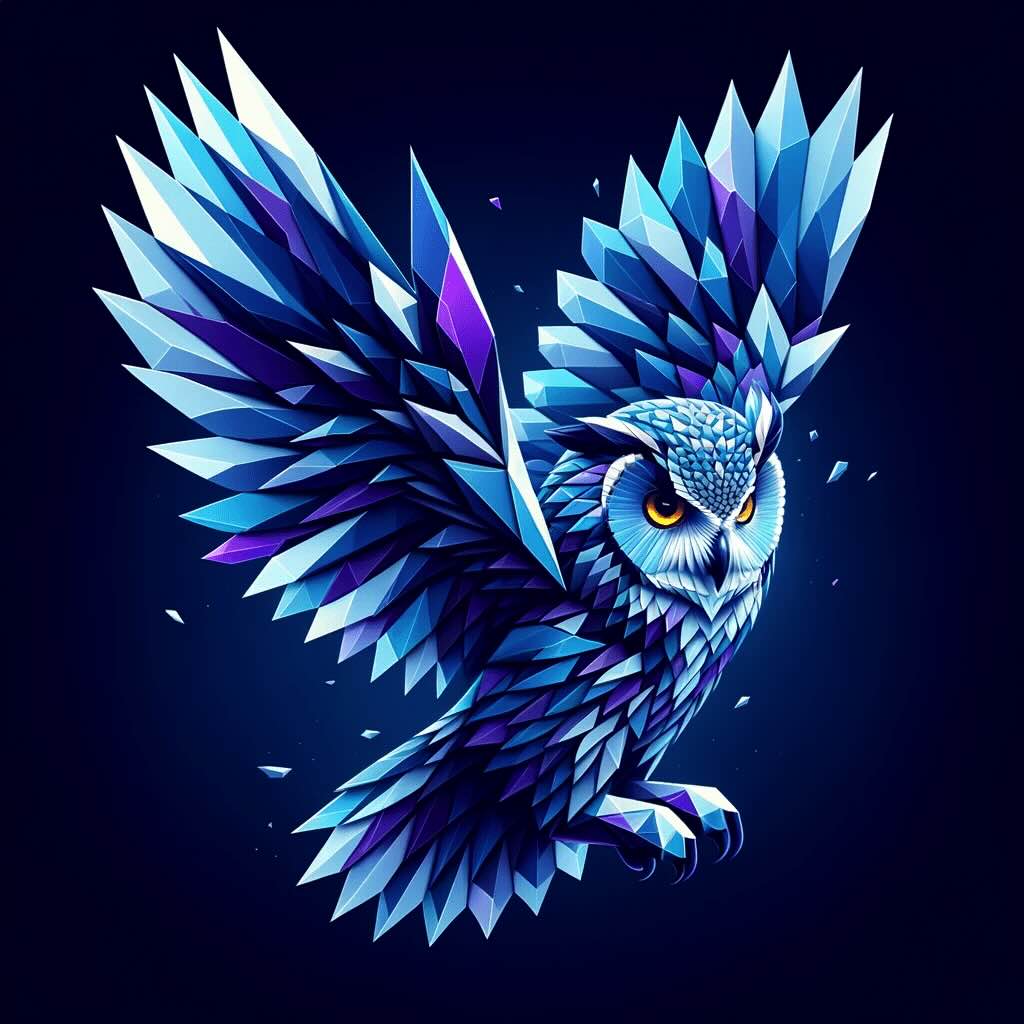} & \includegraphics[width=30mm,height=30mm]{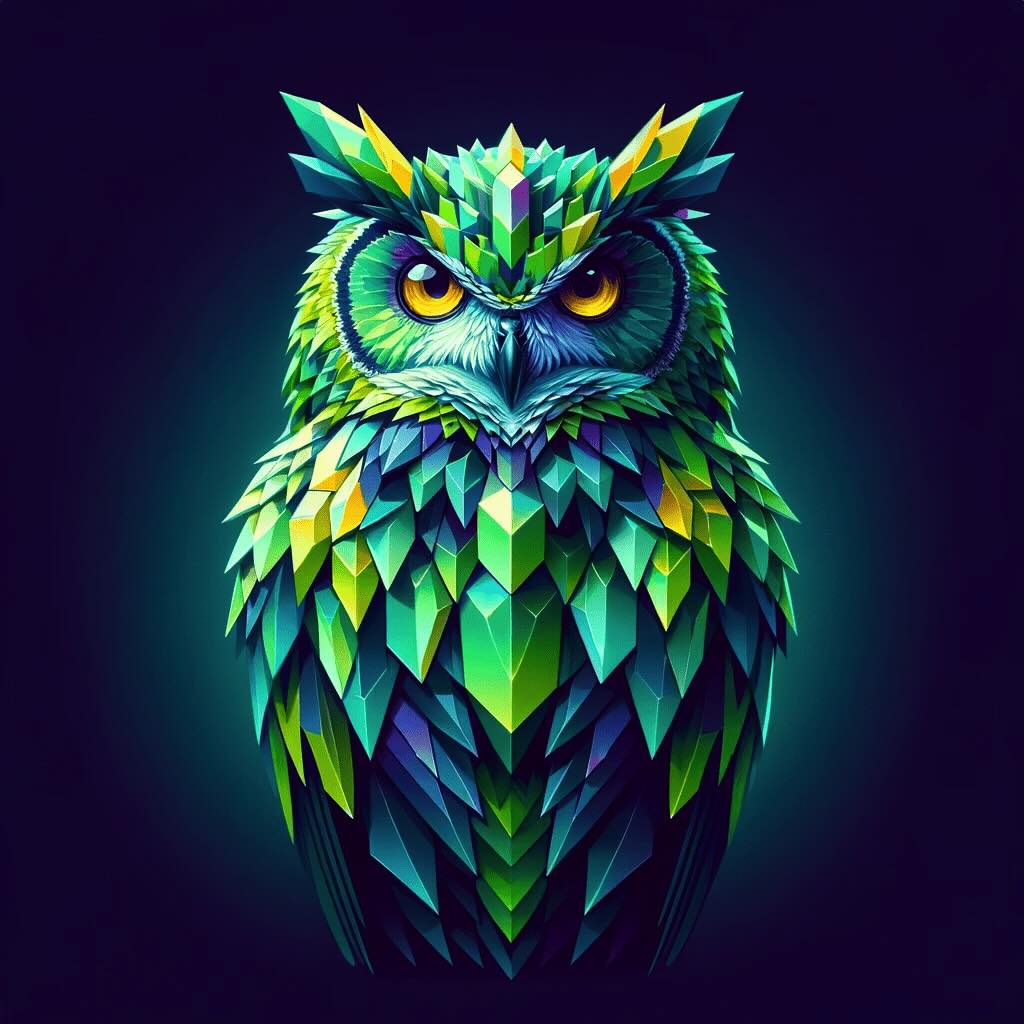} & \includegraphics[width=30mm,height=30mm]{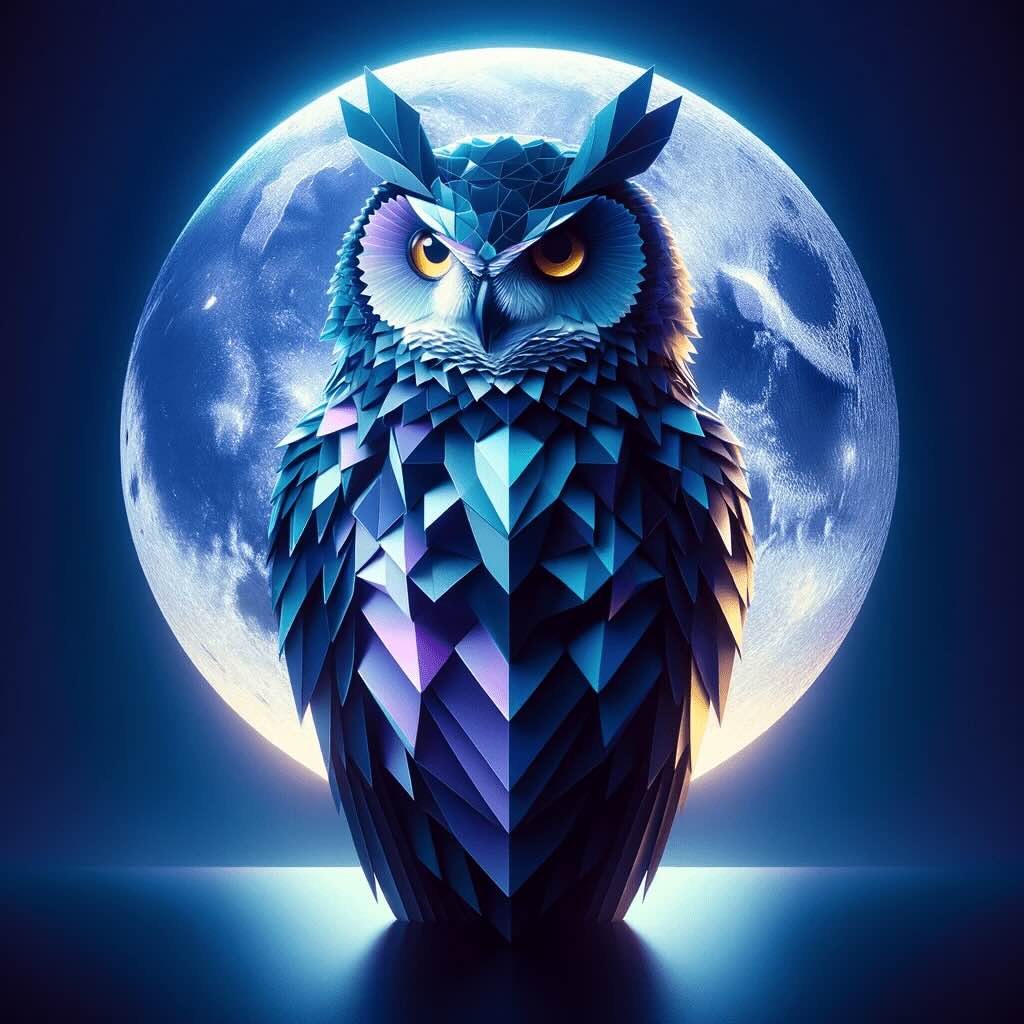} & \includegraphics[width=30mm,height=30mm]{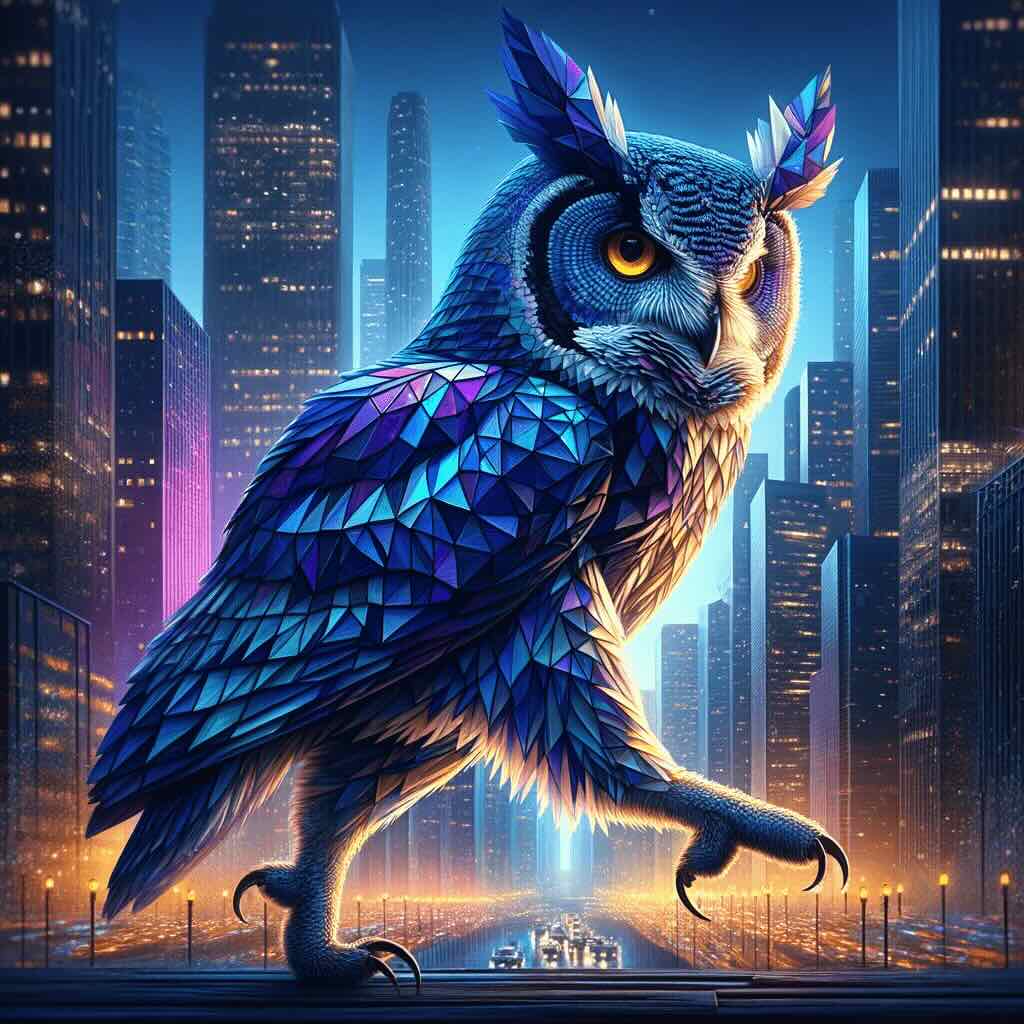} \\
&  & Make the owl fight a hawk. & Make the owl flap its wings. & Make the owl fully green. & Make the owl stand in front of the moon. & Make the owl walk in the city.  \\ 
\bottomrule[1.5pt]
\end{NiceTabular}
}
\caption{\textbf{Customization via prompt inversion.} Users can simply append extra descriptions to the inverted prompts to customize their main characters in queried images.}
\vspace{-2mm}
\label{tab:customization}
\end{table*}

{\bf Task setup.} For {\em T2I generation}, we experiment with a subset of 100 text queries from Winoground~\cite{winoground} that involve complex attribute and relation reasoning, which DALL-E might initially fail to generate. Our framework refines the prompts to capture the user-specified topics using a few (three) iterations. We also attempt a reverse task of {\em prompt inversion}: given a user-specified reference (query) image, our framework reverse-engineers the prompt to have DALL-E generate the same object or scene in the query image (see \autoref{fig:dalle3_inversion}). This enables users to easily make customizations~\cite{dreambooth} (see \autoref{tab:customization}), such as having the character in a reference image perform various actions or change scenes. For this task, we sample 100 random queries from DiffusionDB~\cite{wang2022diffusiondb}. We provide qualitative results in \autoref{tab:opt_dalle}, \autoref{tab:more_t2i}, and \autoref{tab:more_inversion}. We hire two volunteers to assess the faithfulness of the images generated by our method, and to compare these with the images manually prompted by two designers (each with one year of experience in AI content generation), as shown in \autoref{tab:human_study}.

{\bf Remarks on limitations.} While we show promising results, we note some failure cases in \autoref{tab:failure_t2i} and \autoref{tab:failure_inversion} due to the inherent limitations of foundation models. For example, GPT4-V might fail to describe abstract and artistic details, and DALL-E 3 often fails to generate the correct number of objects. We believe that our framework can benefit from more capable foundation models in the future.

\begin{table}[h]
\centering
\renewcommand{\arraystretch}{1.3}
\scalebox{0.74}{
\begin{tabular}{c|c|ccc}
\toprule[1.5pt]
Task                               & Method & Init. (std) & Final (std)  & $\Delta$       \\ \midrule
                                   & Human  & 2.28 (.45)              & 2.86 (.61)                        & {\textcolor{red}{0.58}} \\
\multirow{-2}{*}{Text-to-Image}    & Ours   & 2.62 (.36)              & {3.56 (.54)} & {\textcolor{green}{0.94}} \\ \midrule
                                   & Human  & 1.58 (.48)               & 2.76 (.53)                        & {\textcolor{red}{1.18}} \\
\multirow{-2}{*}{Prompt Inversion} & Ours   & 1.94 (.39)                & {3.68 (.47) } & {\textcolor{green}{1.74}} \\ 
\bottomrule[1.5pt]
\end{tabular}
}
\caption{{\bf Our method enhances faithfulness in T2I generation.} We hire two human annotators to assess the faithfulness of images generated from user queries, e.g., textual topics for Text-to-Image, or reference images for Prompt Inversion. The scores are measured on a 1-to-5 Likert scale, with 1 signifying contradiction and 5 indicating perfect alignment with the user's goal. Our approach benefits from three iterations of prompt optimization and consistently outperforms human-engineered prompts by designers who have one year of experience in AI content generation. }
\vspace{-5mm}
\label{tab:human_study}
\end{table}

\section{Discussion and Limitations}
\label{sec:discussion}

{\bf Summary.} We present the first attempt to leverage LLMs as prompt engineers for VLMs. For one-shot image classification, our method surpasses human-engineered prompts and even rivals white-box approaches. Central to the success of our method is the utilization of conversational feedback, enabling chat-based LLMs to efficiently steer VLMs in the right direction. This process leads to naturally interpretable prompts bearing considerable resemblance to those crafted by humans. Importantly, our natural language prompting setup is a lot more constrained than the assumed scenarios of previous white-box or even some black-box settings~\cite{oh2023blackvip}, because we do not require the model weights and outputs of VLMs. 
Finally, our framework can be extended to generative tasks using the state-of-the-art black-box DALL-E 3.

{\bf Limitations and future work.}
While we try to minimize the overall cost and the total number of API calls, the energy consumption associated with LLMs remains a substantial concern. It is vital to note that we do not intend to compete directly with white-box baselines that can improve visual and text representations with more data. Further details on the higher-shot performance of our method can be found in \autoref{sec:appendix_experiments}. VLMs are trained on noisy and imbalanced web data~\cite{parashar2024neglected}, which may result in biased performance~\cite{mehrabi2021survey}. Lastly, we are limited to costly human evaluation for T2I generation in this study. Future work may adopt automatic evaluation~\cite{lin2024evaluating, lin2023revisiting} for large-scale experiments.

\section{Details of Conversing with ChatGPT}
\label{sec:appendix_setup}

{\bf Multi-turn conversation.} We use ChatGPT to generate a set of new prompts based on the top and bottom performing prompts (line 10 of Algorithm~\autoref{alg:method}). The exact prompts we use are:

\begin{mdframed}
\texttt{Hi ChatGPT, assume you are a pattern learner. I have two lists of CLIP templates: one with good templates and the other with bad templates. There are latent patterns that make a template good or bad. Based on these patterns, give me a better template for image classification while avoiding worse template. \\
Here is the list of good templates:\\
- \textcolor{blue}{good1} \\
- \textcolor{blue}{good2} \\
- \dots \\
Here is the list of bad templates:  \\
- \textcolor{blue}{bad1} \\
- \textcolor{blue}{bad2} \\
- \dots \\
Here are my requirements: \\
- Please only reply with the template. \\
- The template should be fewer than 15 words. \\
- The template should have a similar structure to the above templates. \\
\\
\textcolor{red}{Positive Response (if the new prompt outperforms the top-k)} \\
The performance of the template ``\textcolor{blue}{newTemplate}'' improves to X.XX\%. Please give me a better template. \\
\\
\textcolor{red}{Negative Response} \\
The performance of the template ``\textcolor{blue}{newTemplate}'' drops to X.XX\%. Please give me a better template.\\
}
\end{mdframed}

\begin{figure}[h]
  \vspace{-4mm}
  \centering
    \begin{subfigure}{\columnwidth}
    \includegraphics[width=\textwidth]{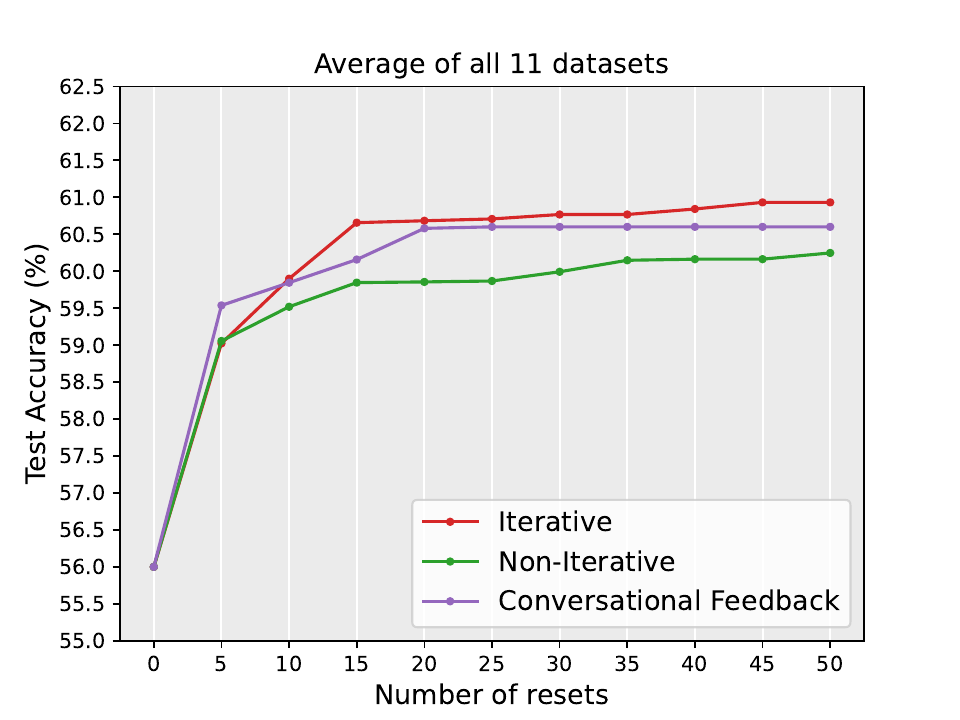}
  \end{subfigure}\hfill
  \begin{subfigure}{\columnwidth}
    \includegraphics[width=\textwidth]{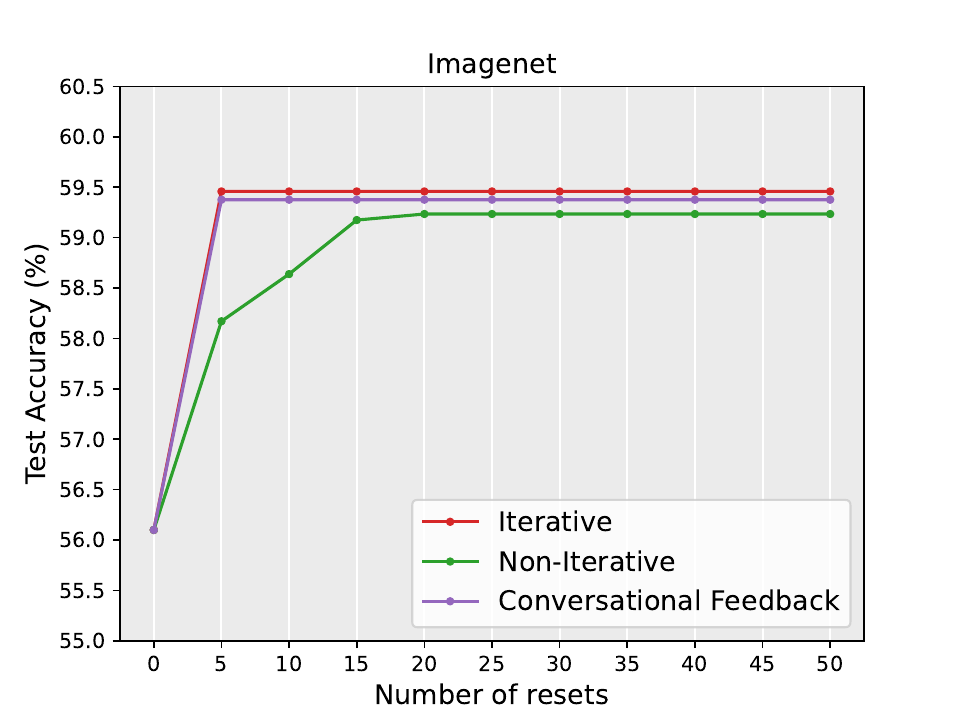}
  \end{subfigure}
  \vspace{-4mm}
  \caption{\textbf{Updating initial prompts can be as effective as multi-turn conversation.} We ablate different ways of conversing with ChatGPT on all 11 datasets (left) and ImageNet (right). Notably, we find that only updating the top-k and bottom-k prompts ({\bf Iterative}) is as performant and thus a cheaper alternative because sending response to ChatGPT costs more input tokens. On the other hand, reusing the initial prompts ({\bf Non-Iterative}) leads to worse overall performance.}
  \label{fig:multi_turn}
  \vspace{-4mm}
\end{figure}

\textbf{Alternative implementation: sending only the initial prompts (default).} Multi-turn conversation requires appending all chat history to ChatGPT's official API at every iteration, which costs more input tokens and money. In \autoref{fig:multi_turn}, we show that one can only send the initial prompts (without any response) to ChatGPT at every iteration to get equivalent and even slightly better performance. However, it is important to also update the top-k and bottom-k prompts at every iteration ({\bf Iterative}) for efficiency. We show that the {\bf Non-Iterative} version that keeps re-using the initial top-k and bottom-k prompts leads to worse performance. Therefore, in our paper, we stick to {\bf Iterative} for all experiments.

\newpage
\textbf{Positive Only (P only).} When using only positive prompts, we can remove negative prompts and provide twice as many positive examples:

\begin{mdframed}
\texttt{Hi ChatGPT, assume you are a pattern learner. I have one list of CLIP templates: one with good templates. There are latent patterns that make a template good. Based on these patterns, give me a better template for image classification. \\
Here is the list of good templates:\\
- \textcolor{blue}{good1} \\
- \textcolor{blue}{good2} \\
- \dots \\
Here are my requirements: \\
- Please only reply with the template. \\
- The template should be fewer than 15 words. \\
- The template should have a similar structure to the above templates. }
\end{mdframed}

\section{Additional Experimental Results}
\label{sec:appendix_experiments}

In this section, we present additional experiments to gain further insights into our method.

\begin{figure}[h]
    \centering

    \includegraphics[width=\columnwidth, clip=true,trim = 0mm 0mm 0mm 0mm]{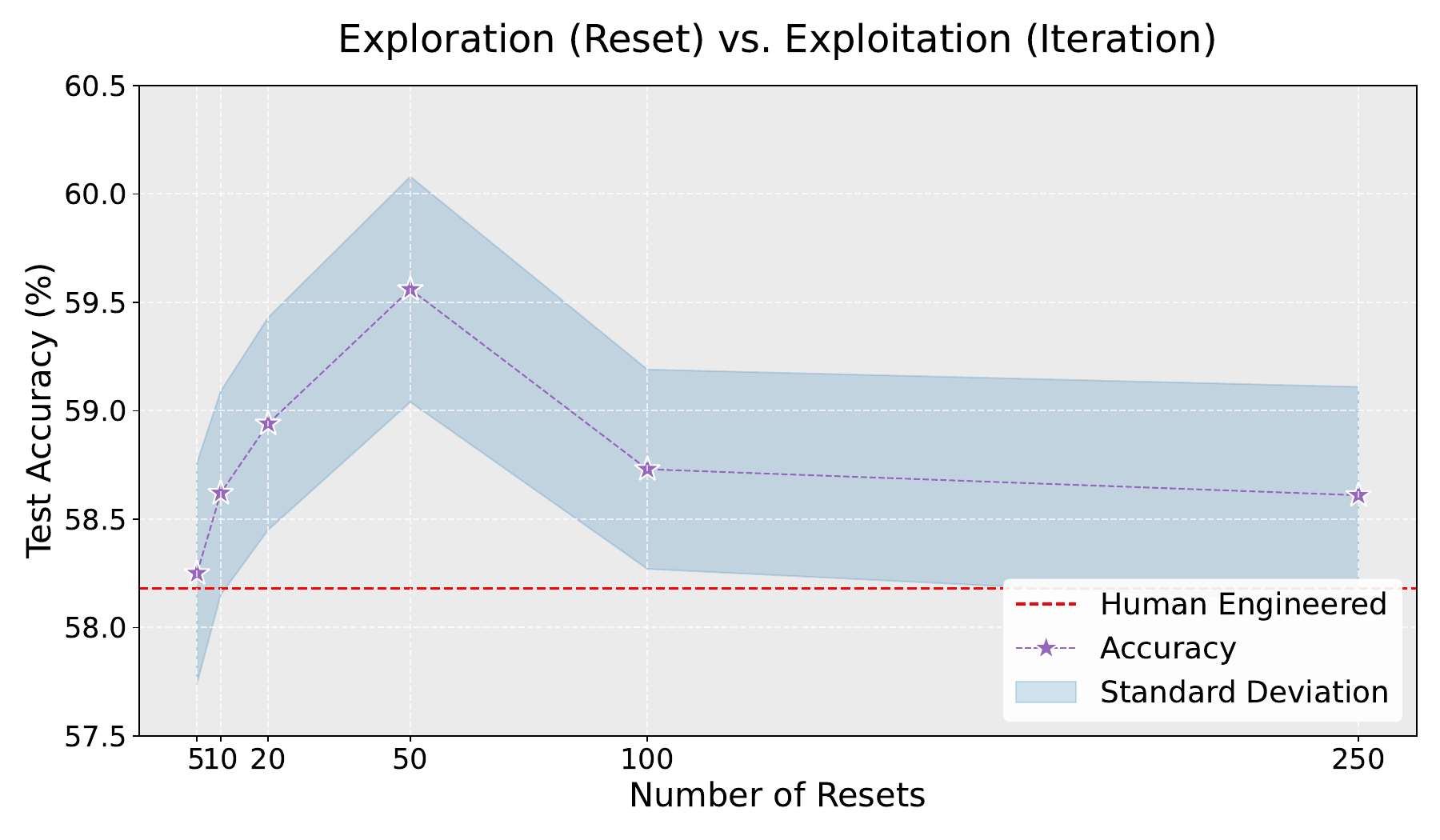}
    \caption{{\bf Balancing exploration and exploitation.} We use a fixed budget of 500 ChatGPT API calls per restart, and ablate the optimal number of resets to use in our algorithm on 1-shot ImageNet. The number of iterations is thus inversely proportional to the number of resets; for example, 10 resets would allow for 50 iterations per reset. We take the average over three runs and also report the standard deviation. We find the optimal balance of exploration and exploitation to be 10 iterations and 50 resets. In contrast, ``pure'' exploration (2 iterations, 250 resets) leads to 0.9\% lower accuracy due to insufficient optimization. On the other hand, when exploitation is overly prioritized (100 iterations, 5 resets), our method gets 1.3\% lower accuracy.}
  \label{fig:exploration_exploitation}
\end{figure}

{\bf Balancing exploration and exploitation can improve the final performance.} Our method extensively leverages the ChatGPT API, necessitating an investigation into strategies for minimizing optimization costs. This leads us to examine the classic dilemma of exploration versus exploitation, a foundational concept in reinforcement learning. Specifically, we use a fixed budget of 500 API calls per restart, and investigate the optimal combination of the number of resets and iterations in \autoref{fig:exploration_exploitation}. For example, we can allocate 50 resets with 10 iterations each to encourage more exploration, or 10 resets with 50 iterations each to foster more exploitation. We find that the optimal balance point is 50 resets of 10 iterations each, and note that no other combination is within 1 standard deviation of the optimal performance. As shown in the performance curve, having too much exploration (250 resets), or too little (5 resets) will result in a roughly $1\%$ decrease in performance. In general, we find it is useful to spend more budget on exploration as ChatGPT can be stuck at local minima within one reset.

\begin{table}[t]
\centering
\scalebox{0.8}{
\begin{tabular}{c|ccc}
\toprule[1.5pt]
\multirow{2}{*}[-1mm]{\textbf{Backbone}} & \multicolumn{3}{c}{\textbf{Method}}
\\ 
\cmidrule(l){2-4}
                   & Our Approach & Hand-Engineered & Linear Probe \\ \midrule
ResNet-50          & 59.6      & 58.2   & 55.9   \\
ResNet-101         & 61.8      & 61.6   & 59.8   \\
ViT-B/32           & 62.6      & 62.0   & 59.6   \\ 
ViT-B/16           & 67.8      & 66.7   & 65.9   \\
\bottomrule[1.5pt]
\end{tabular}
}
\vspace{2mm}
\caption{\textbf{Our method can generalize to various CLIP architectures.} We run our method on 1-shot ImageNet across multiple CLIP backbones, and compare it to the best Human-Engineered prompt and Linear-Probing~\cite{clip} performance. }
\label{tab:backbone}
\end{table}

\begin{table*}[t]
\centering
\renewcommand{\arraystretch}{1.3}
\scalebox{0.7}{
\begin{tabular}{ccccccccccccc}
\toprule[1.5pt]
\multirow{2}{*}[-1mm]{\textbf{Method}} & \multicolumn{11}{c}{\textbf{Dataset}} & \multirow{2}{*}[-1mm]{\bf Avg}
\\ 
\cmidrule(l){2-12}
          &  Caltech & ImageNet & Aircraft & Food & Pets  & Cars  & SUN & UCF & DTD & EuroSAT & Flowers &  \\ \hline

{LAIONCOCO-1M}  &  81.4& 56.2& 17.4& 76.5& 79.6& 51.3& 54.9& 55.8& 43.1& 38.6& 61.3& 56.0 \\ 

{Iterative APE}  &  88.3& 58.1& 17.0& 77.3& 85.1& 54.8& 58.6& 57.4& 41.2& \underline{46.7}& 65.3&  59.0\\ 

{Ours (P only)}  & {\underline{89.0}}     & \underline{59.4}    & {\underline{17.9}}    & {\underline{77.8}}   & {\underline{85.7}}  & {\underline{55.7}}  & {\underline{60.4}}  & {\underline{58.7}}  & \underline{43.6} & \underline{46.7}   & \underline{66.6}  & {\underline{60.1}} \\ 

{Ours (P+N)}  & {{\bf 89.1}}  & {{\bf 59.6}}  & {\bf 18.1} & {{\bf 78.3}}  & {{\bf 88.1}}  & {{\bf 56.2}}  & {{\bf 61.0}}  & \bf{60.2}  & {{\bf 44.8}}  & {\bf 49.0}  & {\bf {67.2}} & {{\bf 61.1}} \\ 

\bottomrule[1.5pt]
\end{tabular}
}
\vspace{2mm}
\caption{\textbf{Comparing our method with our own version of iterative APE~\cite{ape}.} Optimized using 1-shot training sets, we find that both iterative APE and our methods can effectively improve upon the initial sampled prompts. However, our method achieves better performance within the same computational budget, presumably because we provide explicit textual feedback to ChatGPT, leading to faster convergence.} 
\label{tab:ape}
\end{table*}

\begin{table*}[h]
\centering
\renewcommand{\arraystretch}{1.3}
\scalebox{0.7}{
\begin{tabular}{cccccccccccccc}
\toprule[1.5pt]
\multirow{2}{*}[-1mm]{\textbf{Shot}} & \multirow{2}{*}[-1mm]{{\bf Method}} & \multicolumn{11}{c}{\textbf{Dataset}} & \multirow{2}{*}[-1mm]{{\bf Avg}}
\\ 
\cmidrule(l){3-13}
          &  & Caltech & ImageNet & Aircraft & Food & Pets  & Cars  & SUN & UCF & DTD & EuroSAT & Flowers &  \\ \hline

\multirow{2}{*}{1 shot}                        & Ours (P only)  & 89.0 & 59.4 & 17.9 & 77.8 & 87.8 & 55.7 & 60.4 & 58.7&43.6 &46.7  &66.6 & 60.1  \\ 
                        & Ours (P+N) & 89.1& 59.6 & 18.1 & 78.3 & 88.1 & 56.2 & 61.0 & 60.2 & 44.8 & 49.0 &67.2 &  61.1  \\ \hline
\multirow{2}{*}{16 shot}                        & Ours (P only)  & 89.3 & 59.6 & 17.7 & 77.9 & 86.6 & 56.2 & 61.0 & 60.2& 44.0& 49.0 & 66.0&  60.6 \\ 
                        & Ours (P+N) & 89.5 & 59.9 & 18.1 & 78.3 & 88.3 & 56.8 & 60.8 & 60.5&44.9 &51.4  &67.4 &61.4    \\ 

\bottomrule[1.5pt]
\end{tabular}
}
\vspace{2mm}
\caption{\textbf{Higher-shot performance.} We report the 16-shot performance of our method in this table. It is important to note that as the number of shots increases, the role of the natural language prompt diminishes because it will be more effective to tune the visual representations (which requires white-box access to VLMs). }

\label{tab:extra_shots}
\end{table*}

\begin{table*}[h!]
\centering
\scalebox{0.75}{
\begin{tabular}{ccccccccccccc}
\toprule[1.5pt]
\multirow{2}{*}[-1mm]{\textbf{GPT version}} & \multicolumn{11}{c}{\textbf{Dataset}} & \multirow{2}{*}[-1mm]{\textbf{Avg}}
\\ 
\cmidrule(l){2-12}
          & Caltech & ImageNet & Aircraft & Food & Pets  & Cars  & SUN & UCF & DTD & EuroSAT & Flowers &  \\ \hline
gpt-turbo-3.5-0301 & 89.1      & 59.6    & 18.1    & 78.3   & 88.1  & 56.2  & 61.0  & 60.2  & 44.8 & 49.0   & 67.2 & 61.1  \\ 
gpt-4-0314 & 89.1      & 59.6    & 17.9    & 78.5   & 87.7  & 56.2  & 60.3  & 59.9  & 45.0 & 48.0   & 67.6  & 60.9 \\
\bottomrule[1.5pt]
\end{tabular}
}
\caption{\textbf{ChatGPT versus GPT4.} Our approach is equally effective using other versions of ChatGPT.}
\label{tab:gpt4}
\end{table*}

\textbf{Reimplementing (iterative) APE for VLM optimization.} We attempt to implement our own version of iterative APE using the given prompts in~\cite{ape} while making minimal changes such that it fits in our automatic prompt-searching system. For a fair comparison, we reuse exactly the same initial sampled prompts from LAIONCOCO-1M for iterative APE because their ``instruction-induction'' paradigm cannot be applied to VLM optimization settings. The results are shown in \autoref{tab:ape}. We find that iterative APE shows inferior performance to our method, presumably because we leverage more textual feedback for more efficient search. The exact prompt we use is shown below: 

\begin{mdframed}
\texttt{Hi ChatGPT, generate a single variation of the following template while keeping the semantic meaning: \\
- \textcolor{blue}{template} \\
Here is my requirement:\\
- Please return a single template starting with '-'}
\end{mdframed}

\textbf{Comparison of CLIP backbones.} To verify that our method scales properly to other CLIP backbones, we test our method on ImageNet using four different CLIP backbones: ResNet-50, ResNet-101, ViT-B/32, and ViT-B/16. We compare our method with hand-engineered prompts, and a linear probe (linear classification on the visual embeddings). \autoref{tab:backbone} shows the results of the experiment, where we see that our method outperforms the baselines consistently. Thus, our method scales appropriately with larger and more powerful models.

\textbf{Results on higher shots.} We additionally test the generalization ability of our method given more data (16 shots), with results shown in \autoref{tab:extra_shots}. We observe that our method gains small but incremental improvements given more data, and using both top-k and bottom-k prompts (P+N) consistently outperforms top-2k prompts (P only).

\textbf{Results using GPT4.} We run our approach using the same hyperparameters and initial prompts using GPT4 in \autoref{tab:gpt4}. It shows that our approach is equally effective using other versions of ChatGPT, but interestingly, there is no performance benefit of using GPT4. This may be because our hyperparameters were optimized on ChatGPT, and are suboptimal for GPT4.

{\bf Cost analysis. }We use GPT3.5 which costs $\$0.0015$ per 1000 tokens. In our default setup, we use an average of 500 tokens per API call. We use a total of 500 API calls (50 resets and 10 iterations) for a total of 250,000 tokens per restart, and thus each run costs around 50 cents. Since we use 20 restarts per dataset, the total cost over the suite of 11 datasets is around $\$100$ for each of the three folds.

\clearpage
\section{T2I Experimental Details}
\label{sec:t2i_details}

In this section, we include implementation details and more qualitative results for T2I generation experiments.

{\bf Image generation using DALL-E 3.} We use the below template to generate images without changing the prompts.

\begin{mdframed}
    {\tt Create this exact image without any changes to the prompt: \textcolor{blue}{\{prompt\}}. }
\end{mdframed}

{\bf T2I generation (\autoref{fig:dalle3}).} We use DALL-E 3 to expand the query text to a longer prompt for the first image. Next, we send generated image, query text, and current prompt to GPT4-V for prompt optimization.
\begin{mdframed}
    {\tt
    \textcolor{red}{Prompt for DALLE-3 (first round):} Create an image that shows \textcolor{blue}{\{query text\}}. \\ \\ \textcolor{red}{Prompt for GPT-4V:} 
    \texttt{Do you think this image \textcolor{blue}{\{generated image\}} correctly depicts  \textcolor{blue}{\{query text\}}? If not, briefly explain why and suggest modifications. Then, help me adjust the prompt to make it correct: \textcolor{blue}{\{prompt\}}. Please provide a response in a JSON file format containing: (1) "feedback" summarizing the key points, and (2) "new\_prompt" with the revised text.}
    }
\end{mdframed}

{\bf Prompt inversion (\autoref{fig:dalle3_inversion}).} We use GPT4-V to generate the initial prompt given the query image. Next, we send query image, generated image, and current prompt to GPT4-V for prompt optimization.

\begin{mdframed}
    {\tt \textcolor{red}{Prompt for GPT-4V (first round):} Generate a detailed text prompt to recreate the attached image \textcolor{blue}{\{query image\}} using an image generator. \\ \\ \textcolor{red}{Prompt for GPT4-V:} Compare the original image \textcolor{blue}{\{query image\}} and generated image \textcolor{blue}{\{generated image\}}, analyze their differences, and then propose changes to update the original prompt in-place: \textcolor{blue}{\{prompt\}}. Please provide a response in a JSON file format containing: (1) "feedback" summarizing the key points, and (2) "new\_prompt" with the revised text.}
\end{mdframed}

{\bf Failure cases.} We show some failure cases of our method in \autoref{tab:failure_t2i} and \autoref{tab:failure_inversion}. We note that these queries are especially challenging even for state-of-the-art VLMs because they require complex reasoning abilities. We expect better performance of our framework using stronger generative models in the future.

{\bf Prompt inversion on natural images.} In addition to sampling queries from DiffusionDB~\cite{wang2022diffusiondb}, we also attempt at prompt inversion with natural images, as shown in \autoref{tab:real_img_inversion}. %

{\bf Human studies.} We hire two graphical designers who have one year of experience using AI content creation tools such as Stable Diffusion and Midjourney to manually design the prompts for DALL-E 3. We also hire two volunteers to assign a Likert scale score between the generated image and user query according to \autoref{tab:likert}.

\begin{table*}[h]
\centering
\scalebox{0.72}{
\begin{NiceTabular}{M{0.2\linewidth} M{0.16\linewidth} M{0.16\linewidth} M{0.62\linewidth}}
\CodeBefore
    \Body
\toprule[1.5pt]
\textbf{User Query} & \textbf{Init. Image} & \textbf{Final Image} & \textbf{Final Prompt} \\ \midrule
\Block{1-4}{{\bf Text-to-image generation}} \\
The unmasked wrestler hits the masked wrestler. & \includegraphics[width=30mm,height=30mm]{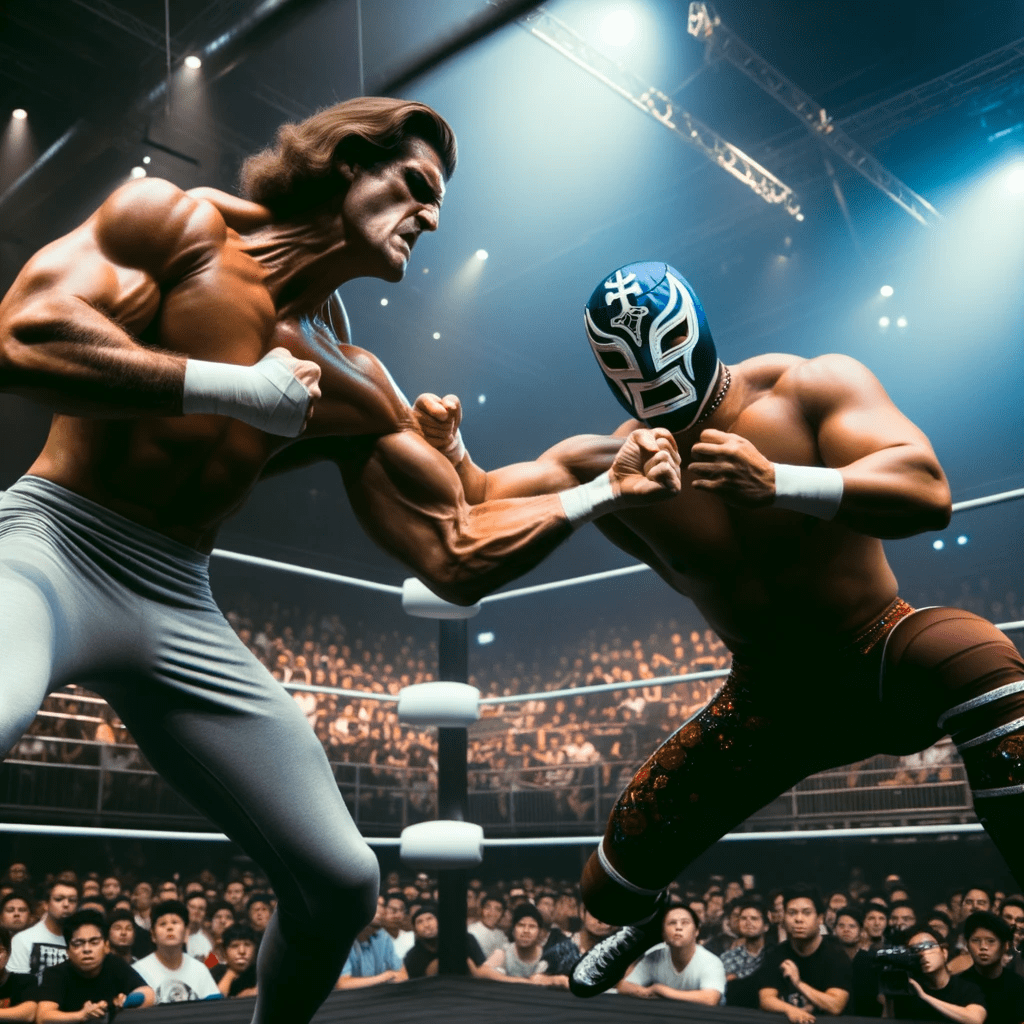} & \includegraphics[width=30mm,height=30mm]{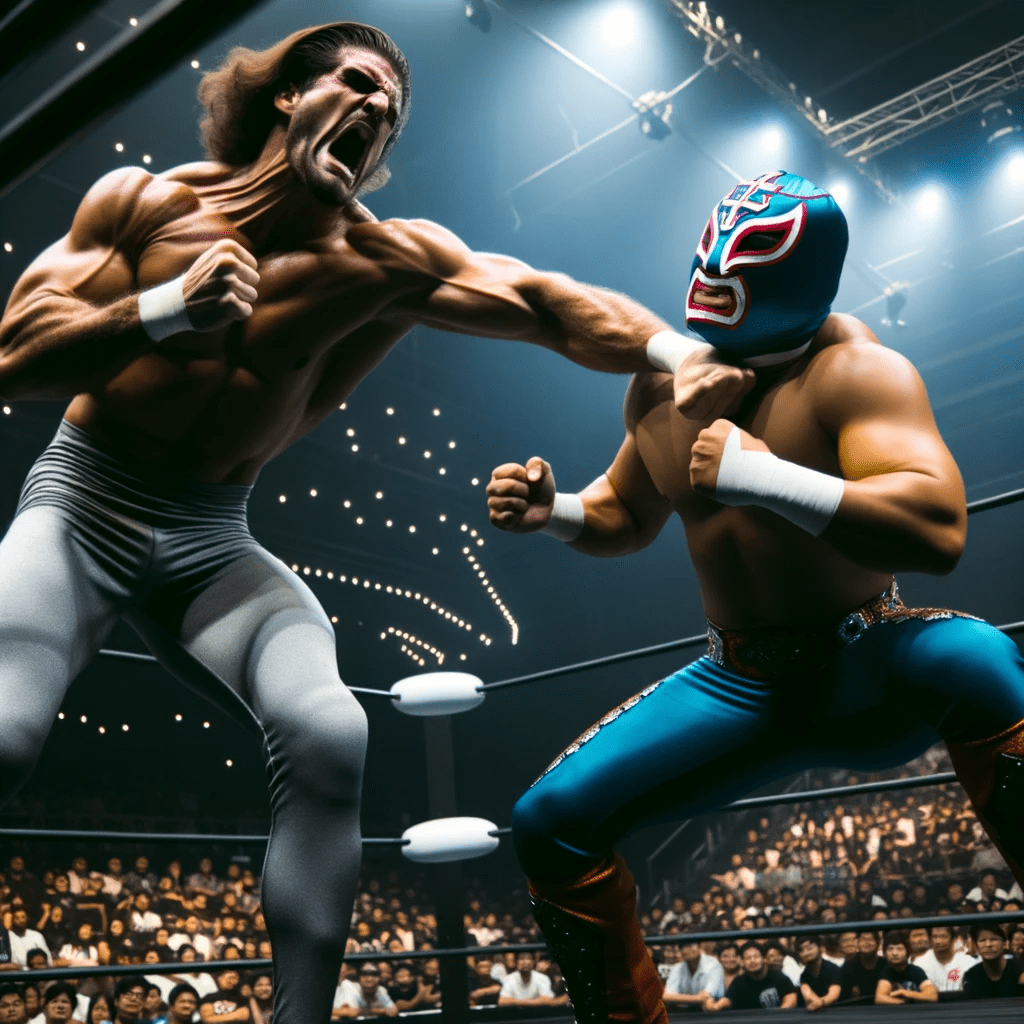} & Photo of a wrestling ring where an unmasked male wrestler with a muscular physique is in the midst of delivering a powerful blow to a masked male wrestler donning a lucha libre style mask. The spectators in the background are on the edge of their seats, watching the action closely.  \\ \midrule

The person with earrings pays the person without earrings & 
\includegraphics[width=30mm,height=30mm]{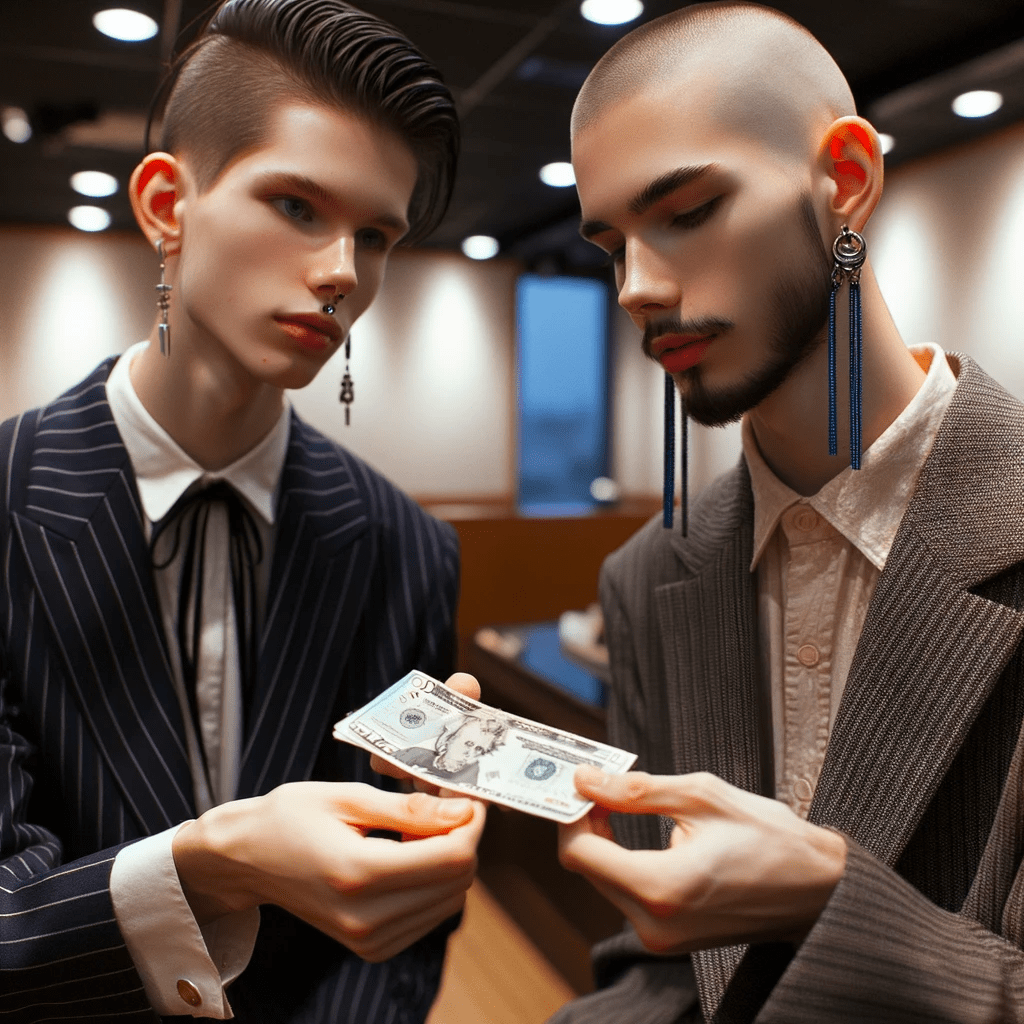} & \includegraphics[width=30mm,height=30mm]{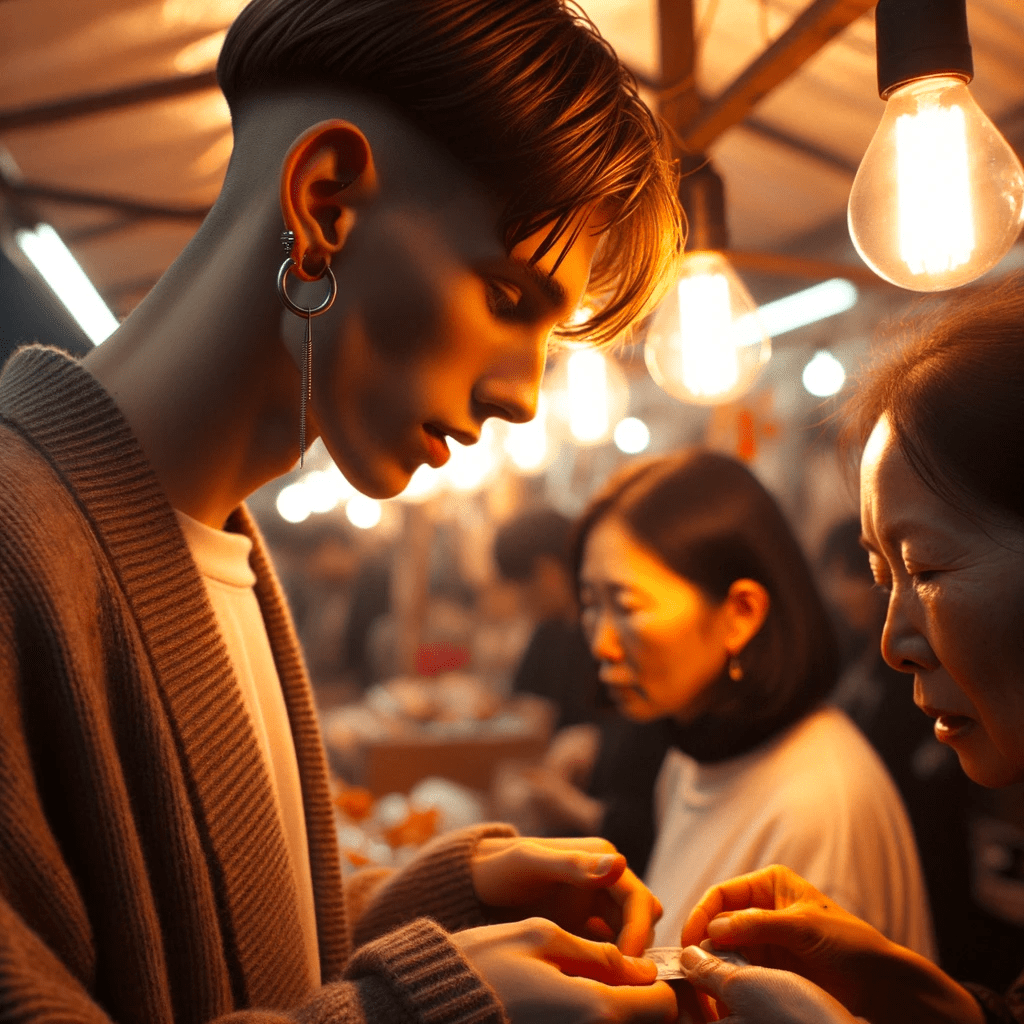} & Photo of a person with a short haircut and noticeable earrings in the process of paying a long-haired vendor without earrings at a market stall, with warm lighting.  \\ \midrule

A bird eats a snake & 
\includegraphics[width=30mm,height=30mm]{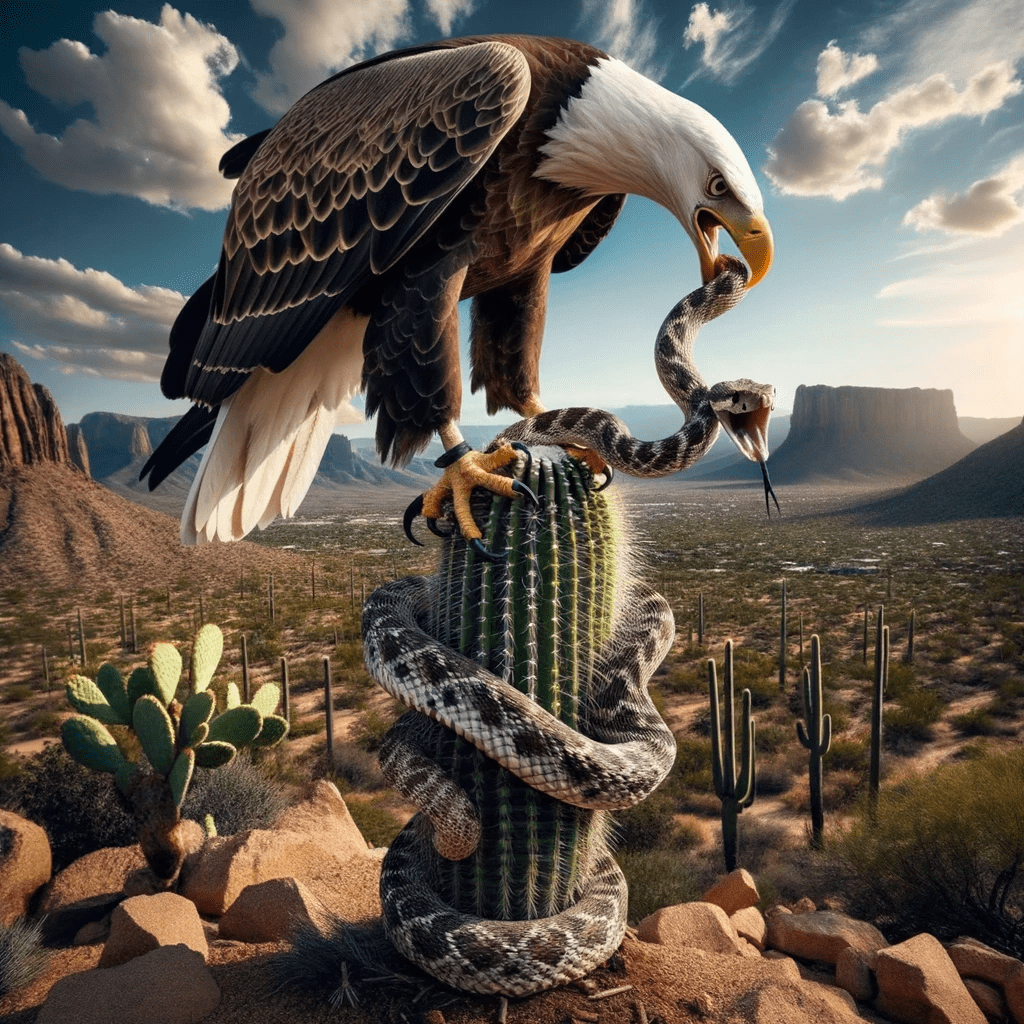} &  \includegraphics[width=30mm,height=30mm]{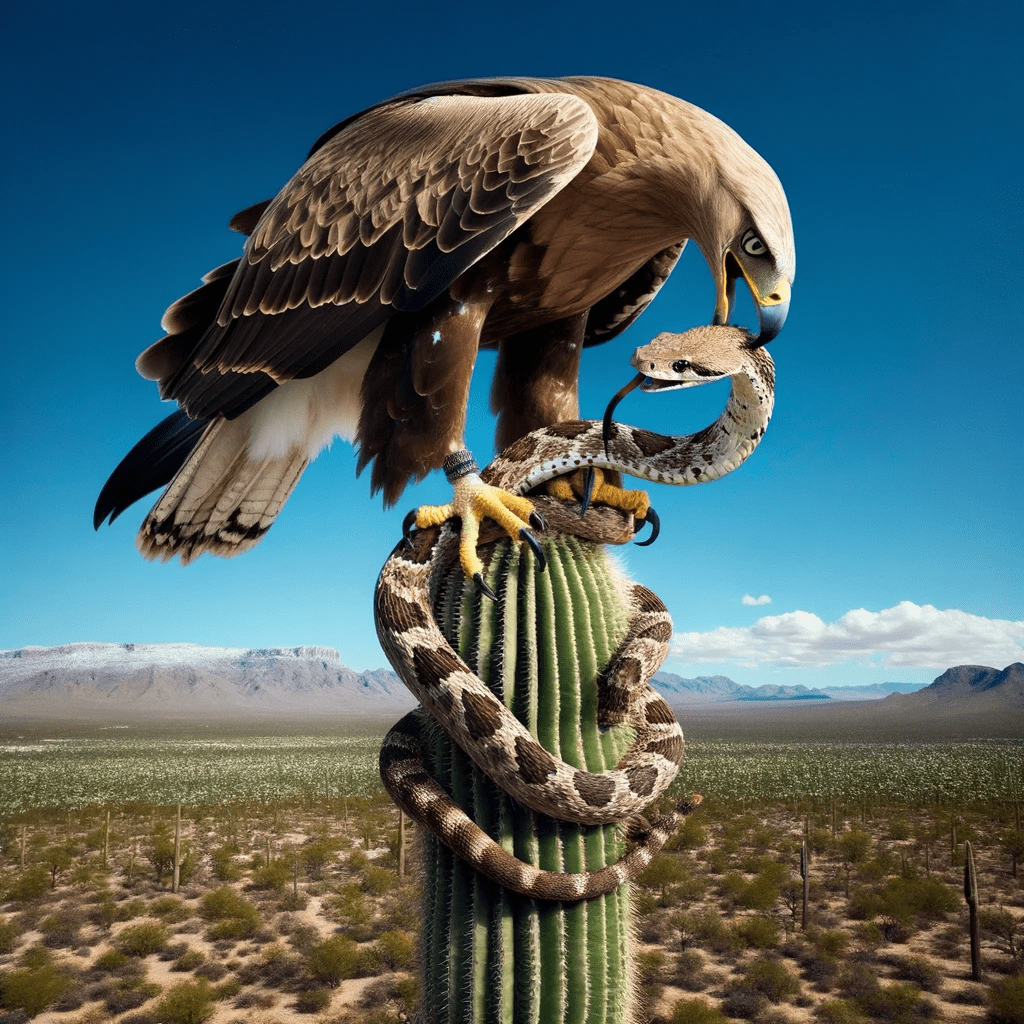} & Photo of a vast desert landscape under a clear blue sky. In the foreground, a large, powerful eagle with brown feathers and piercing eyes is perched confidently on a tall, green cactus. The eagle tightly clenches a rattlesnake in its strong talons. The snake's rattle is visible, and it appears to be struggling. The eagle's beak is wide open, showing its sharp beak, indicating it's about to consume the snake.  \\ \midrule

A shorter person is covering the eyes of a taller person. & \includegraphics[width=30mm,height=30mm]{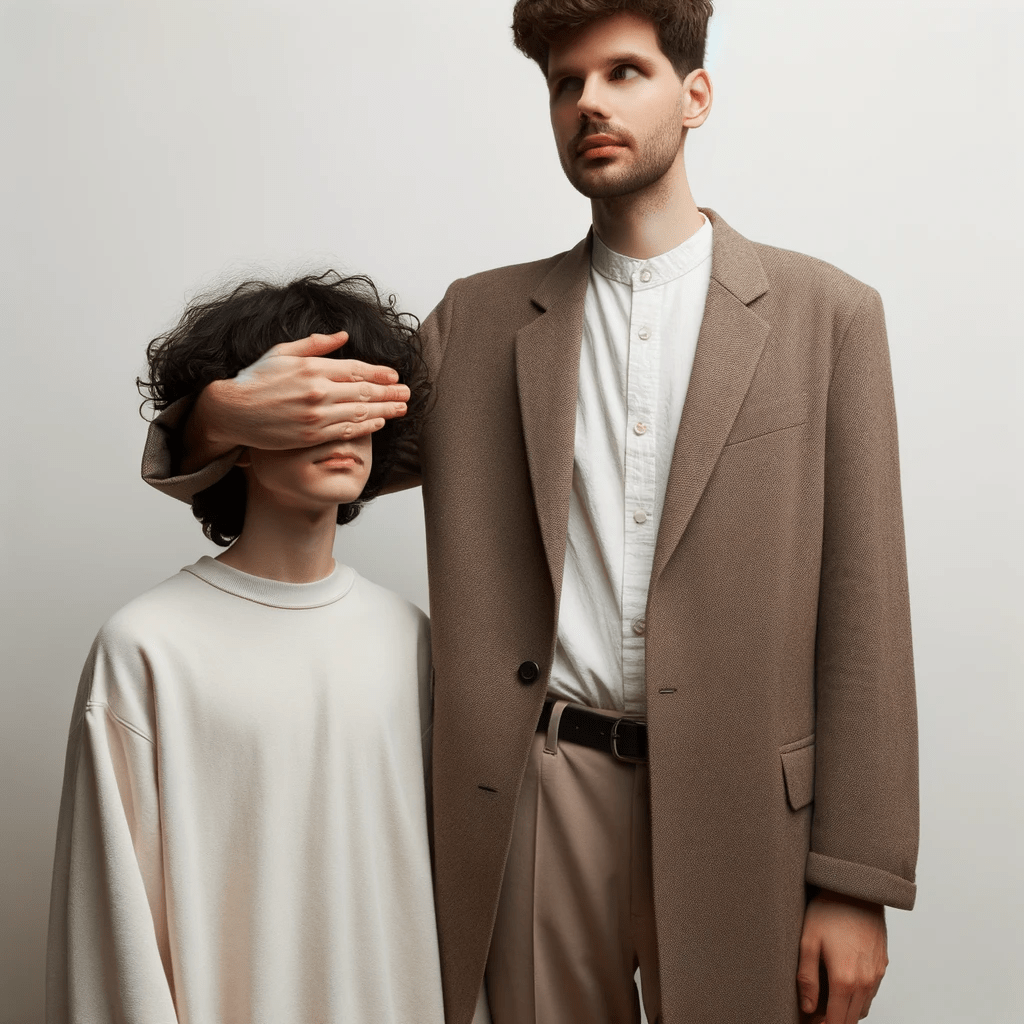}  & \includegraphics[width=30mm,height=30mm]{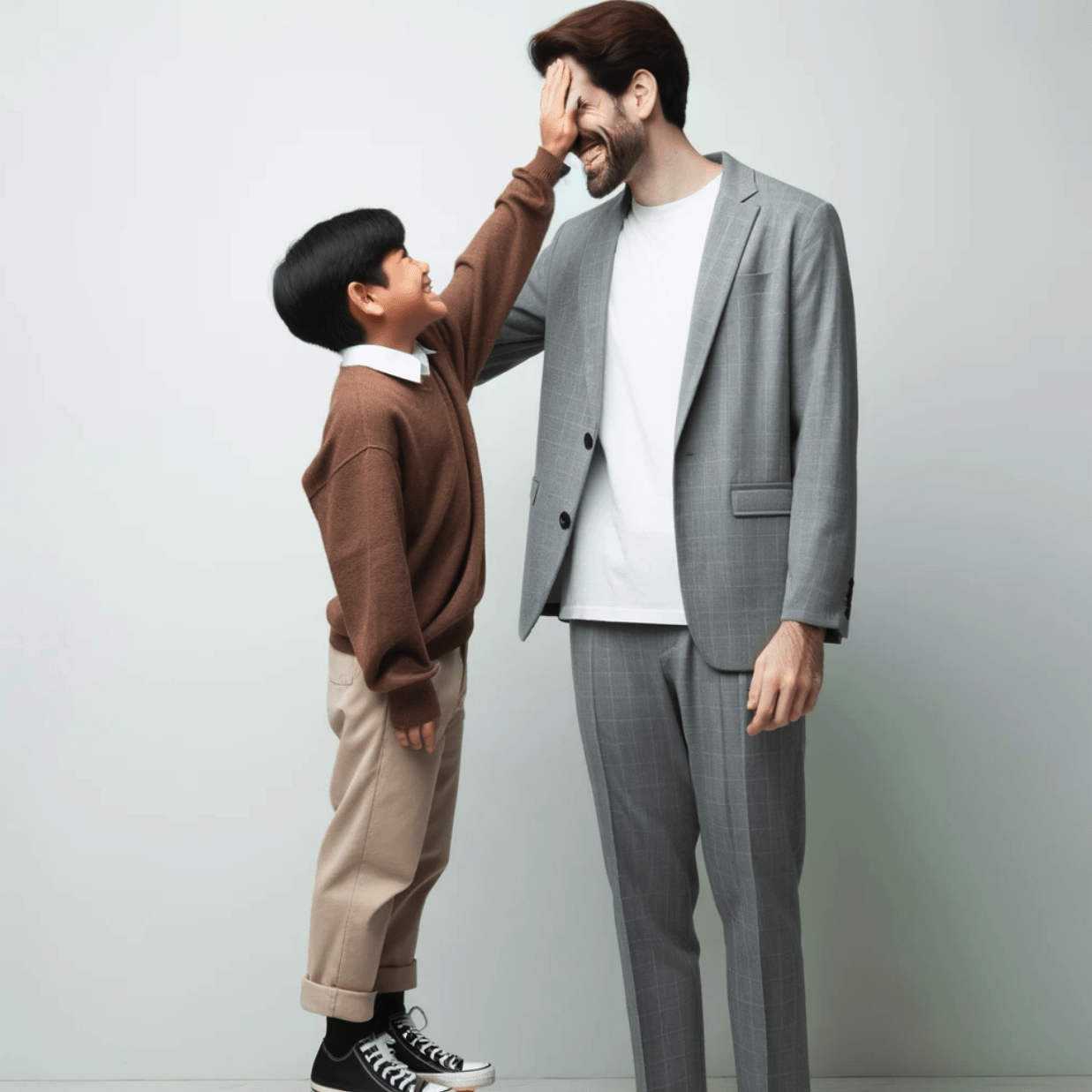} & A shorter individual reaching up to cover the eyes of a taller person standing in front of them. The shorter person is on their tiptoes, trying to reach the taller person's eyes. Both of them are smiling and seem to be enjoying the moment. The backdrop is simple and unobtrusive to maintain focus on the subjects.\\ \midrule

There is less milk than orange juice. & \includegraphics[width=30mm,height=30mm]{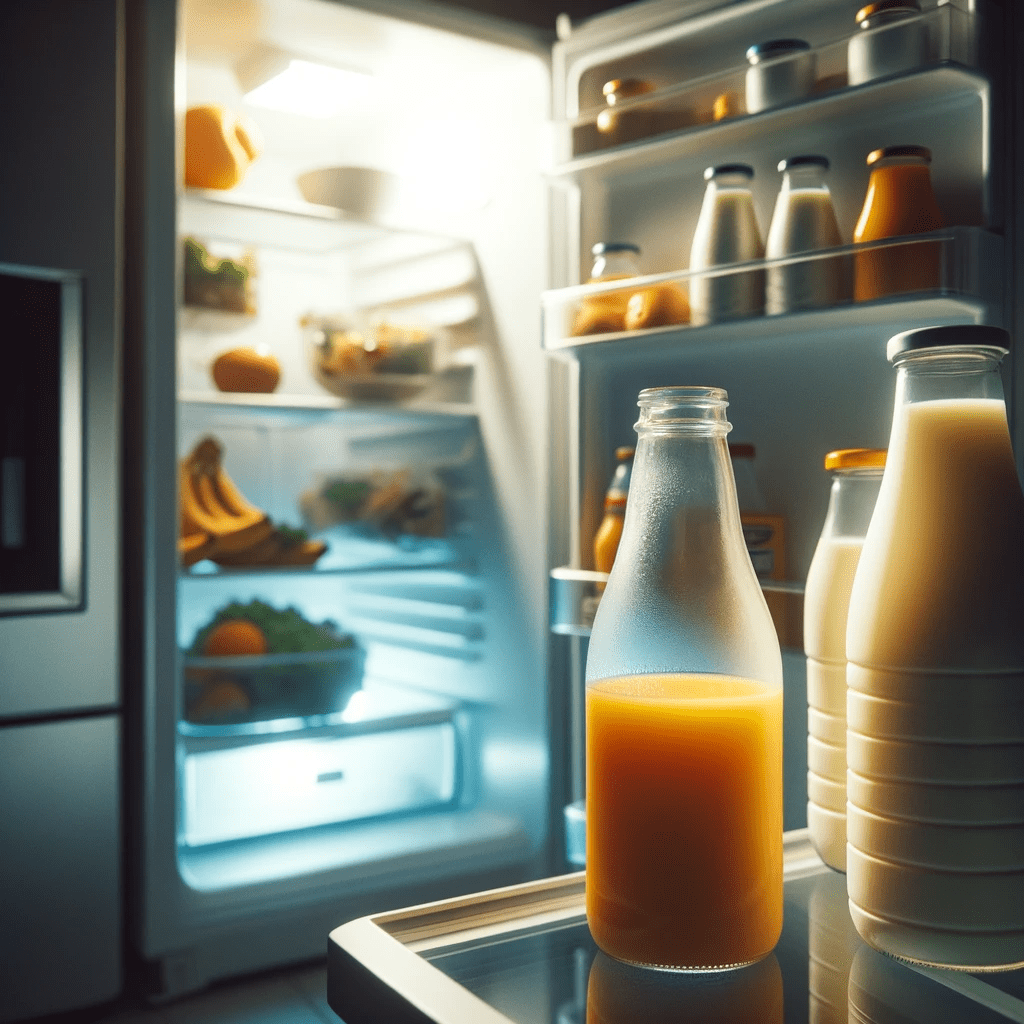} & \includegraphics[width=30mm,height=30mm]{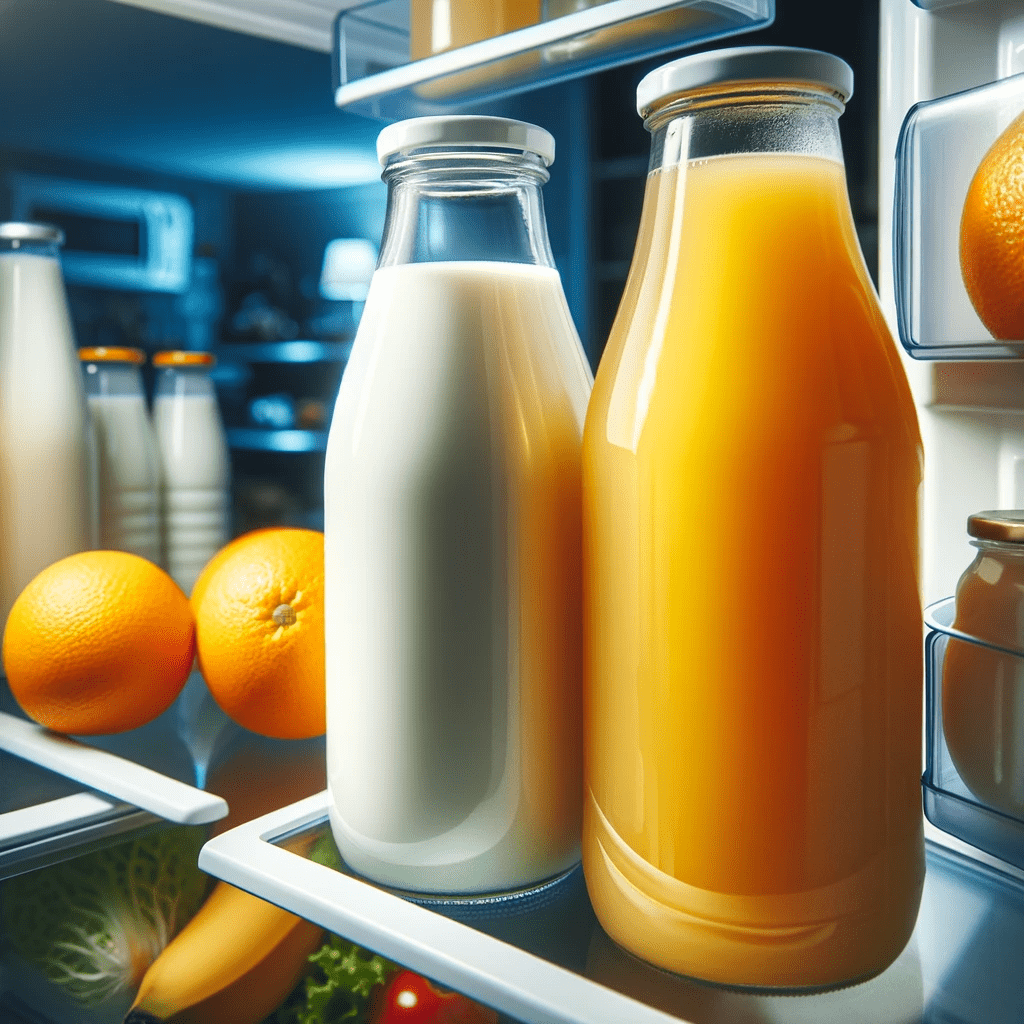} & A kitchen scene showing a fridge door open. Inside, there's a clear glass bottle of milk and a larger bottle of orange juice. Both the milk and orange juice bottles are almost full, with a nearly equal amount visible at the top. The fridge shelves are well-lit, and the rest of the fridge contains a variety of food items including fruits and vegetables. The focus is on the milk and orange juice bottles that are prominently displayed on the shelves.
\\ \midrule

Getting a horse wet. & \includegraphics[width=30mm,height=30mm]{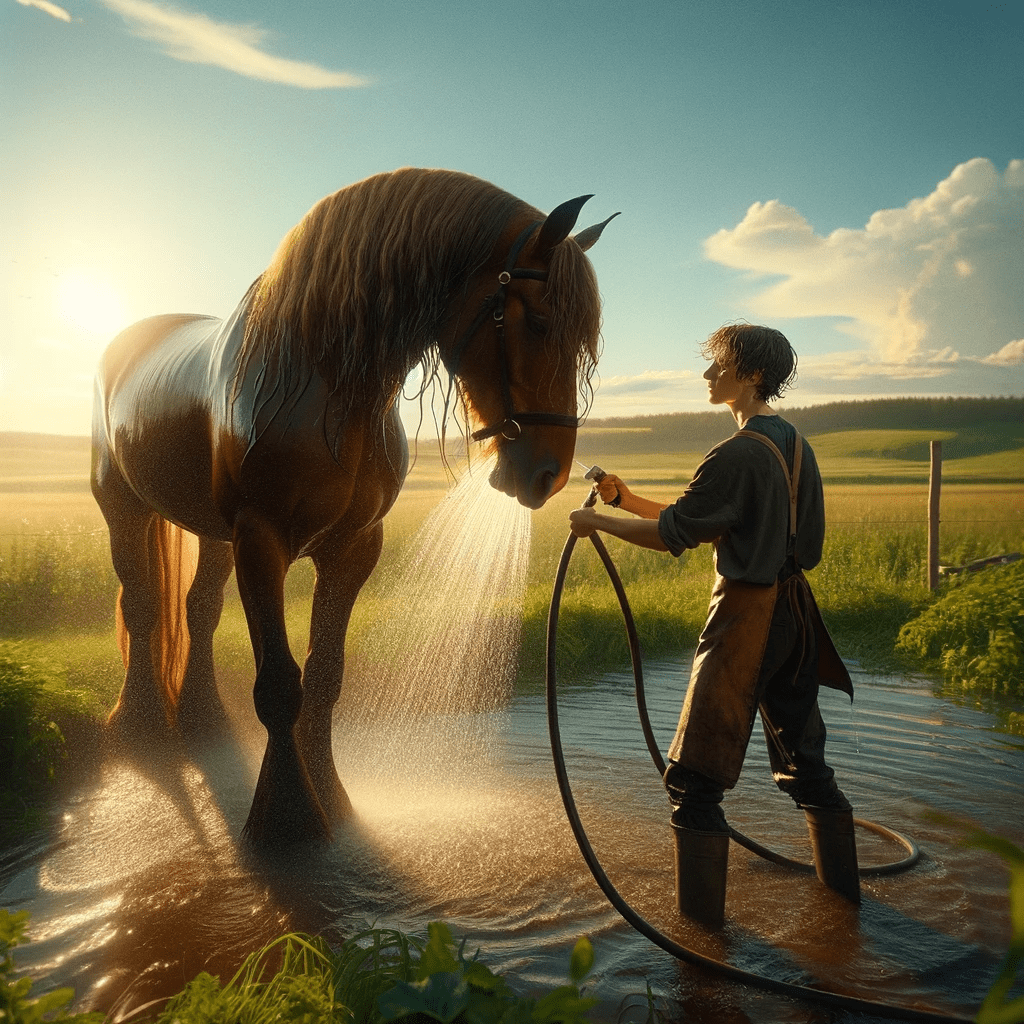}  & \includegraphics[width=30mm,height=30mm]{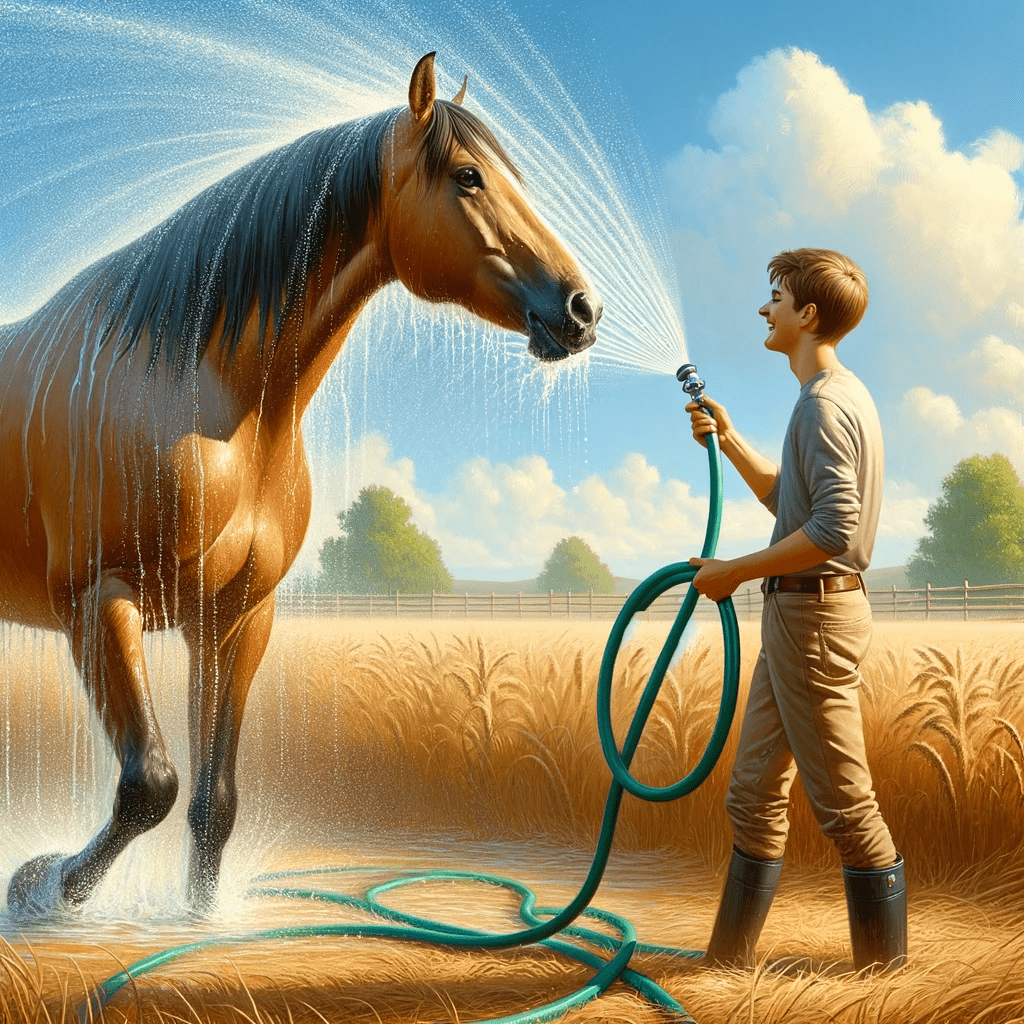} & A scene depicting a person using a hose to gently spray water on a horse in an open field. The horse appears calm and enjoys the water, with droplets of water glistening on its coat. The person is smiling, dressed in casual outdoor attire. The background features a clear blue sky and a few trees, creating a serene and peaceful setting. The horse is a beautiful chestnut color, and the person is Caucasian with short brown hair.
\\ \midrule

Some are parking in a train. & \includegraphics[width=30mm,height=30mm]{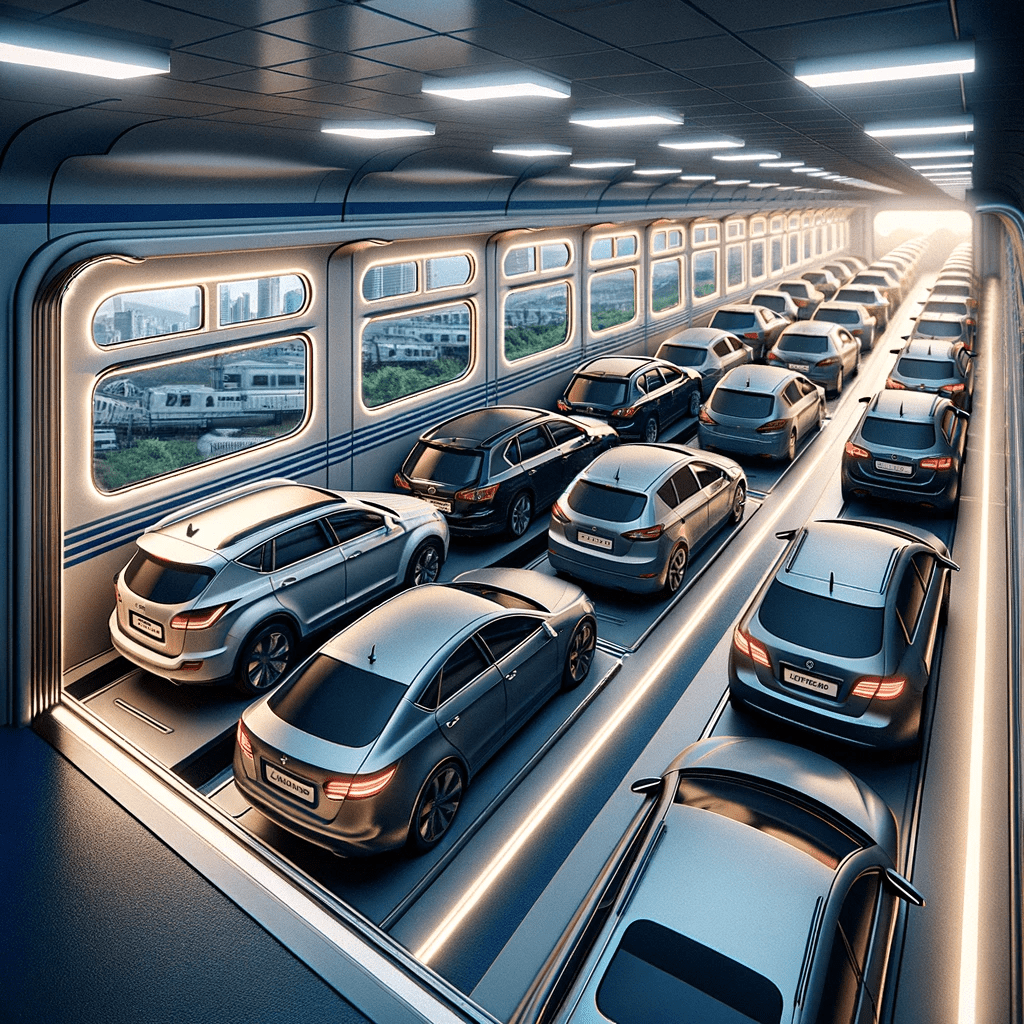}  & \includegraphics[width=30mm,height=30mm]{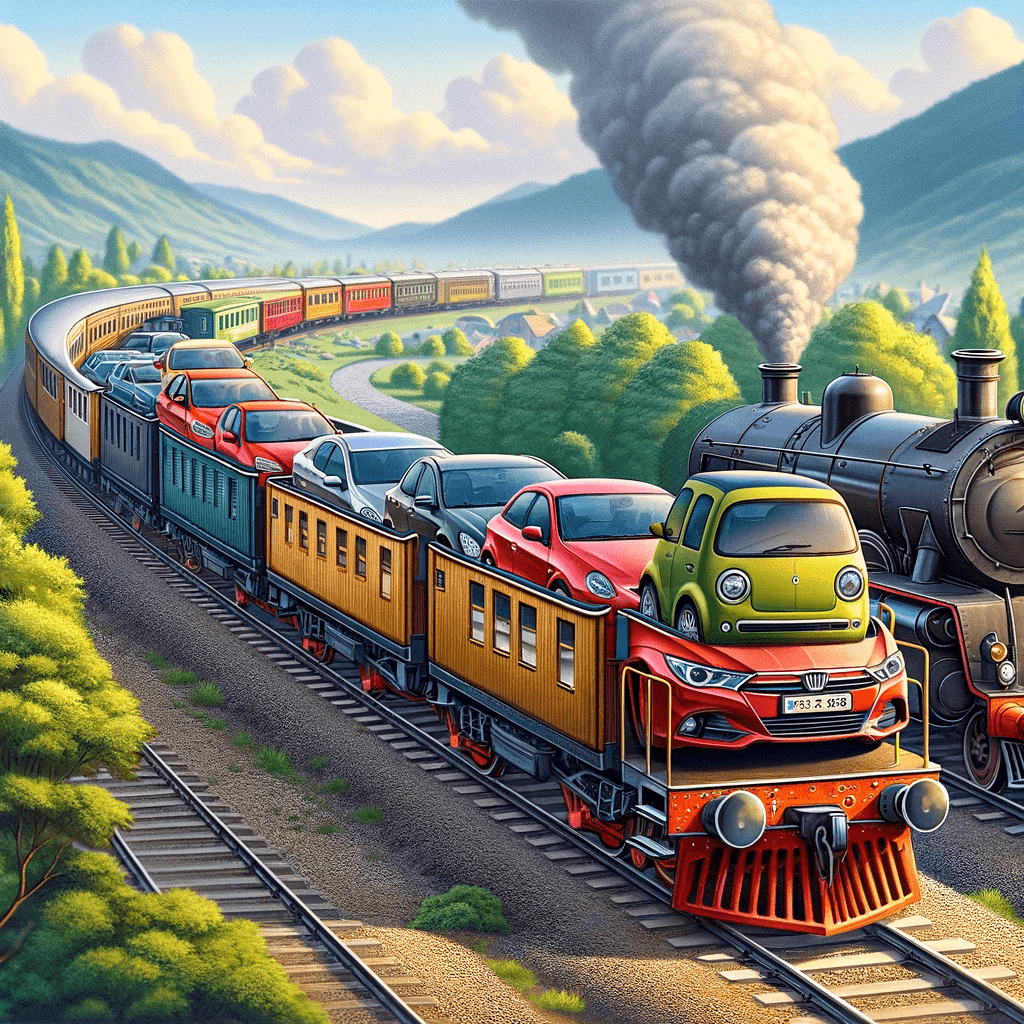} & A whimsical scene depicting a train where some of the carriages are designed as parking spaces, with various types of cars parked inside them. The train is moving through a picturesque landscape, with mountains in the background and a clear blue sky overhead. The cars in the train's parking carriages include a red sports car, a green SUV, and a yellow compact car. The train itself is a classic steam locomotive with a touch of modern design, emitting a puff of steam as it chugs along the tracks.
\\ \midrule

The white wall will soon be painted blue. & \includegraphics[width=30mm,height=30mm]{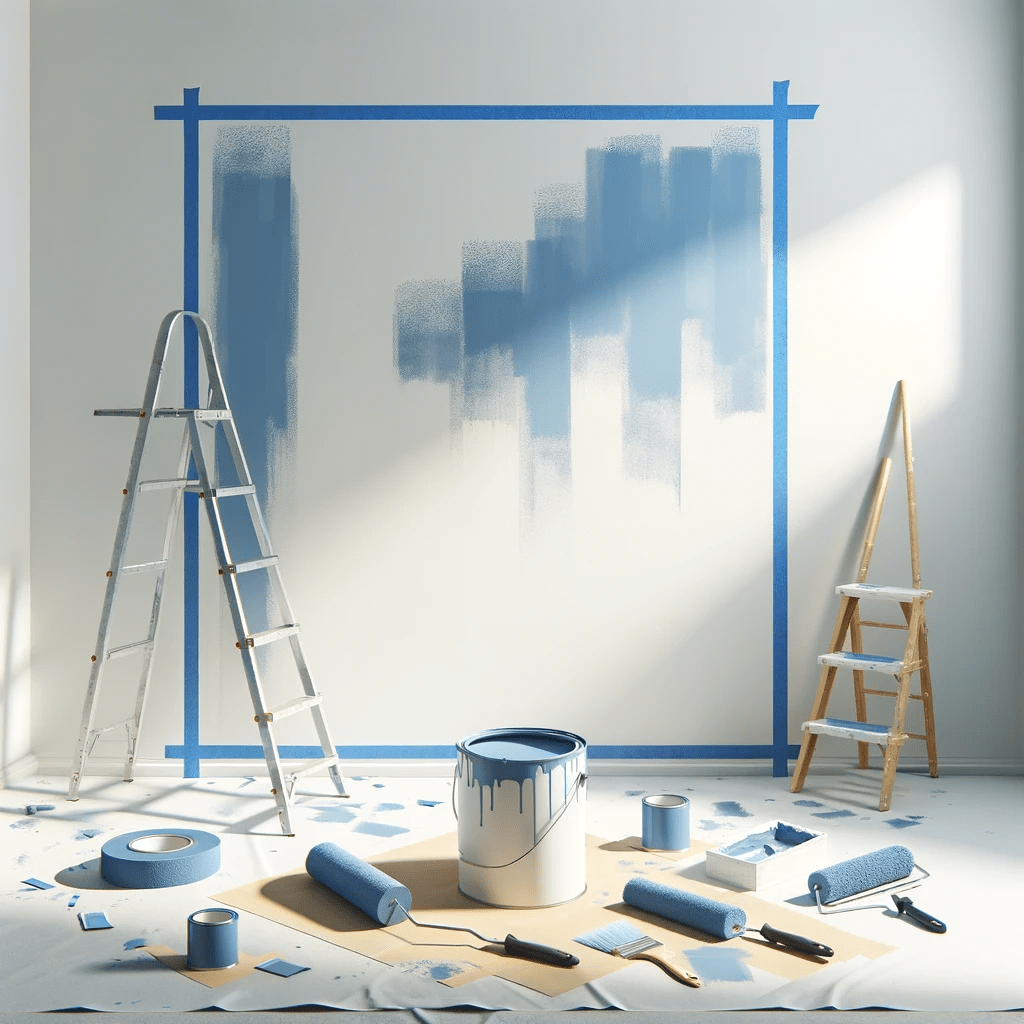}  & \includegraphics[width=30mm,height=30mm]{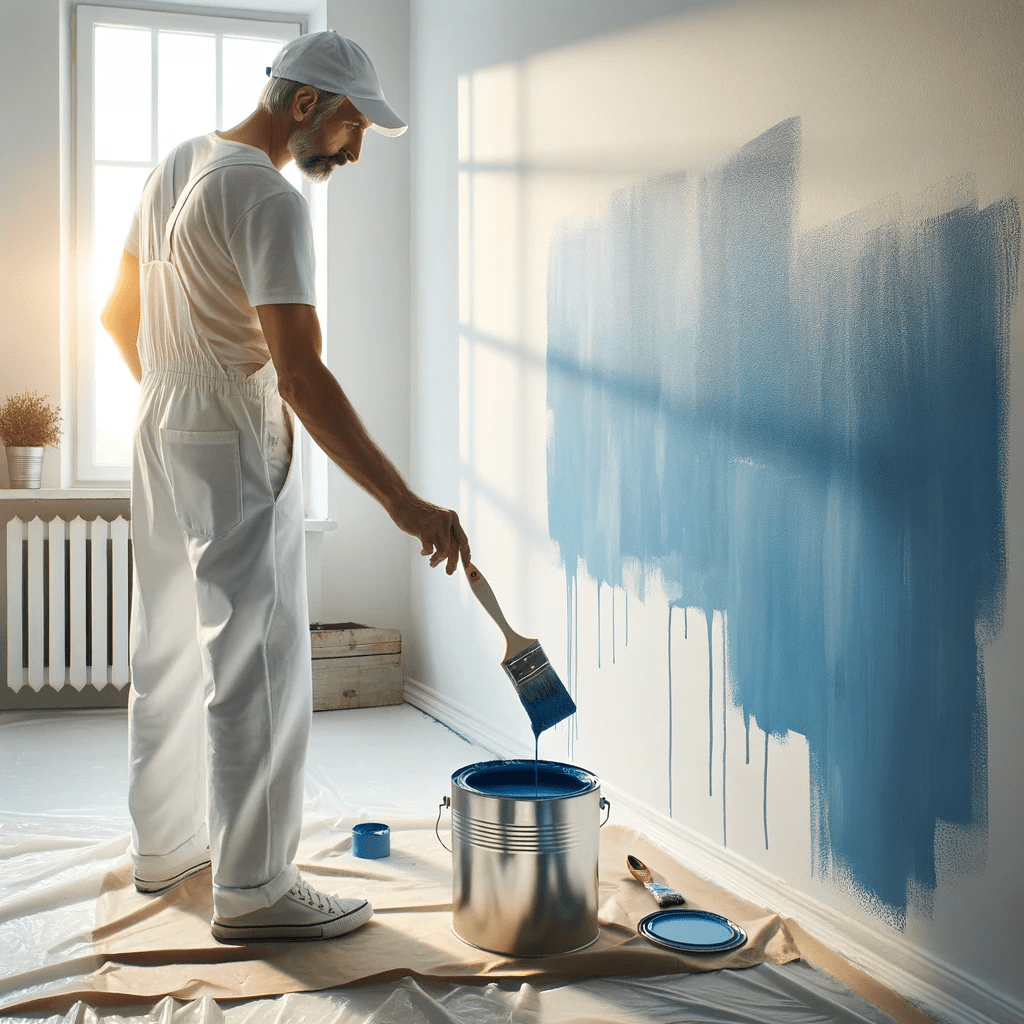} & A white wall in a room, with a paint can and a paintbrush beside it. The can is open and filled with blue paint, ready for use. A painter, a middle-aged Caucasian man wearing a white painter's outfit and a cap, is dipping the brush into the blue paint, preparing to start painting the wall. The room has a window with daylight coming through, casting a bright ambiance over the scene. 
\\ 

\bottomrule[1.5pt]
\end{NiceTabular}
}
\caption{\textbf{More results of T2I optimization.}}
\label{tab:more_t2i}
\end{table*}

\begin{table*}[h]
\centering
\scalebox{0.72}{
\begin{NiceTabular}{M{0.16\linewidth} M{0.16\linewidth} M{0.16\linewidth} M{0.7\linewidth}}
\CodeBefore
    \Body
\toprule[1.5pt]
\textbf{User Query} & \textbf{Init. Image} & \textbf{Final Image} & \textbf{Final Prompt} \\ \midrule
\Block{1-4}{{\bf Prompt inversion}} \\
\includegraphics[width=30mm,height=30mm]{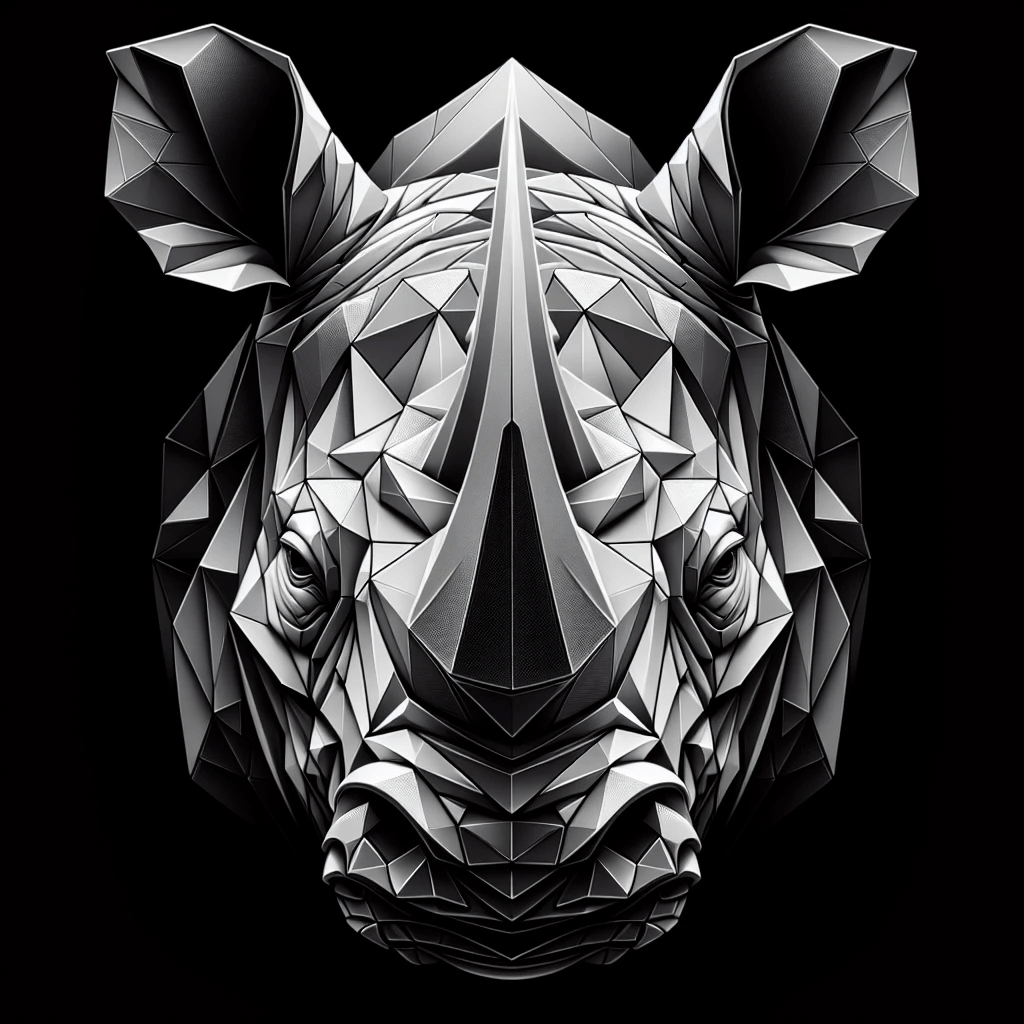} & \includegraphics[width=30mm,height=30mm]{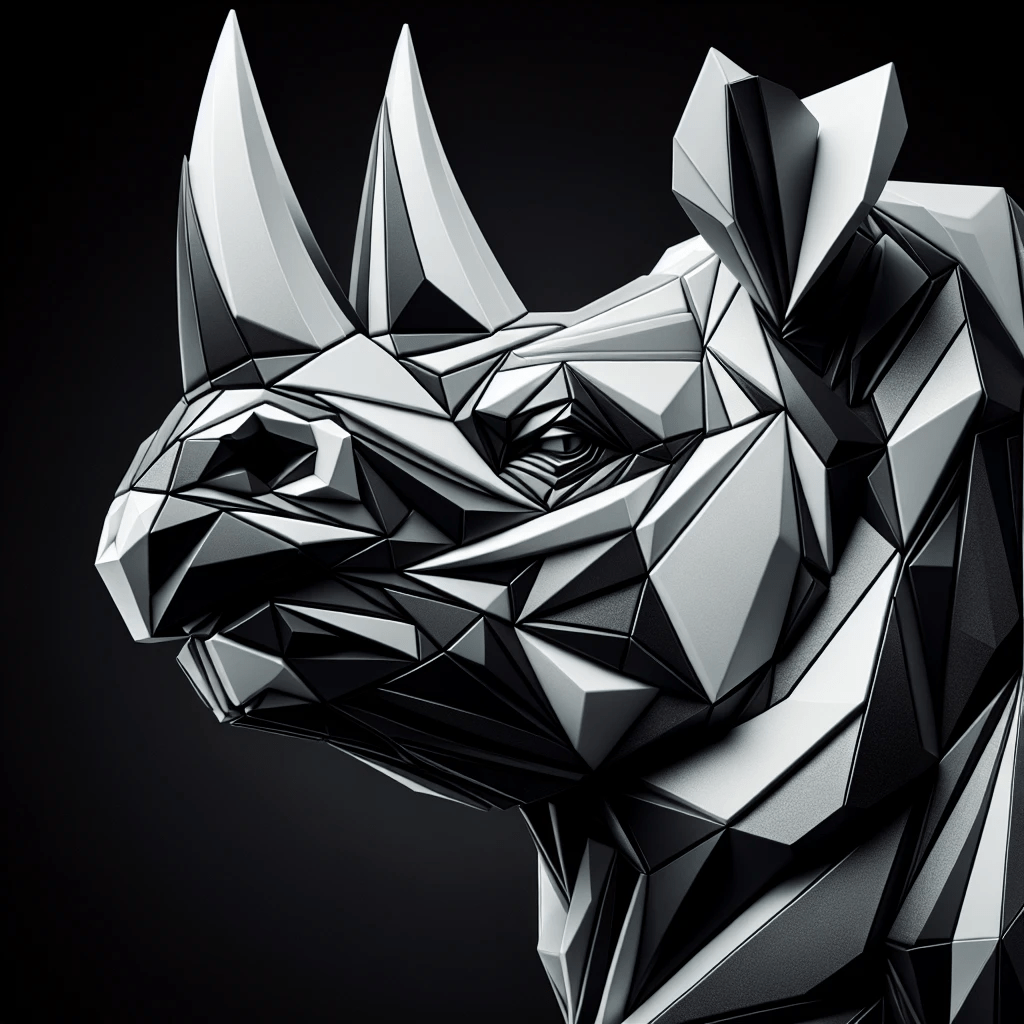} & \includegraphics[width=30mm,height=30mm]{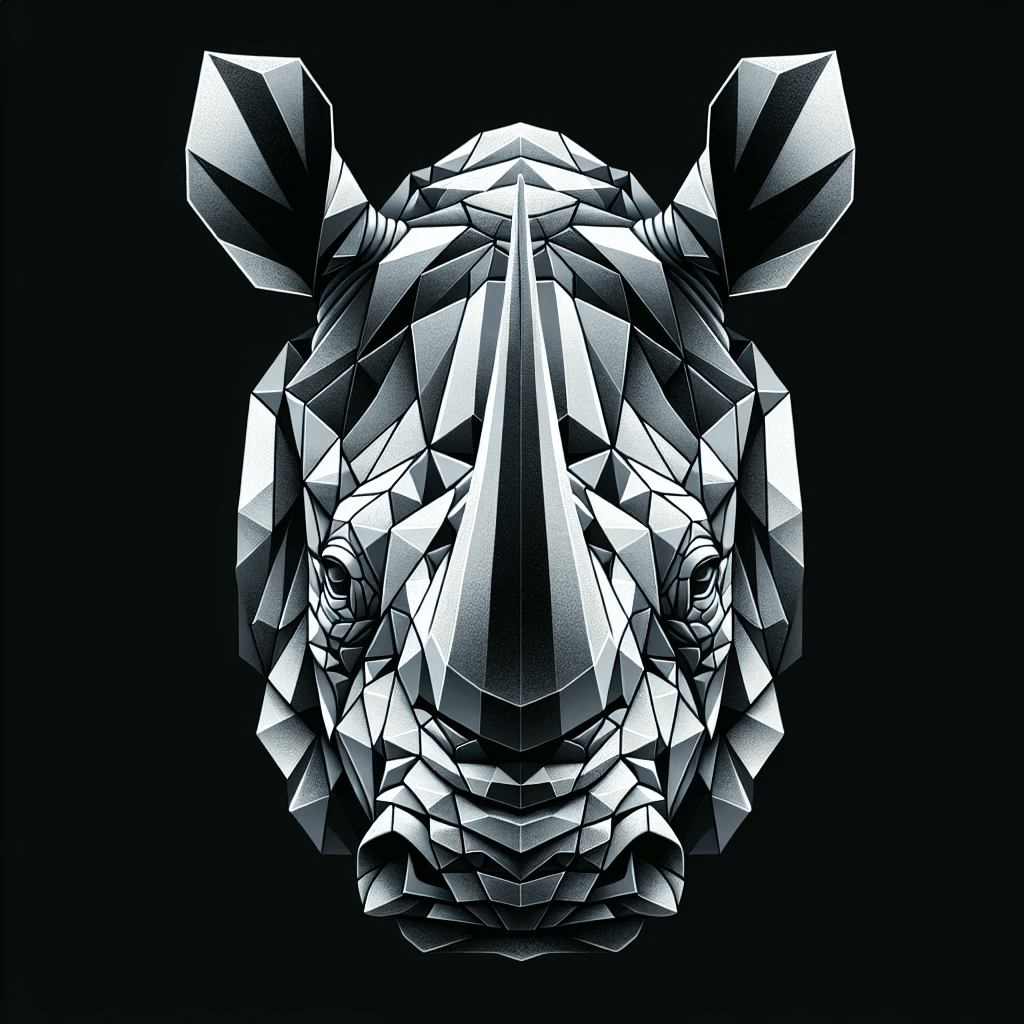} & Create a digital artwork of a stylized, geometric rhinoceros head with a dynamic array of sharp, crystalline facets in a monochromatic palette of black, white, and gray. The design should feature intricate shadows and highlights to produce a three-dimensional illusion, with a focus on accurately representing the creature's contours and muscle structure. Adjust the composition to show the rhinoceros head from a frontal perspective, ensuring that both the horn and the ears are symmetrically aligned in the center. Emphasize the geometric nature of the facets by making them more pronounced and varied in shape, creating a complex mosaic that captures the interplay of light and shadow. Add a slight glow to the edges of the facets to enhance the three-dimensional effect and the metallic quality of the artwork. Display the rhinoceros head against a pitch-black background, with a light source positioned to cast dramatic, high-contrast illumination that emphasizes its multifaceted texture. Incorporate a subtle reflective sheen on the surface to suggest a sleek, metallic finish, and ensure the rhinoceros's eye is detailed and expressive, contributing to the overall lifelike appearance of the artwork.  \\ \midrule

\includegraphics[width=30mm,height=30mm]{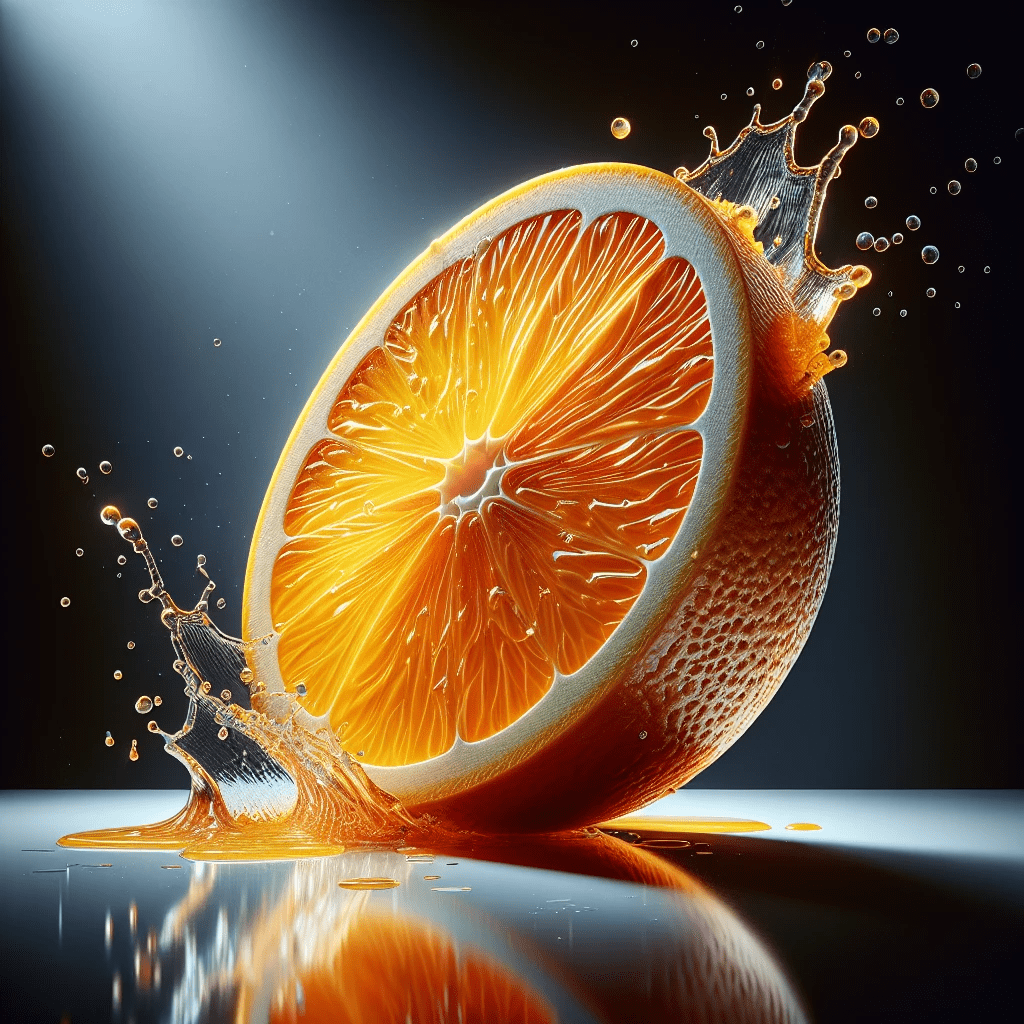} & \includegraphics[width=30mm,height=30mm]{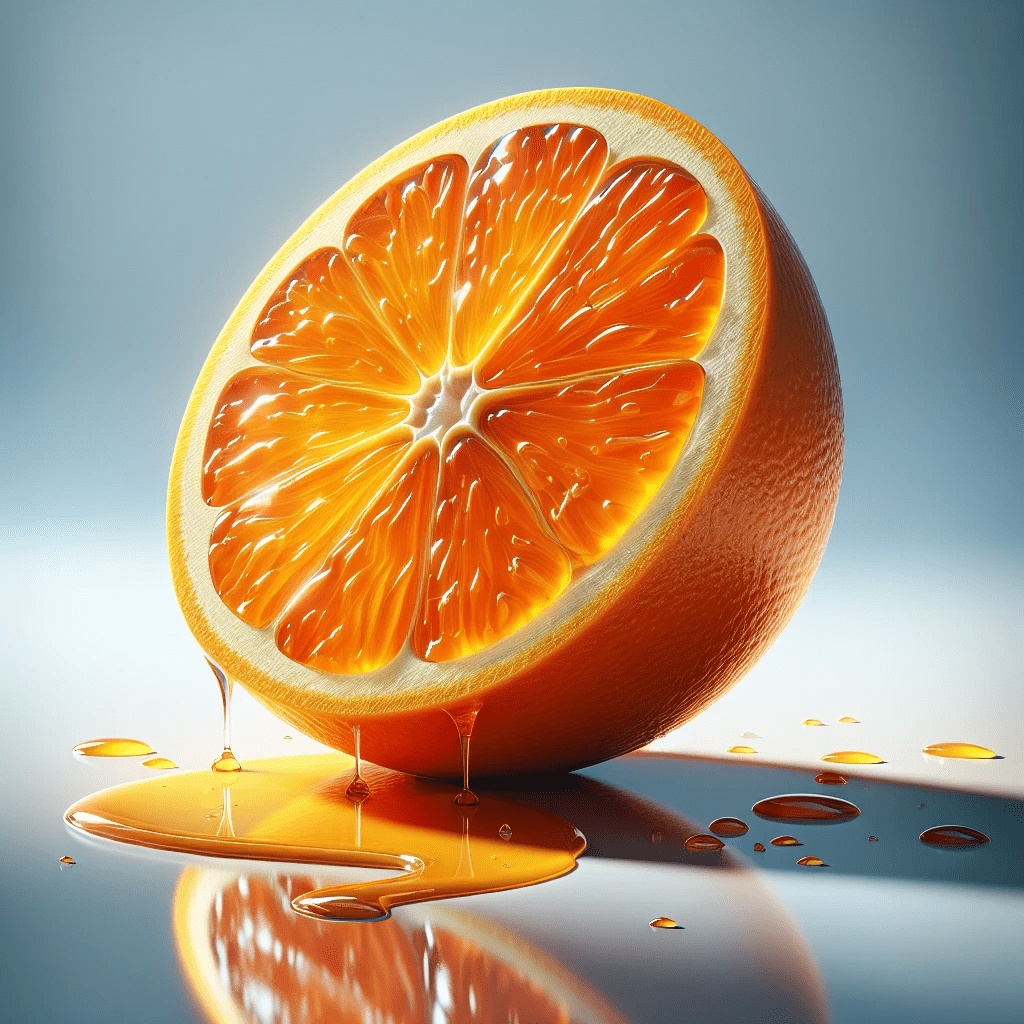} & \includegraphics[width=30mm,height=30mm]{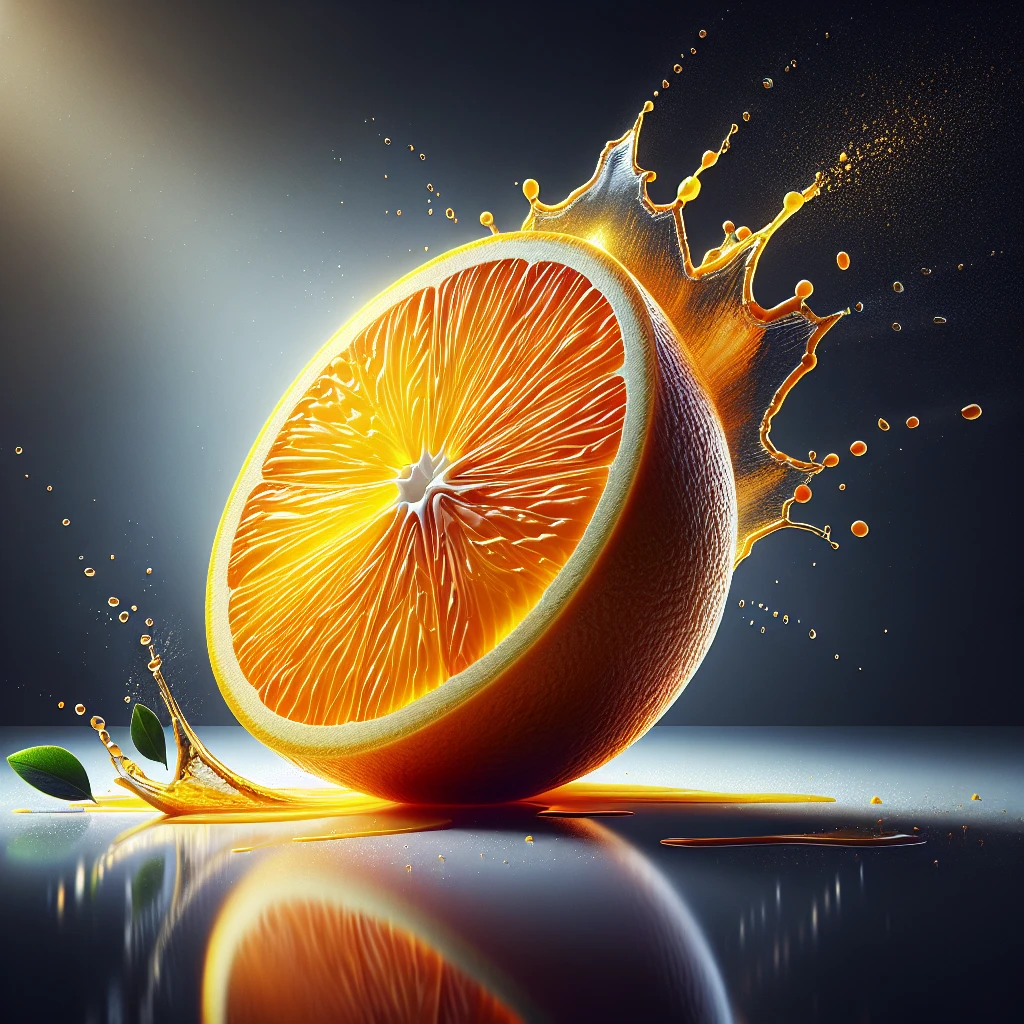} & A hyper-realistic full slice of an orange with intricate details, including the textured pulp and clearly defined rind, positioned off-center on a reflective gradient surface transitioning from white to dark. The orange's juicy texture is accentuated by a dynamic splash of juice, with droplets captured mid-air, creating an energetic and lively scene. The lighting is dramatic and contrasting, with a spotlight effect casting a pronounced shadow to one side to enhance the three-dimensional effect and emphasize the vibrant orange color. Include a clear reflection on the surface and a small stem attached to the orange slice to underscore the realism and freshness. Enhance the composition by ensuring the orange slice is angled slightly, with the splash of juice originating from the lower right side, to add a sense of motion and vitality.  \\
\midrule

\includegraphics[width=30mm,height=30mm]{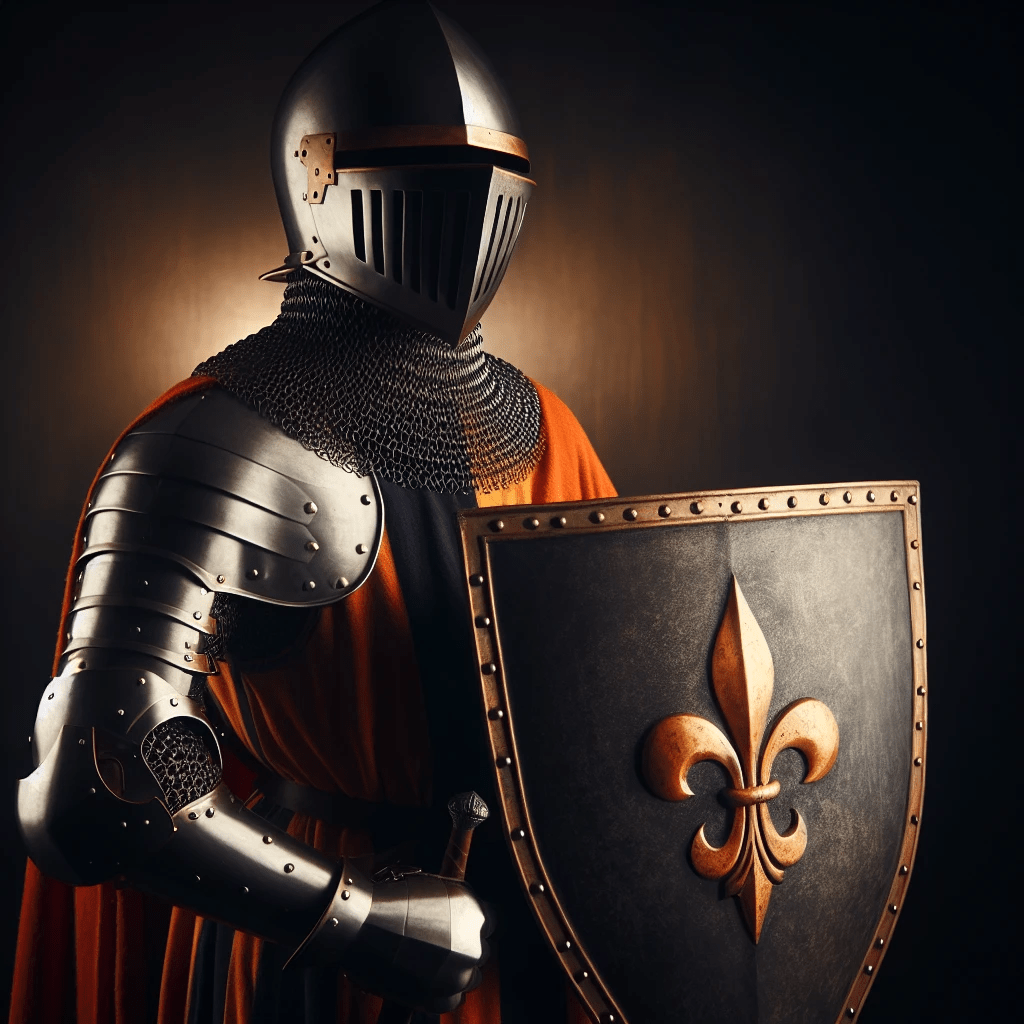} & \includegraphics[width=30mm,height=30mm]{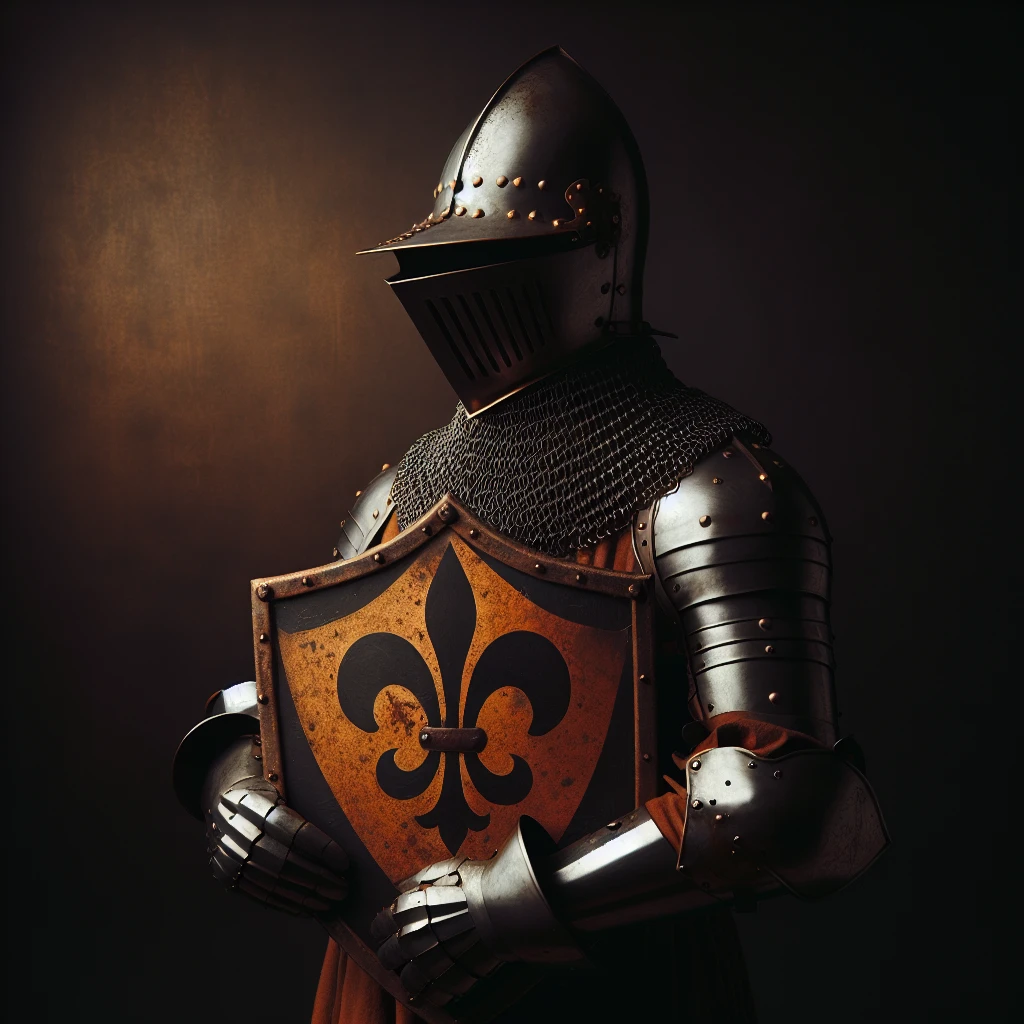} & \includegraphics[width=30mm,height=30mm]{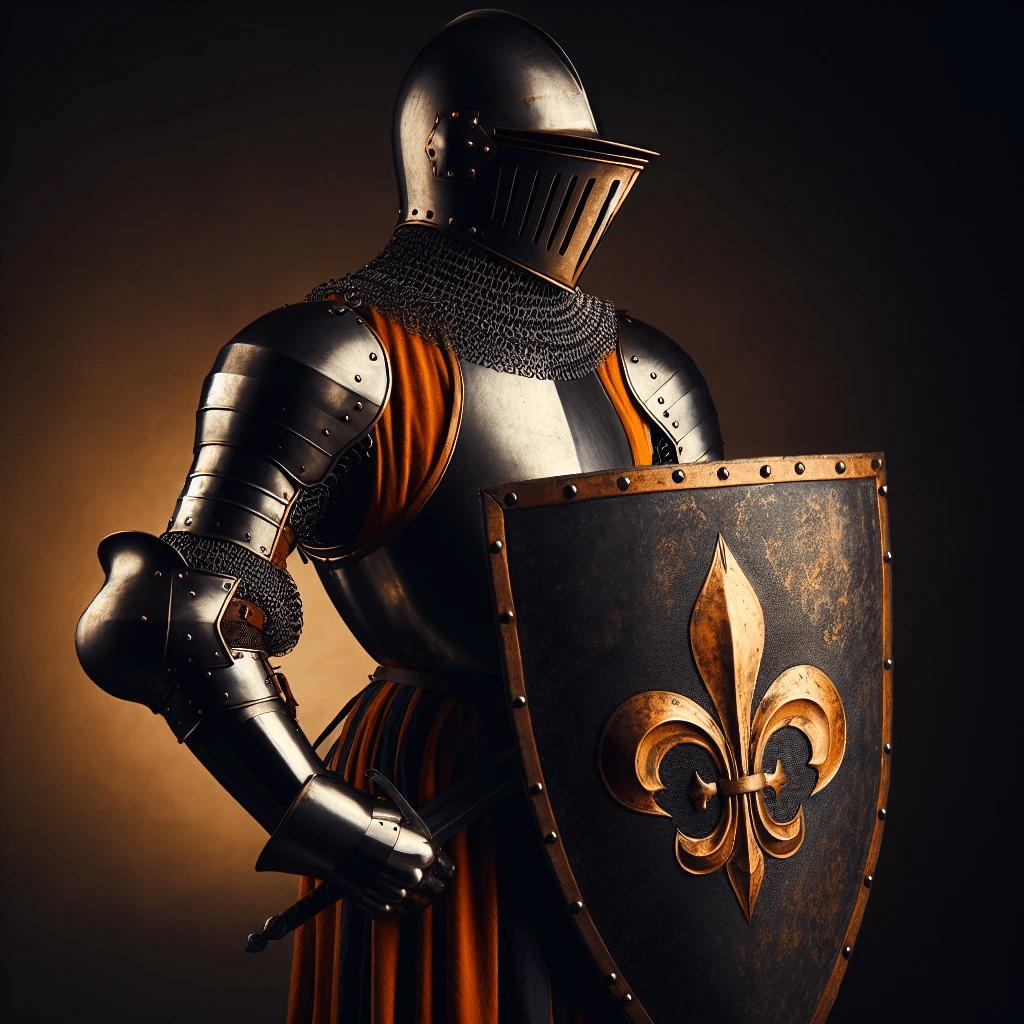} & A medieval knight in full armor stands with a shield, the dark background highlighting his silhouette against a subtle warm glow. His helmet features a visor with a single vertical slit, and his armor includes a chainmail coif beneath a segmented plate gorget and articulated plate gauntlets, with layered plate armor and flared ridged pauldrons. The knight's shield is centered and bears a detailed, embossed golden fleur-de-lis on a field of weathered steel, surrounded by rivets. The vibrant orange cloak drapes over both shoulders and behind his back, adding a touch of regal color to the composition. His stance is grounded and balanced, with his left arm extended, presenting the shield, and his right hand resting on the pommel of his sword, exuding a calm and noble demeanor.  \\ \midrule

\includegraphics[width=30mm,height=30mm]{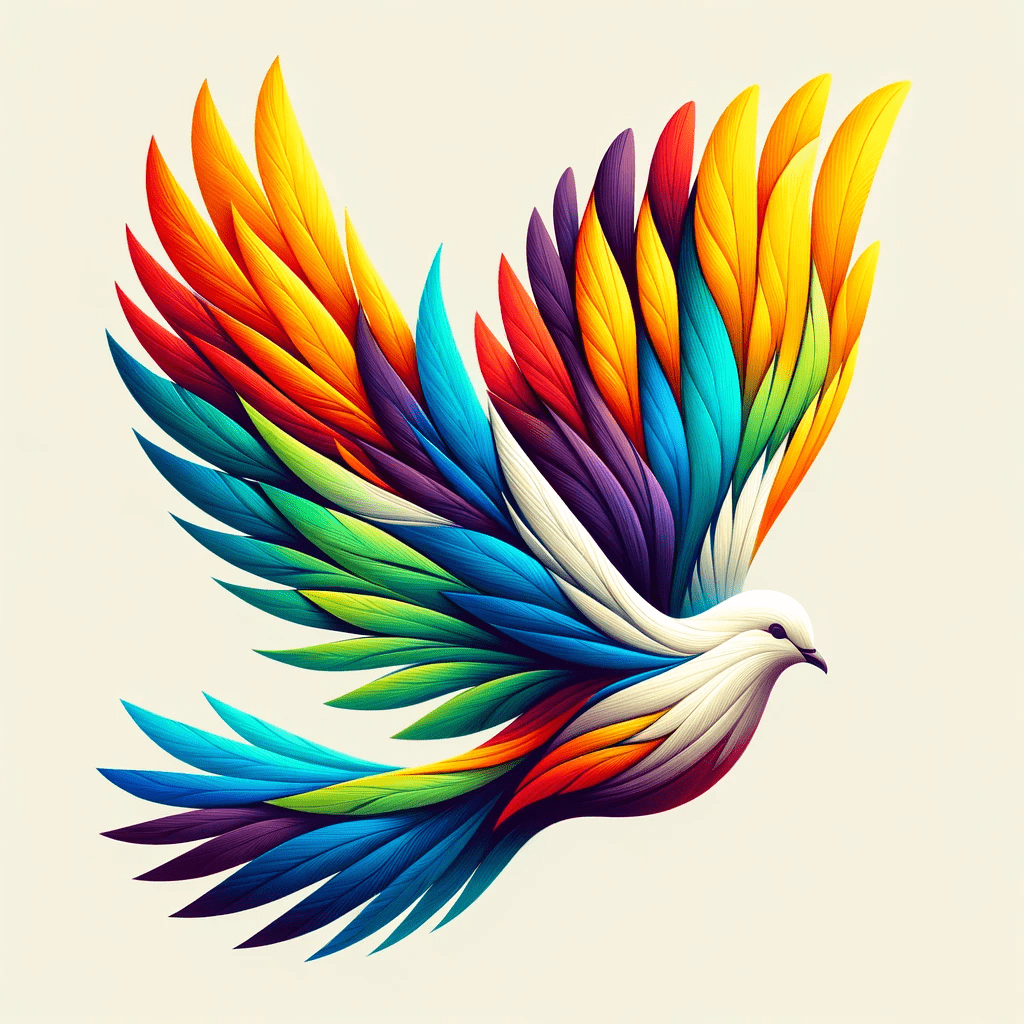} & \includegraphics[width=30mm,height=30mm]{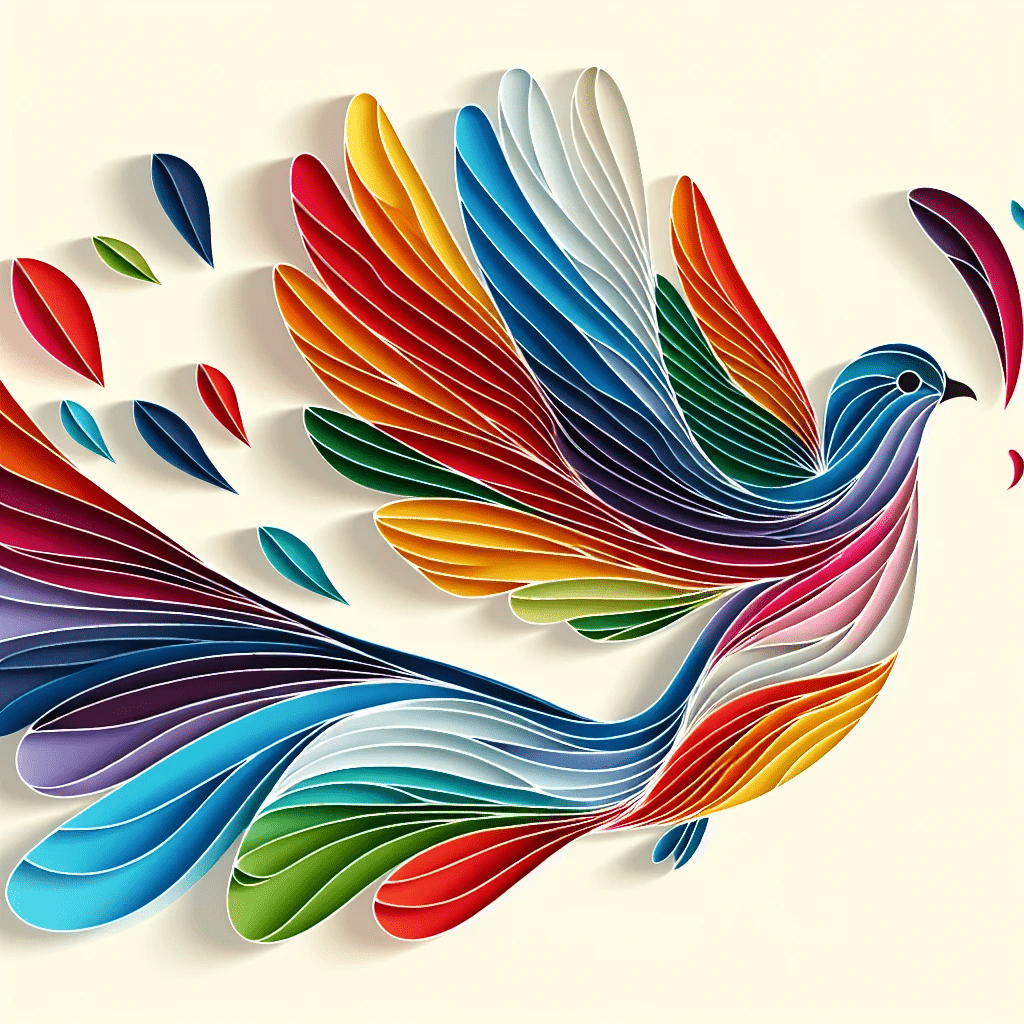} & \includegraphics[width=30mm,height=30mm]{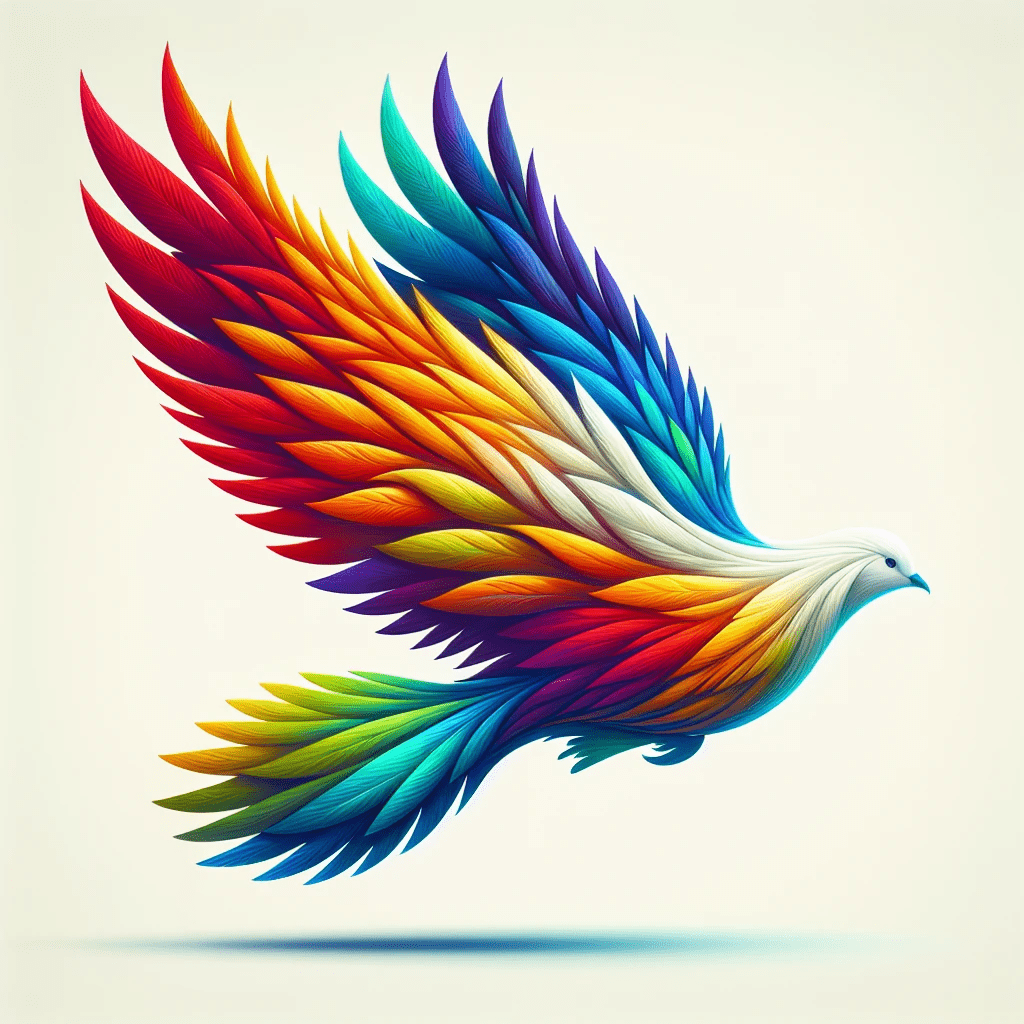} & Create a stylized illustration of a dove in flight, with feathers that transition smoothly through a spectrum of colors including red, orange, yellow, green, blue, indigo, and violet. The dove's plumage should resemble a dynamic, three-dimensional arrangement of vibrant, overlapping feathers, giving a sense of movement and freedom. The style should be a fusion of semi-realistic and digital art, with a focus on vivid colors and a clean, light background that emphasizes the artwork's lively and spirited nature. Adjust the feather arrangement to be more structured and flame-like, with the feathers at the tips being more elongated and pointed to enhance the sense of elegance and flow.  \\
\midrule

\includegraphics[width=30mm,height=30mm]{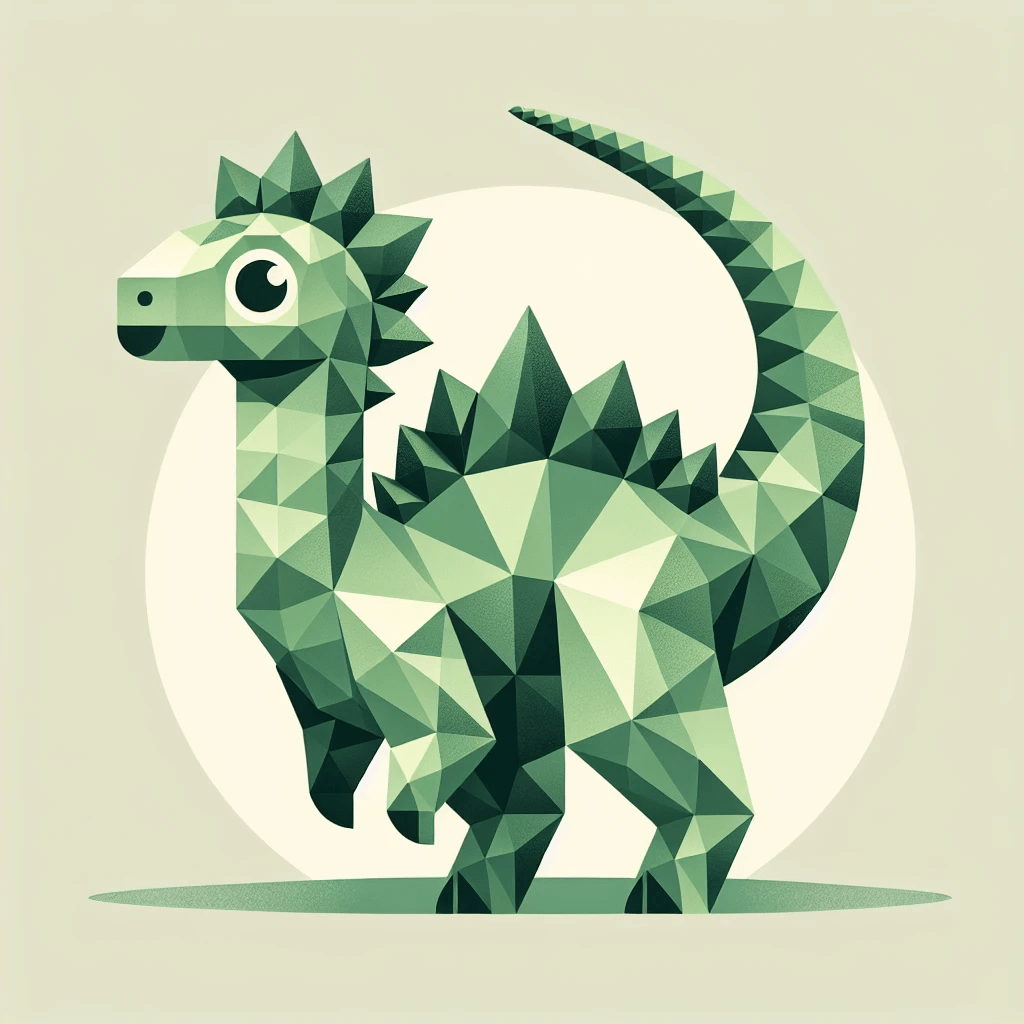} & \includegraphics[width=30mm,height=30mm]{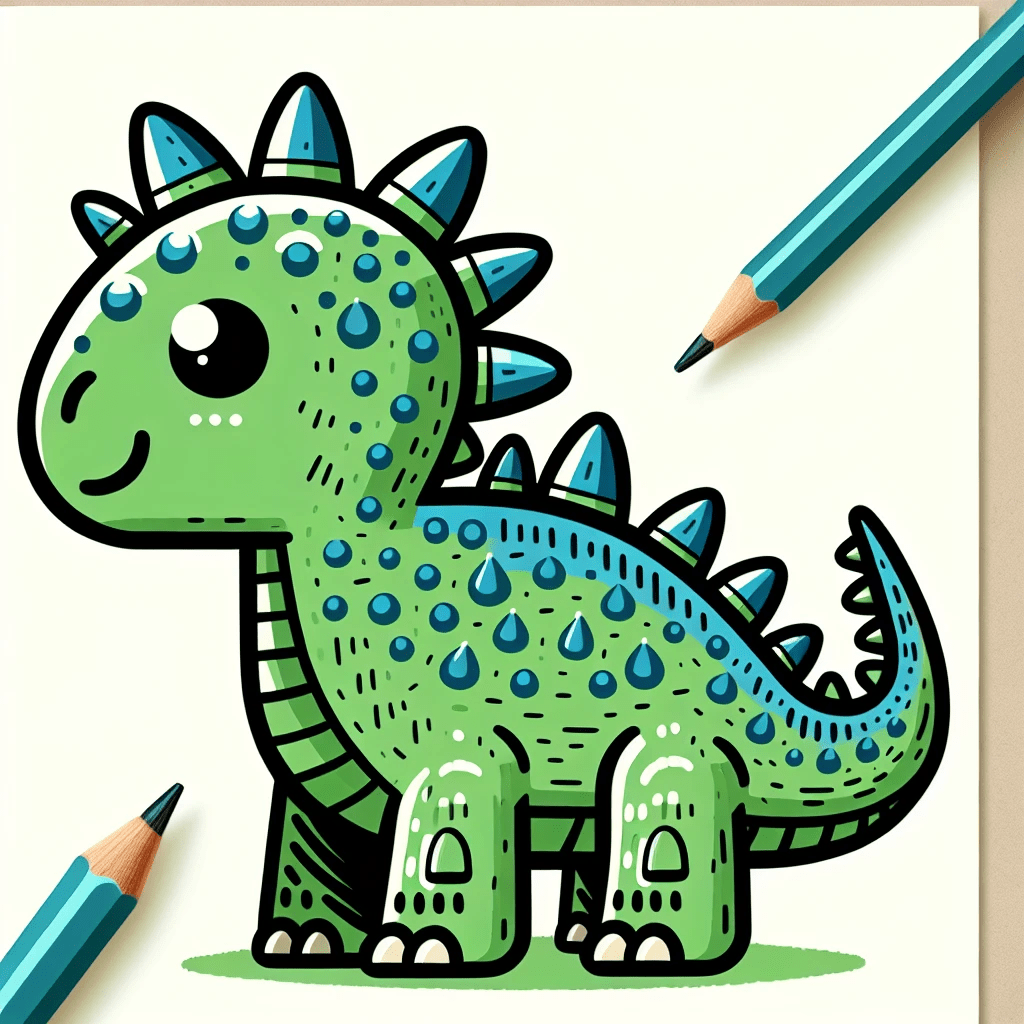} & \includegraphics[width=30mm,height=30mm]{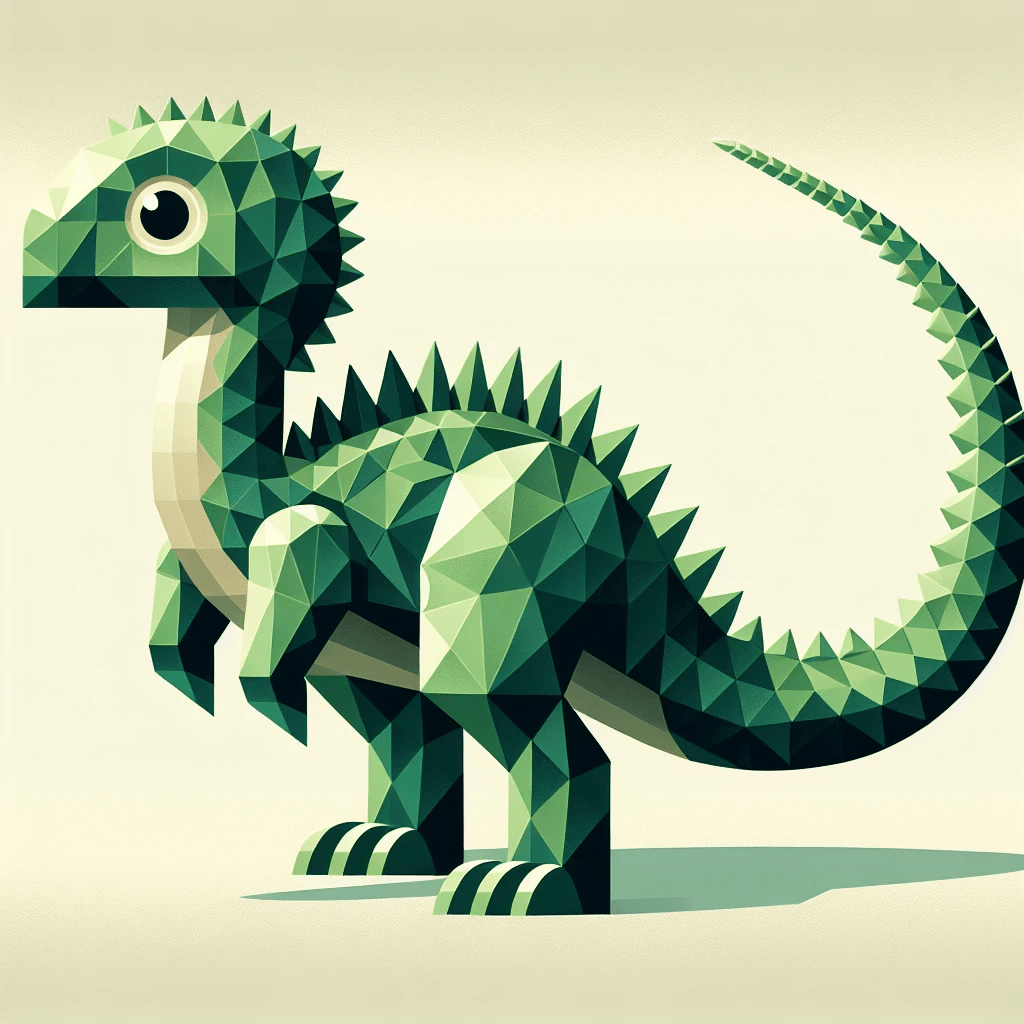} & Create an illustration of a stylized, geometric dinosaur with a textured body in two shades of green: a lighter green for the main body and a darker green for the spiky plates along its back. The dinosaur should have a friendly demeanor, with a long, curved tail and a smooth, rounded head featuring two small, circular white eyes with black pupils. It should stand on two legs with small, rounded feet, each with three visible toes. The background should be a flat, light beige color, with a simple, elongated shadow extending to the right of the dinosaur, indicating a soft light source to the left.  \\ \midrule

\includegraphics[width=30mm,height=30mm]{t2i/show_example3_original.jpg}  & \includegraphics[width=30mm,height=30mm]{t2i/show_example3_init.jpg} & \includegraphics[width=30mm,height=30mm]{t2i/show_example3_final.jpg} & Generate an image of a cartoon-style polar bear with gleefully closed eyes and a wide, toothy grin, revealing just a hint of its tongue. The bear should look exuberant, standing on its hind legs with arms open wide as if ready for a hug. The bear's fur should appear extremely soft and fluffy, with a pronounced blush of rosy pink on both cheeks and belly, enhancing its charm. Adorn the bear with a cozy, chunky-knit scarf, vibrant red with prominent, horizontal white stripes, stylishly wrapped around its neck and draping with a dense tassel fringe at the ends. Situate the bear against a gentle pastel pink backdrop, scattered with delicate, small snowflakes, conveying the splendor and coziness of festive winter cheer.
\\ \midrule

\includegraphics[width=30mm,height=30mm]{t2i/show_example5_original.jpg}  & \includegraphics[width=30mm,height=30mm]{t2i/show_example5_init.jpg} & \includegraphics[width=30mm,height=30mm]{t2i/show_example5_final.jpg} & An anthropomorphic duck standing confidently with hands on hips, styled as a classic film noir detective. The duck has a calm and cool expression, wearing a tan detective's fedora hat and a matching double-breasted trench coat, buttoned up, with a broad collar, epaulets, and a belted waist. The character has a white shirt and a patterned tie with a diagonal stripe design underneath. The character has orange webbed feet and a large, prominent beak. The lighting is dramatic, with a strong contrast between light and shadow, creating a focused shadow on the background that mimics the character's silhouette. The overall color palette is warm with a gentle light source coming from the side, casting the background in a gradient from warm beige to shadows, giving the image a mysterious and dramatic appearance. \\

\bottomrule[1.5pt]
\end{NiceTabular}
}
\caption{\textbf{More results of prompt inversion.}}
\label{tab:more_inversion}
\end{table*}

\begin{table*}[h]
\centering
\scalebox{0.75}{
\begin{NiceTabular}{M{0.25\linewidth} M{0.25\linewidth} M{0.55\linewidth} M{0.20\linewidth}}
\CodeBefore
    \Body
\toprule[1.5pt]
\textbf{User Query} & \textbf{Final Image} & \textbf{Final Prompt} & \textbf{Reason for Failure} \\ \midrule
\Block{1-4}{{\bf Text-to-image generation}} \\

There are fewer forks than spoons. & \includegraphics[width=45mm,height=45mm]{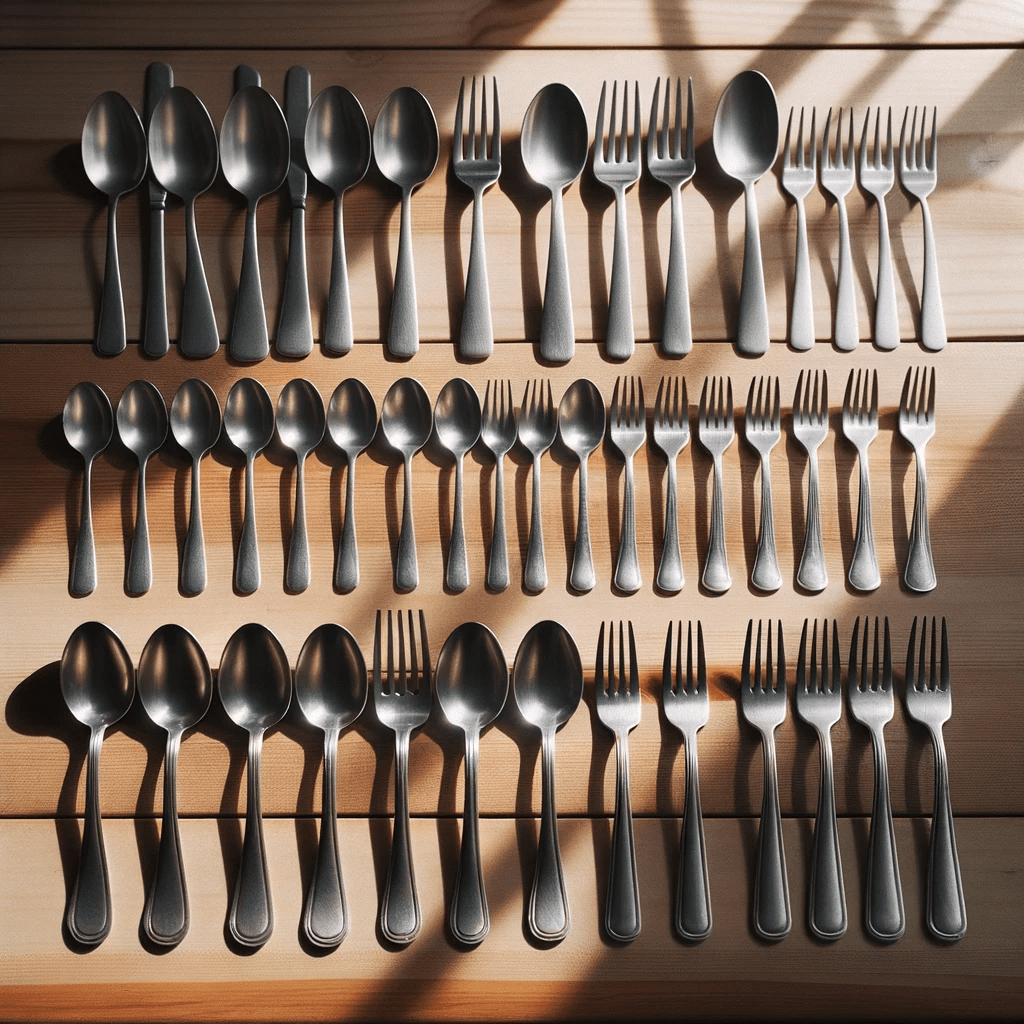} & Photo of a wooden dining table with a collection of spoons and forks laid out. It's clear upon counting that there are fewer spoons than forks. The spoons are stainless steel and placed on the left while the forks, also stainless steel, are on the right. Sunlight filters from a nearby window, casting soft shadows.  & Failure to compare the exact number of forks and spoons in an image. \\ \midrule

 The blue balloon is being lifted by the weight. & \includegraphics[width=45mm,height=45mm]{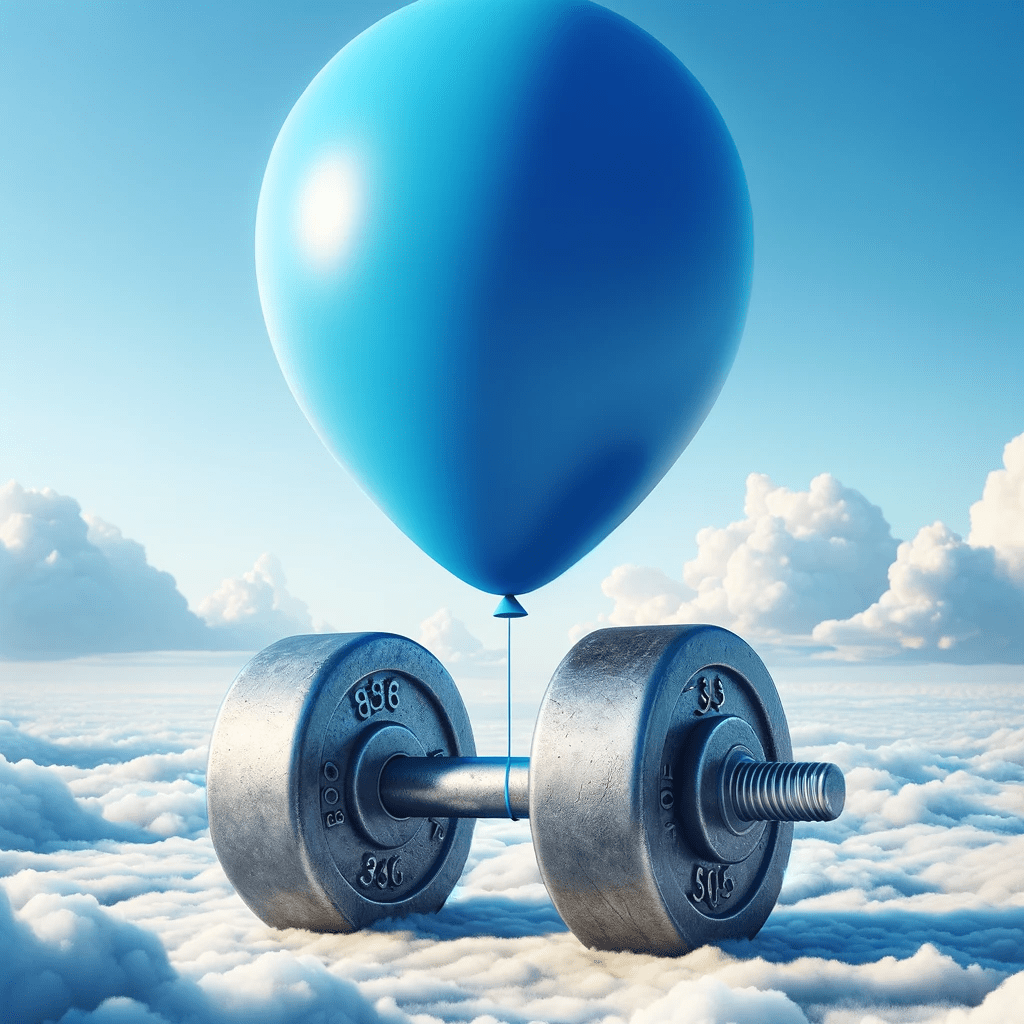} & A surreal scene where a blue balloon appears to be lifting a weight. The weight is realistically depicted, with clear markings indicating its heaviness, while the balloon is oversized and vibrant blue, floating in a clear sky. The scene plays with the concept of reverse gravity, where the balloon, typically light and airy, is paradoxically lifting the heavy weight, creating a visually intriguing and thought-provoking image.  & Failure to incorporate unusual / counterintuitive situations. \\ \midrule

Seat numbers increasing from right to left. & \includegraphics[width=45mm,height=45mm]{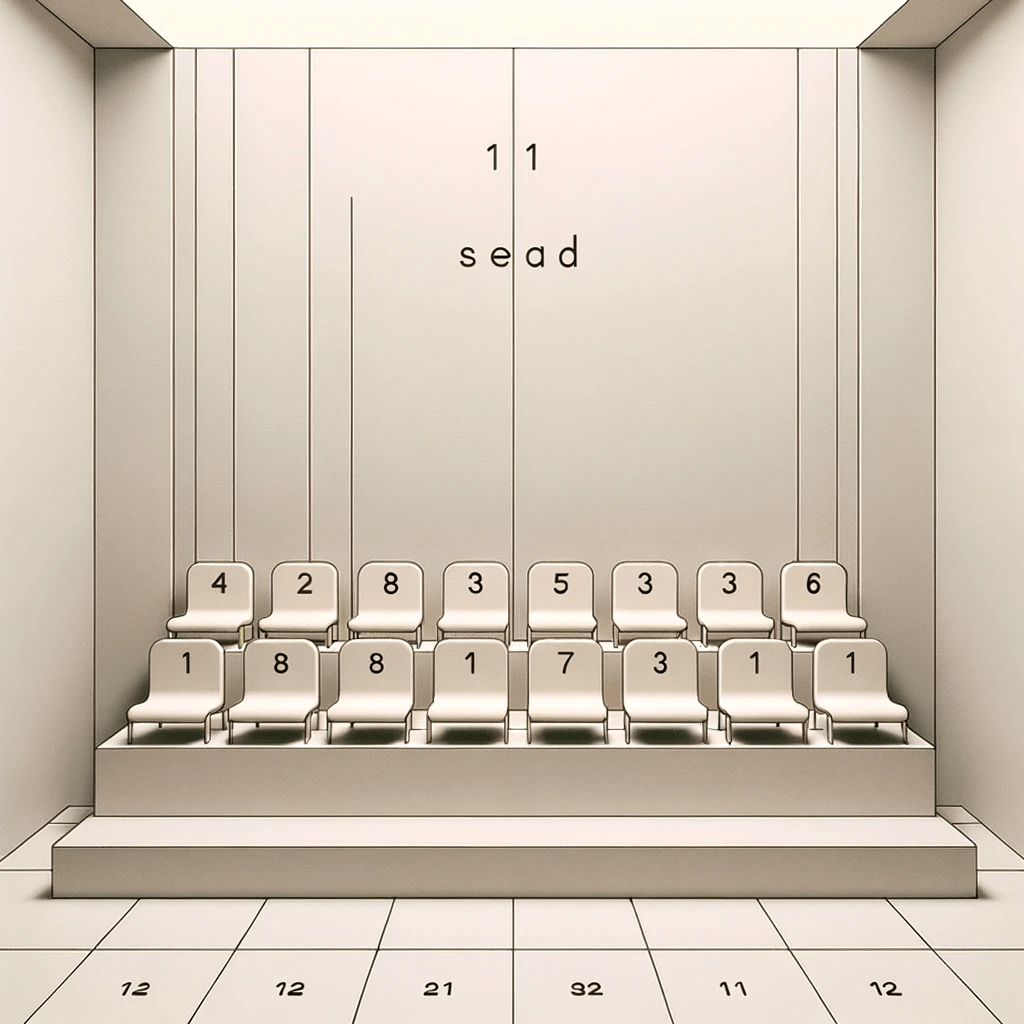} & Illustration of a series of seats in a minimalist room. The seats on the right start with the number '1' and the numbers increase as we move to the left. The room has a neutral color palette to ensure focus on the seat numbers.  & Failure to sort the chair numbers in increasing order.  \\ 
\bottomrule[1.5pt]
\end{NiceTabular}

}
\caption{\textbf{Failure cases of T2I generation.} We note that some Winoground queries that involve commonsense reasoning (e.g., mathematical reasoning, counting) are still too challenging even for DALL-E 3. We expect better results with stronger generative models in the future.}
\label{tab:failure_t2i}
\end{table*}

\begin{table*}[h]
\centering
\scalebox{0.75}{
\begin{NiceTabular}{M{0.25\linewidth} M{0.25\linewidth} M{0.55\linewidth} M{0.20\linewidth}}
\CodeBefore
    \Body
\toprule[1.5pt]
\textbf{User Query} & \textbf{Final Image} & \textbf{Final Prompt} & \textbf{Reason for Failure} \\ \midrule
\Block{1-4}{{\bf Prompt Inversion}} \\

\includegraphics[width=45mm,height=45mm]{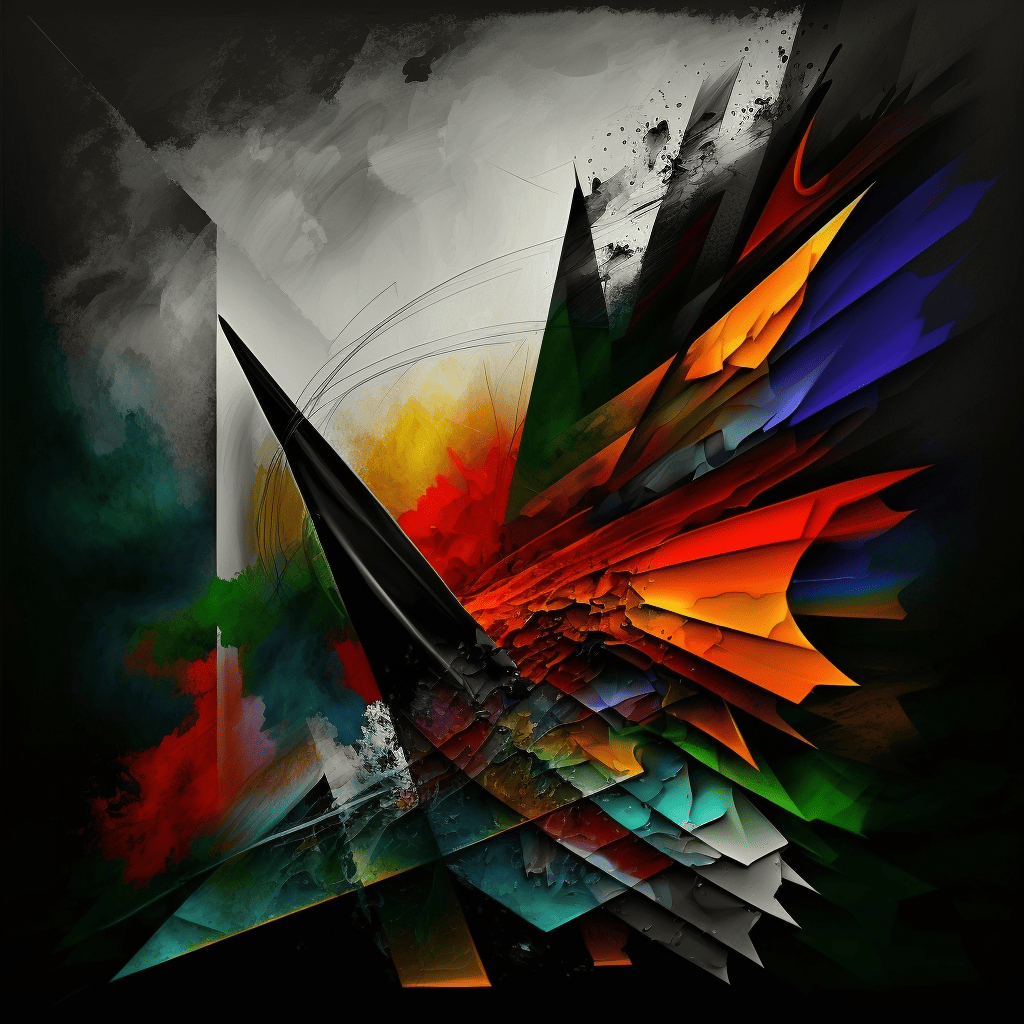} & \includegraphics[width=45mm,height=45mm]{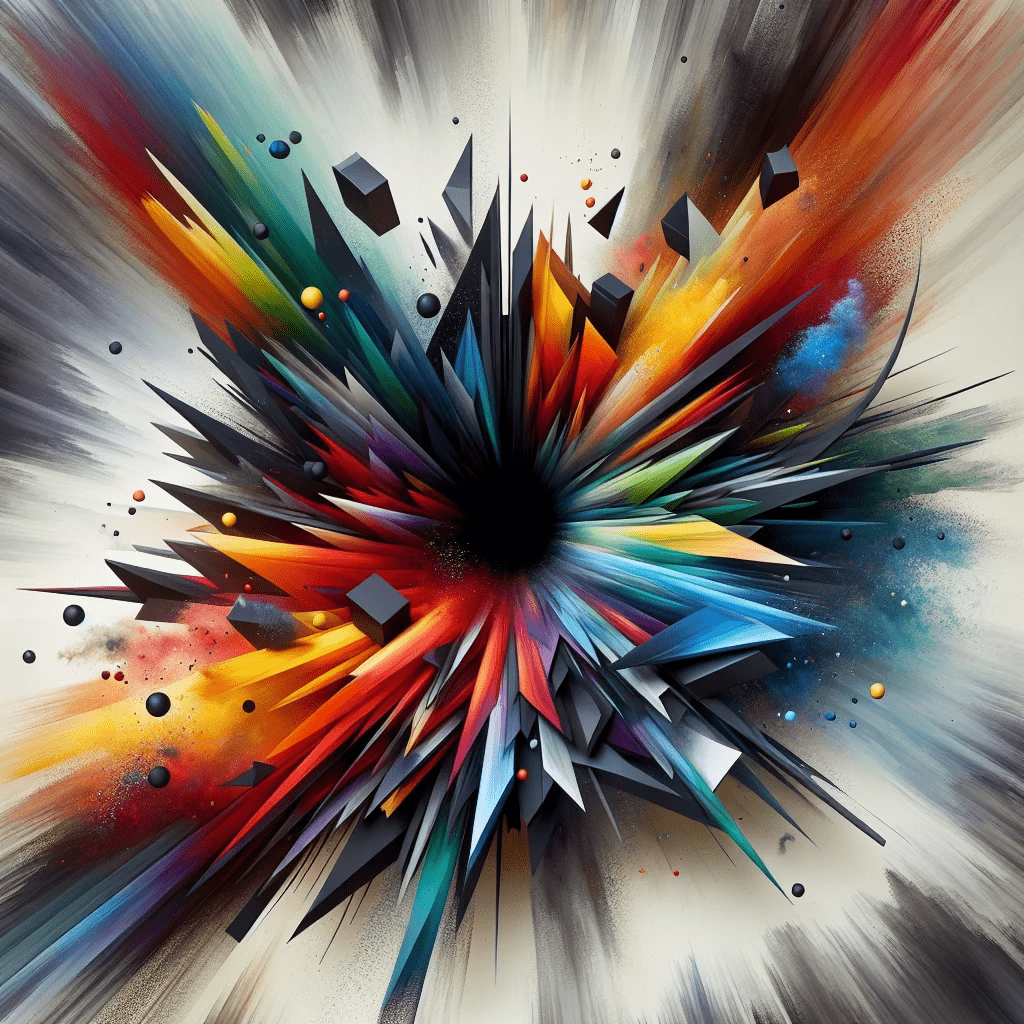} & Create an abstract composition with a dynamic array of shattered, angular shapes emanating from a central point towards the edges of the image. Intensify the contrast by incorporating a deep black void at the core, surrounded by a gradient of vivid colors like red, orange, yellow, green, blue, and indigo transitioning from warm to cool tones to represent this burst of shapes. Add a contrasting background with subtle grayscale gradients, smudges, and paint splatters to enhance the sense of explosion and movement. Include sharp, crisp edges on the shapes to give a sense of three-dimensionality and depth. Ensure the overall effect is of a high-contrast, visually impactful piece that combines both geometric and organic elements, with a clear distinction between the vibrant center and the muted, textured periphery. Adjust the composition to have a more chaotic arrangement of shapes with varying sizes and directions, and incorporate a mix of both soft and hard edges to add complexity. Emphasize a more pronounced use of shadows and highlights to give the shapes a more tangible feel and enhance the illusion of depth and volume.  & Highly challenging abstract details with complex atypical shapes are difficult to describe in detail, even for GPT-4V.  \\ \midrule

\includegraphics[width=45mm,height=45mm]{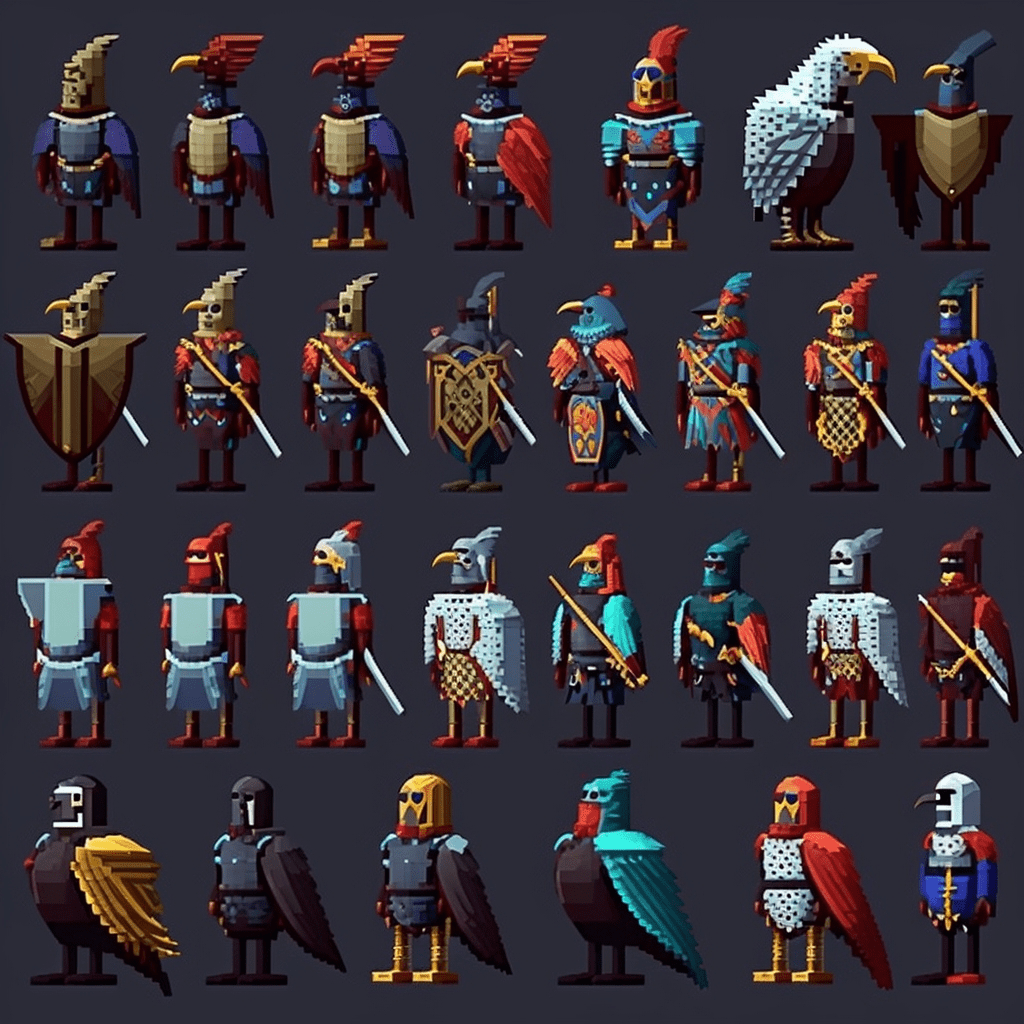} & \includegraphics[width=45mm,height=45mm]{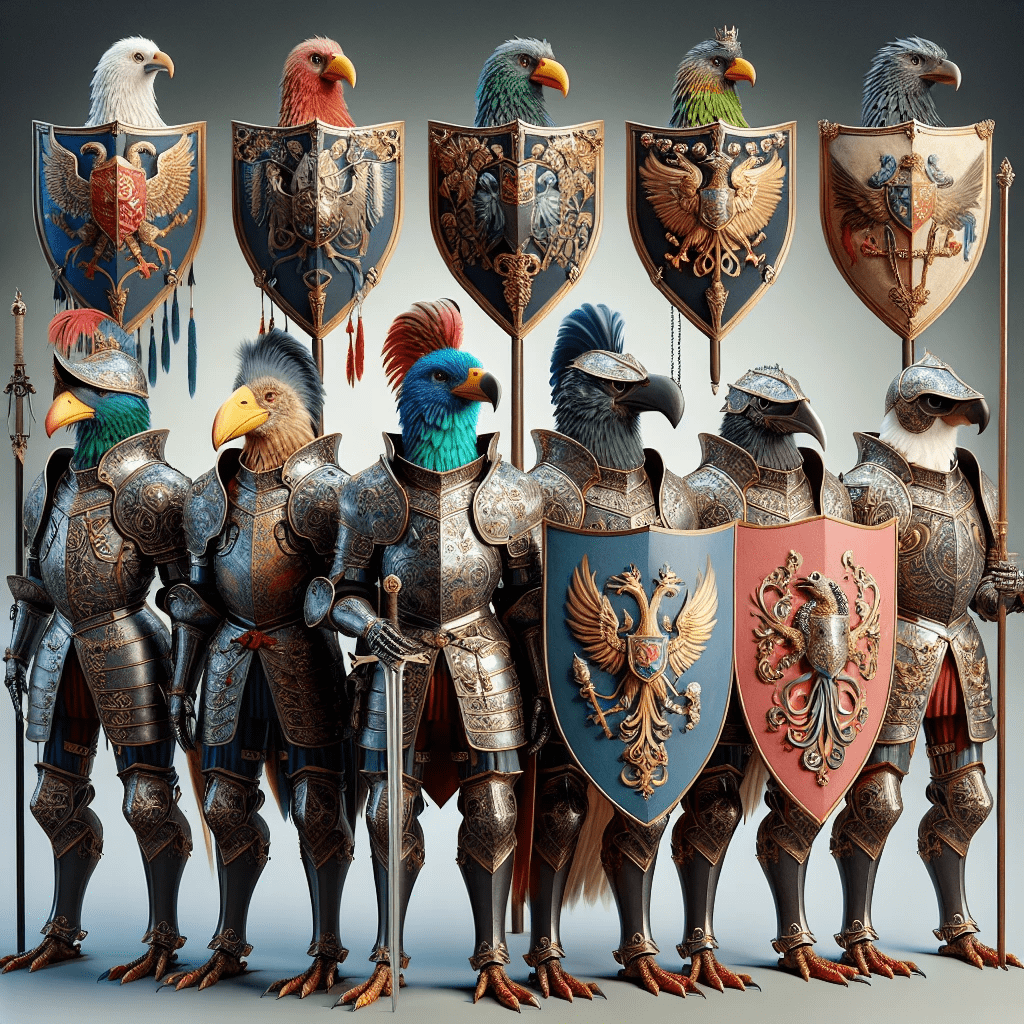} & Create a collection of highly detailed, anthropomorphic bird knights with meticulously crafted medieval armors and heraldic shields, standing in a 3x3 grid formation against a smooth, gradient background. Each bird should display intricate feather patterns and vibrant colors true to real bird species, with helmets thoughtfully designed to accommodate their beaks and crests. The armor should be complete with ornate shoulder plates, breastplates, gauntlets, and greaves, while the shields are to be kite or tower shield shaped, adorned with elaborate coat of arms featuring mythical creatures. Armaments will include finely wrought swords, lances, and axes. Aim for a high-fidelity 3D rendering style with a sophisticated color palette and dynamic lighting to accentuate the textures and metallic sheen of the armors, ensuring each knight is posed in a stately and dignified manner.  & Failure to determine the exact number of objects in an image, especially if the values are greater than 10 or are not in a uniform pattern (grid-shaped).  \\ \midrule

\includegraphics[width=45mm,height=45mm]{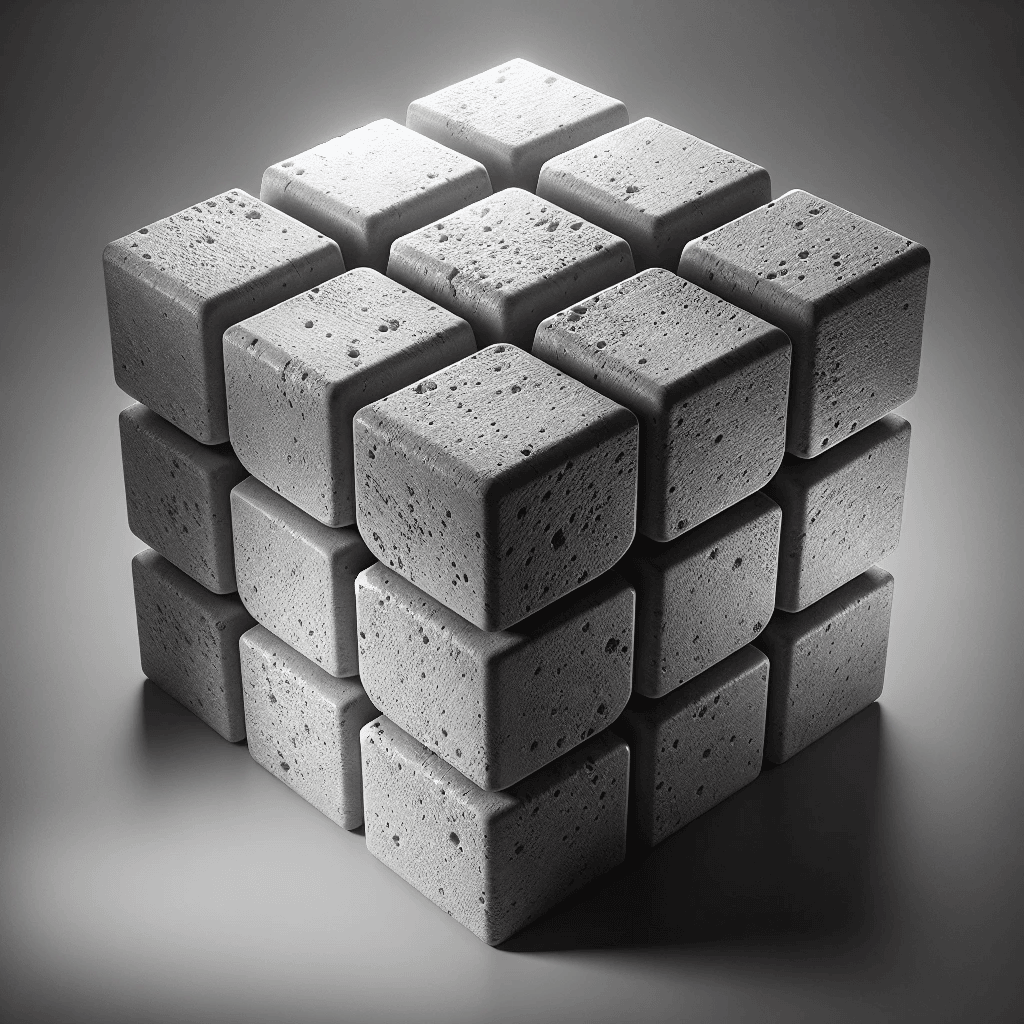}  & \includegraphics[width=45mm,height=45mm]{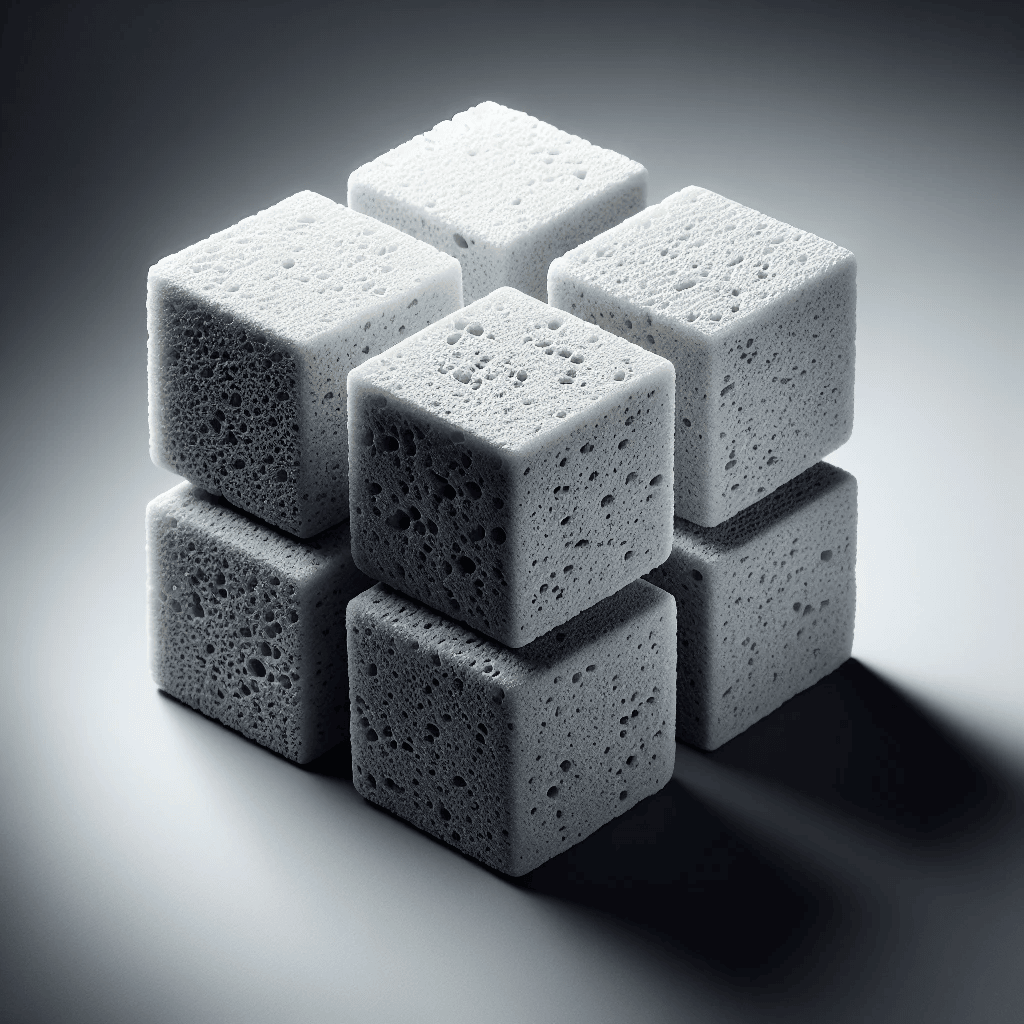}  & Create a 3x3x3 cube arrangement of light grey pumice stones with visible pores and rough texture, each stone equally sized and cube-shaped, on a gradient dark to light gray background with soft focused shadows and a glossy surface reflective of studio lighting. Adjust the lighting to create a more pronounced contrast, highlighting the top edges of the cubes and casting a subtle shadow on the right side, ensuring the image is sharp and high-resolution at a close-up angle to showcase the detail of the stones' textures, with the topmost center stone slightly brighter as if catching more light.  & DALLE-3 fails to understand the meaning of a 3x3x3 set of cubes multiple times, and performs poorly on geometry and patterns in 3 dimensions.  \\ \midrule

\includegraphics[width=45mm,height=45mm]{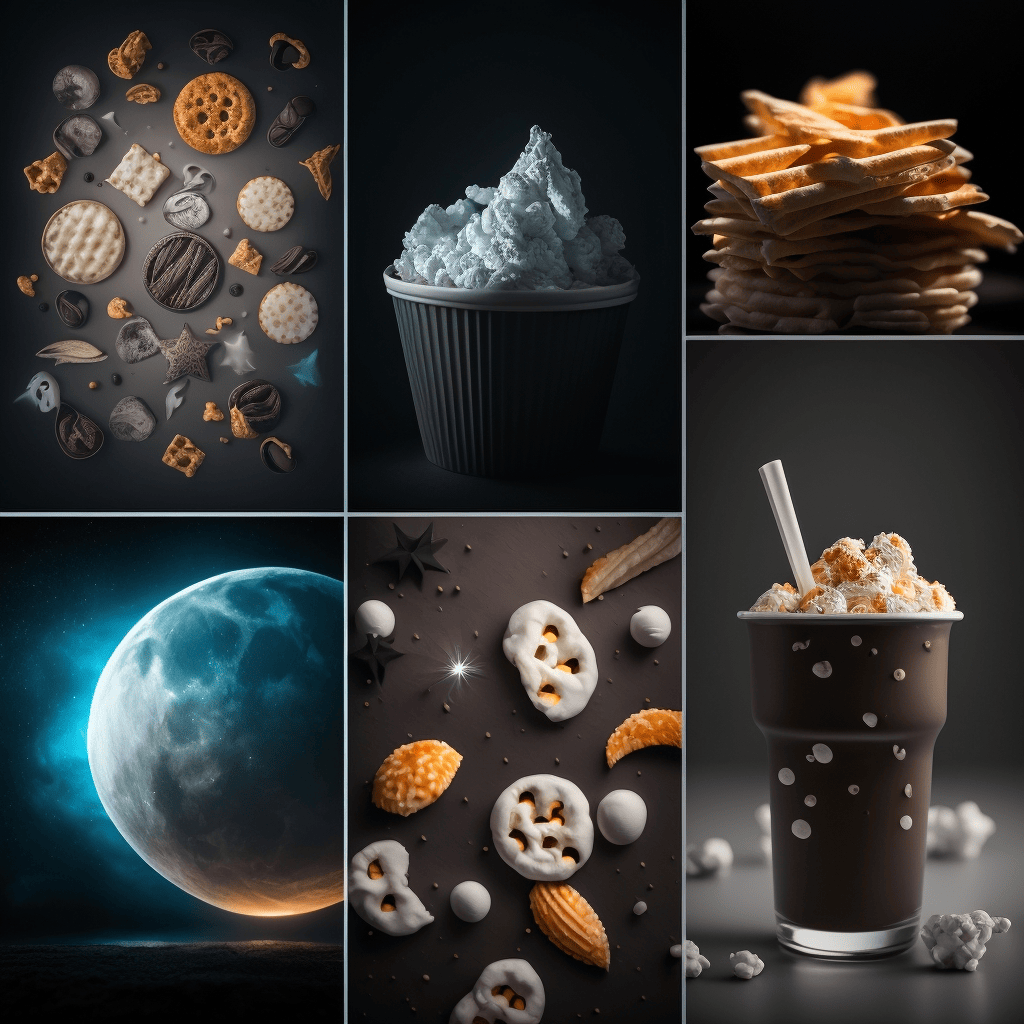}  & \includegraphics[width=45mm,height=45mm]{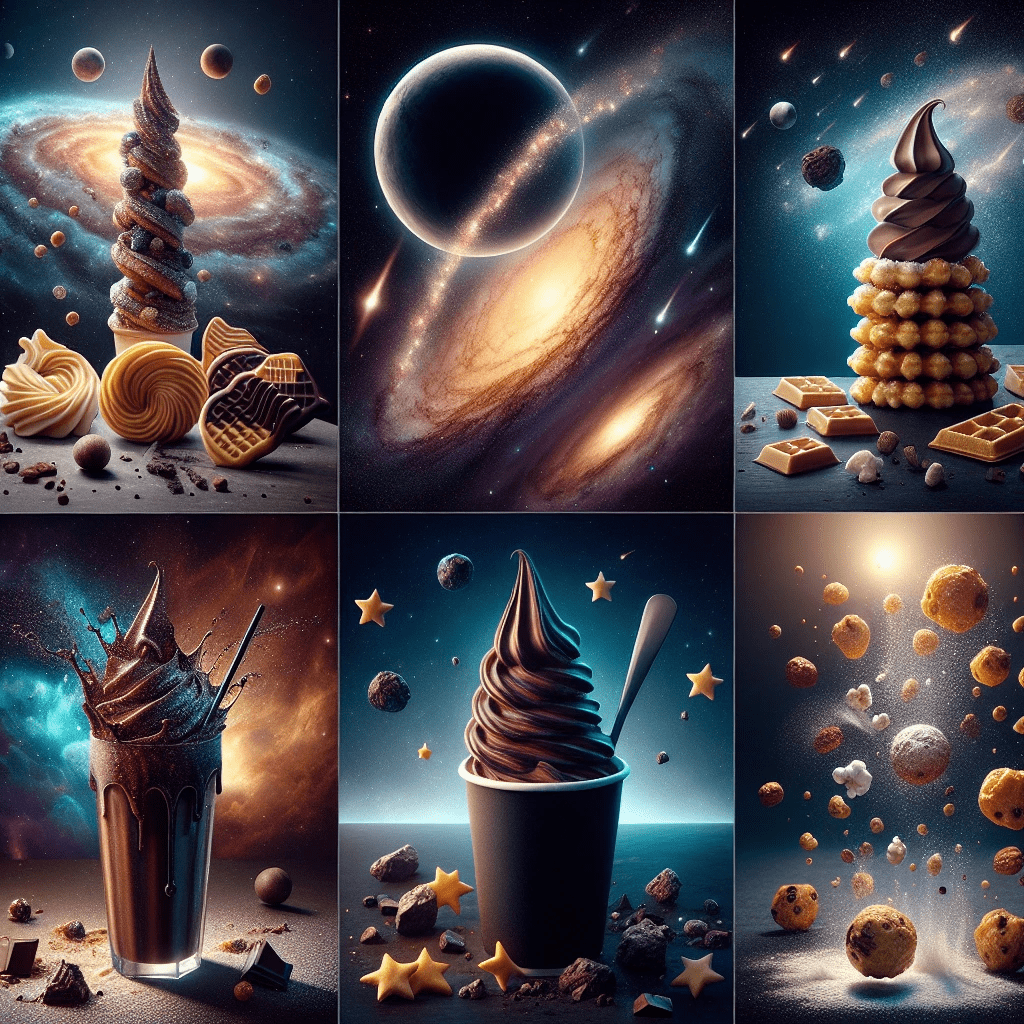} & Create a collage of six images with a cosmic gastronomy theme: top-left depicts an assortment of cookies and chocolates arranged to mimic a galaxy on a space-like background; top-middle features a swirl of soft-serve ice cream in a dark cup, resembling a nebula against a starry sky; top-right displays a stack of golden brown waffles with a dusting of powdered sugar, resembling a celestial body; bottom-left shows a large, detailed moon looming over a twilight horizon; bottom-middle captures various sweets and snacks cascading onto a shadowy surface, evoking a meteor shower; bottom-right presents a cup filled with popcorn and a straw, giving the illusion of a galaxy-themed beverage, set against a backdrop of floating popcorn and sparkling stars.  & DALLE-3 fails to disambiguate details between panels, eventually confusing GPT-4V as well from the comparison.  \\ 
\bottomrule[1.5pt]
\end{NiceTabular}

}
\caption{\textbf{Failure cases of prompt inversion.} We find that our method produces suboptimal results for challenging query images. These involve images that are too abstract to describe, contain too many objects, require geometric reasoning, or involve multiple panels.}
\label{tab:failure_inversion}
\end{table*}

\begin{table*}[h]
\centering
\scalebox{0.72}{
\begin{NiceTabular}{M{0.16\linewidth} M{0.16\linewidth} M{0.16\linewidth} M{0.7\linewidth}}
\CodeBefore
    \Body
\toprule[1.5pt]
\textbf{User Query} & \textbf{Init. Image} & \textbf{Final Image} & \textbf{Final Prompt} \\ \midrule
\Block{1-4}{{\bf Prompt inversion}} \\
\includegraphics[width=30mm]{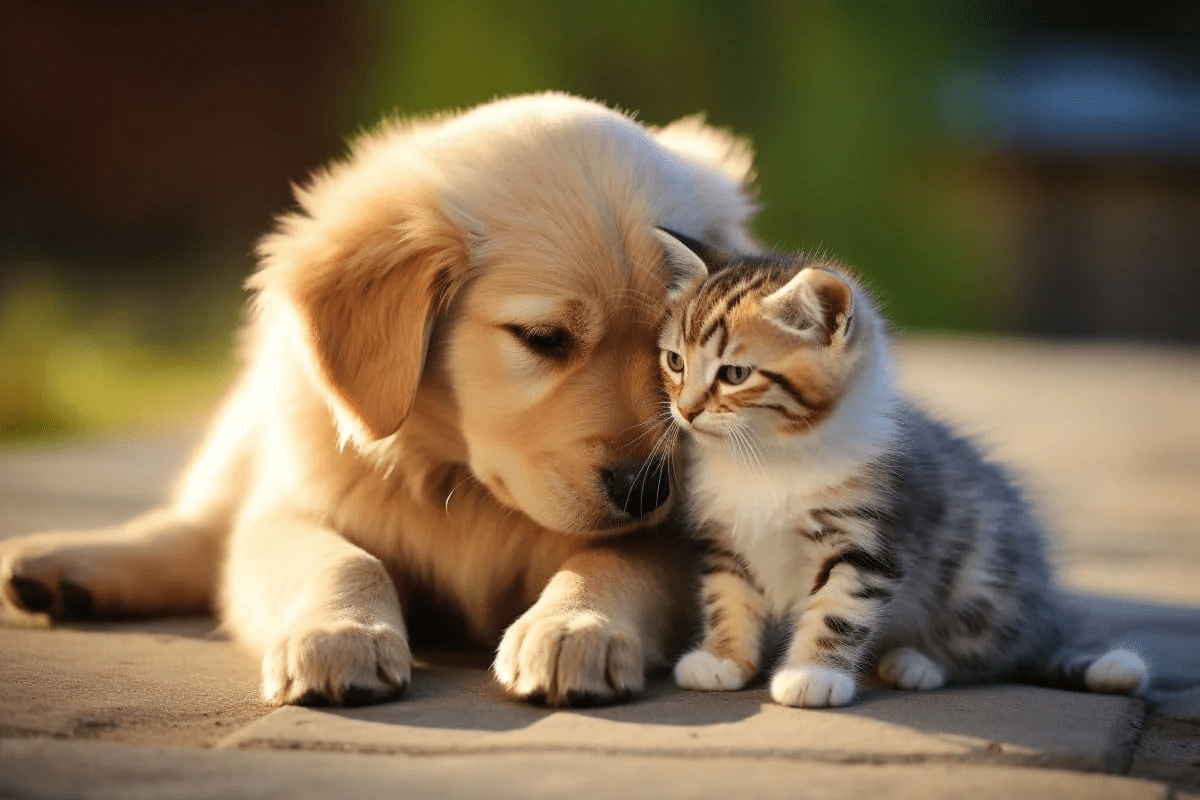} & \includegraphics[width=30mm,height=30mm]{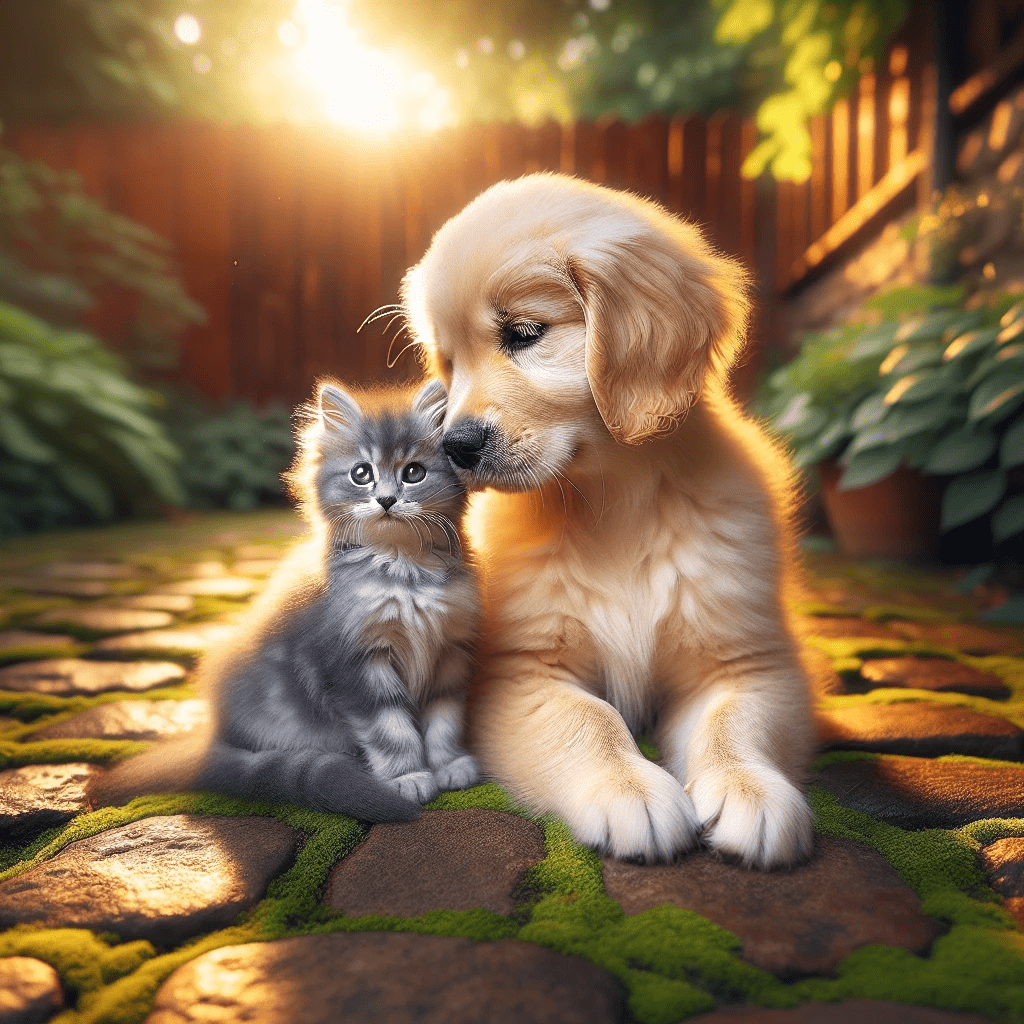} & \includegraphics[width=30mm,height=30mm]{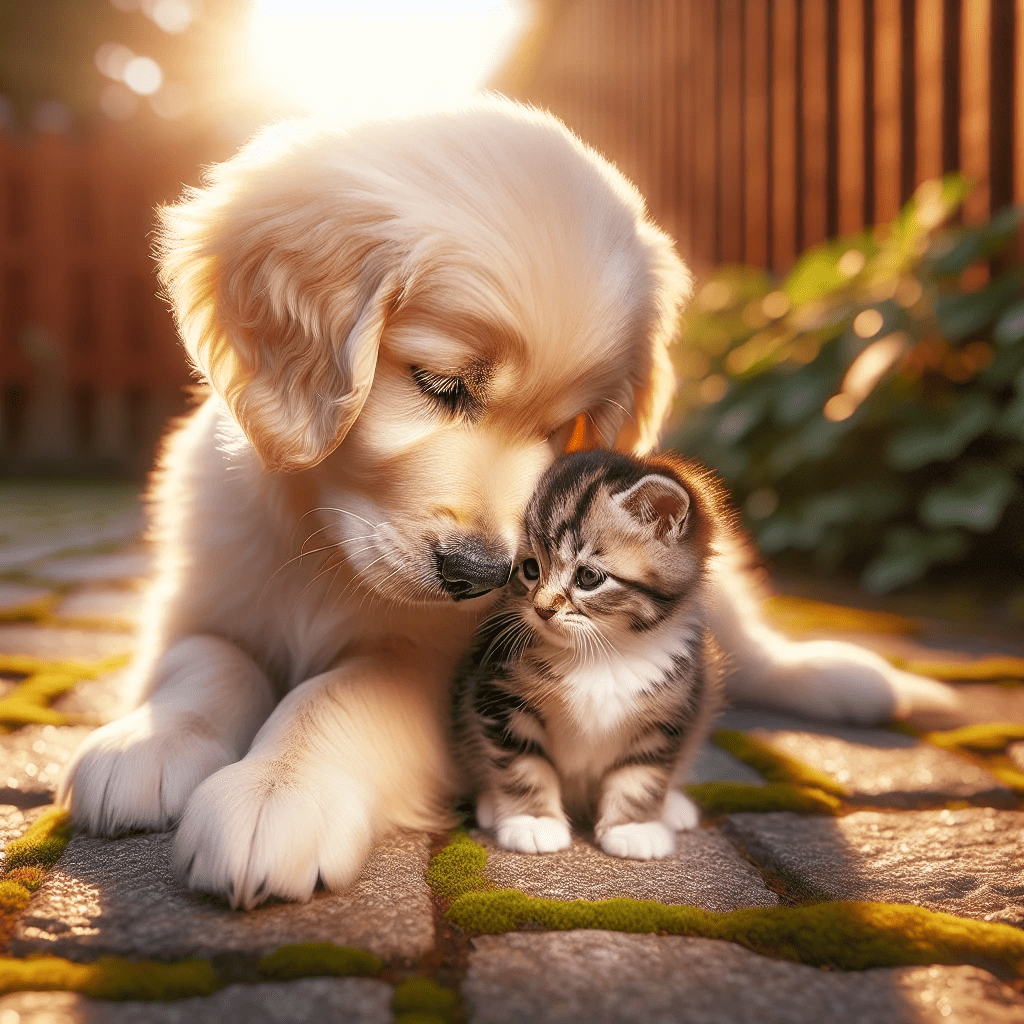} & A young golden retriever puppy with a soft, fluffy coat and gentle eyes, tenderly nuzzling a small American Shorthair kitten with a curious and attentive expression. Both animals are sitting close together on a sunlit cobblestone path with patches of vibrant green moss, with the puppy's paw affectionately resting on the kitten, in the warm ambiance of a backyard during the golden hour. The background features a soft bokeh of lush greenery and the warm tones of a wooden fence, evoking a serene garden or park. The scene captures a moment of affection and camaraderie, showcasing the endearing connection between the two different species, with the sunlight casting a gentle glow and creating soft shadows on their fur.  \\ \midrule

\includegraphics[width=30mm]{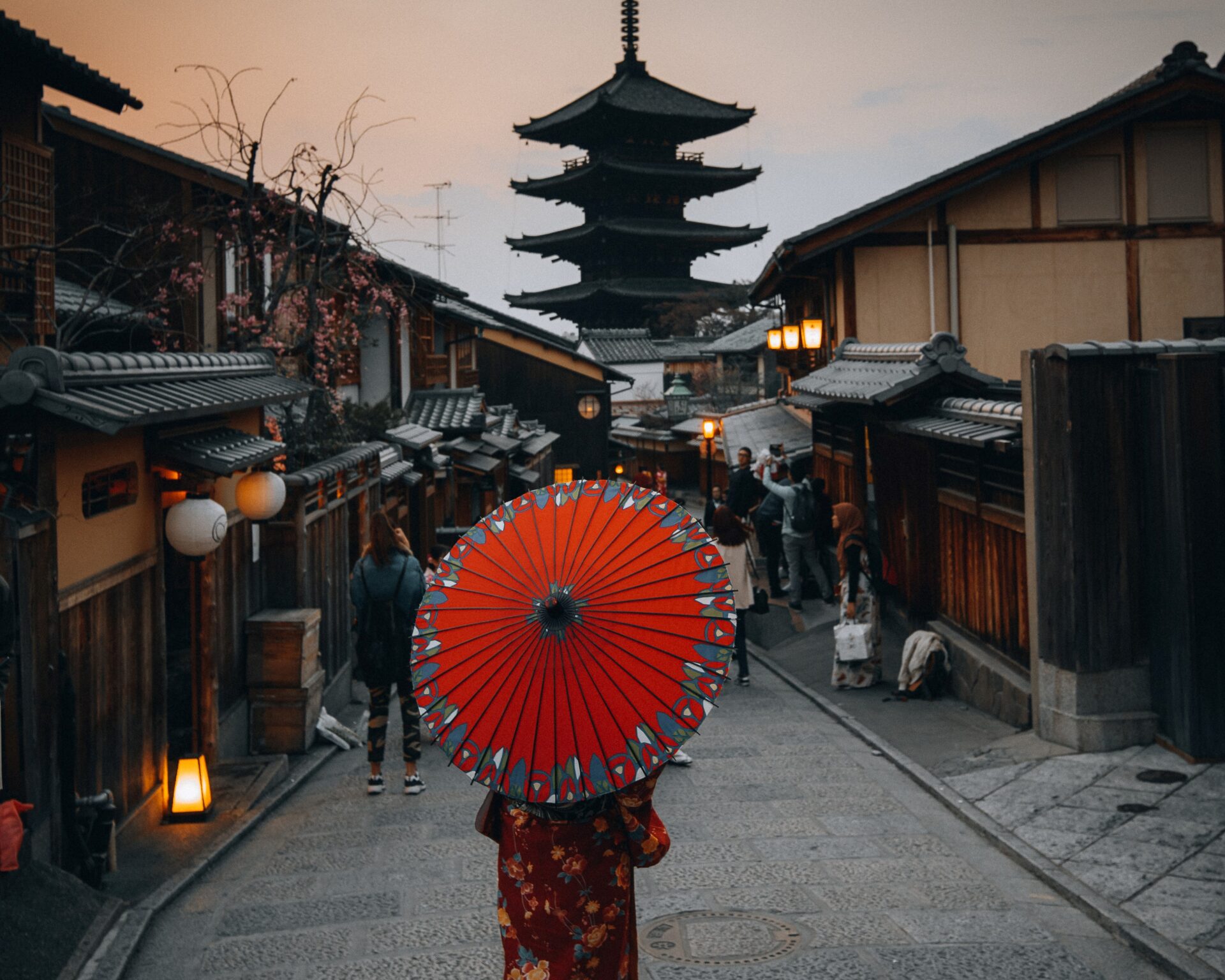}  & \includegraphics[width=30mm,height=30mm]{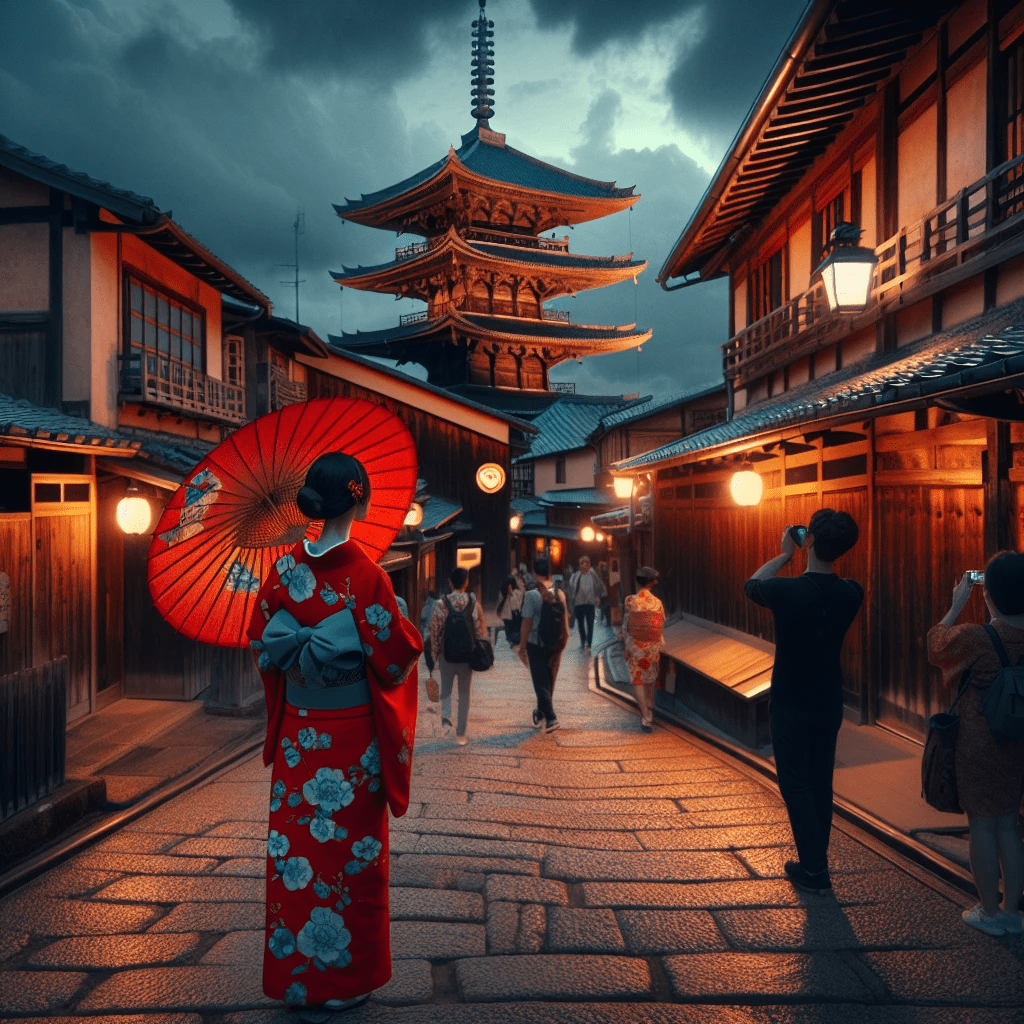} & \includegraphics[width=30mm,height=30mm]{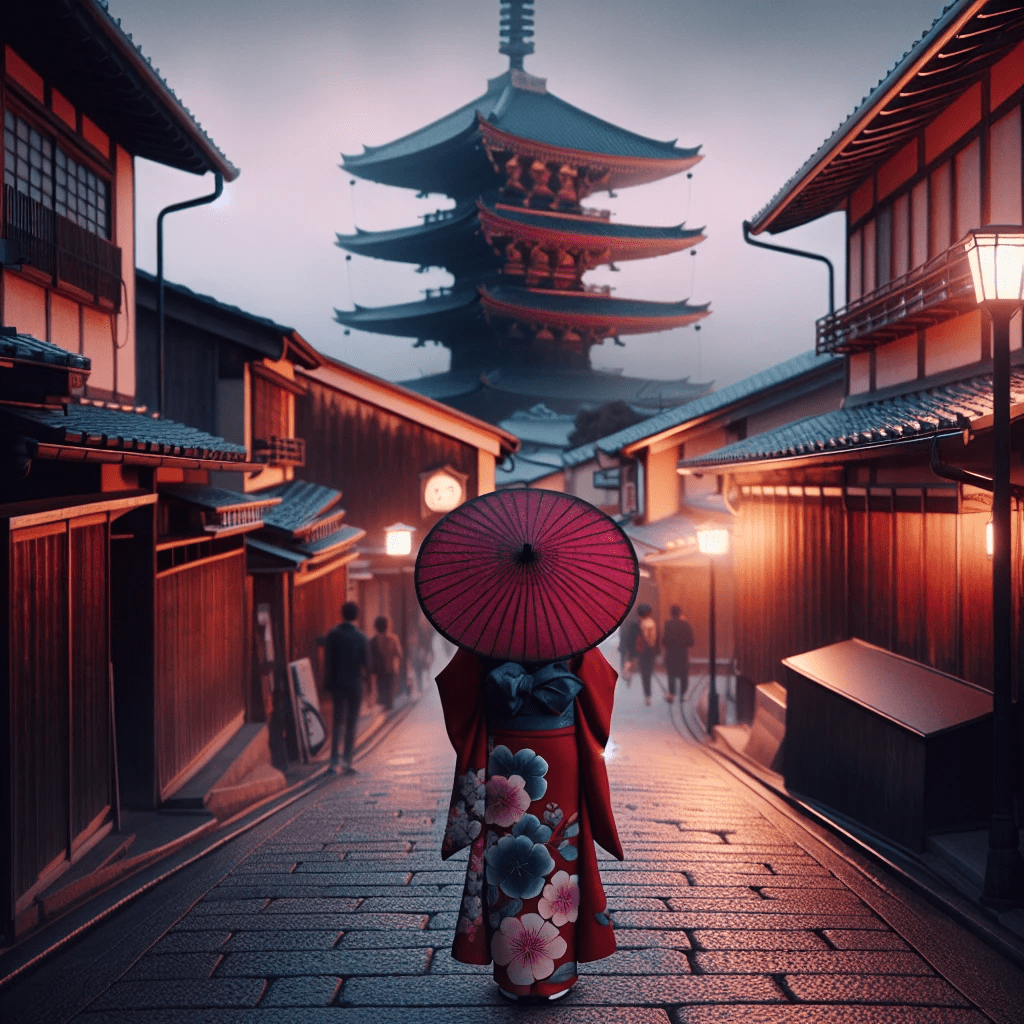} & A person in a luxurious crimson kimono, embossed with bold indigo floral patterns, stands diminutive at the lower center of a photorealistic Japanese street as dusk settles in. Facing a grand five-tiered pagoda that ascends into the hazy sky, they hold an expansive crimson paper umbrella aloft, masking their upper body and creating an arresting visual anchor. The alley, bathed in the soft glow from the traditional wooden buildings’ lanterns, stretches around them, while the cobblestone path gleams under the ambient light. In the background, life continues as silhouettes of pedestrians engage in subdued conversations or pause to photograph the scene, adding layers of depth and motion to the tranquil tableau. The pagoda, a silhouette against the misty heavens, invites the viewer's gaze upward, reinforcing the composition's sense of depth and perspective.
\\ \midrule

\includegraphics[width=30mm]{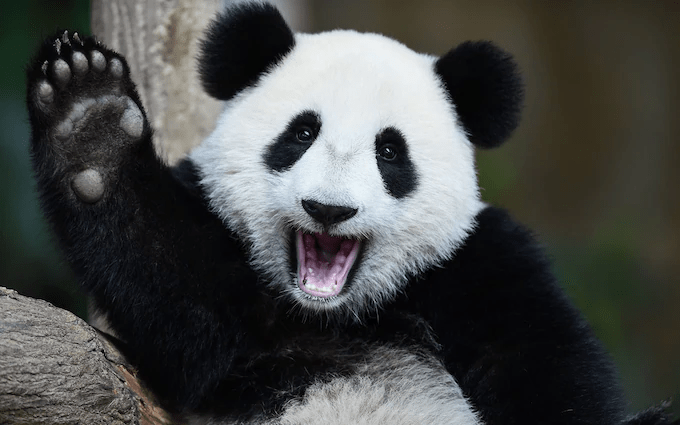}  & \includegraphics[width=30mm,height=30mm]{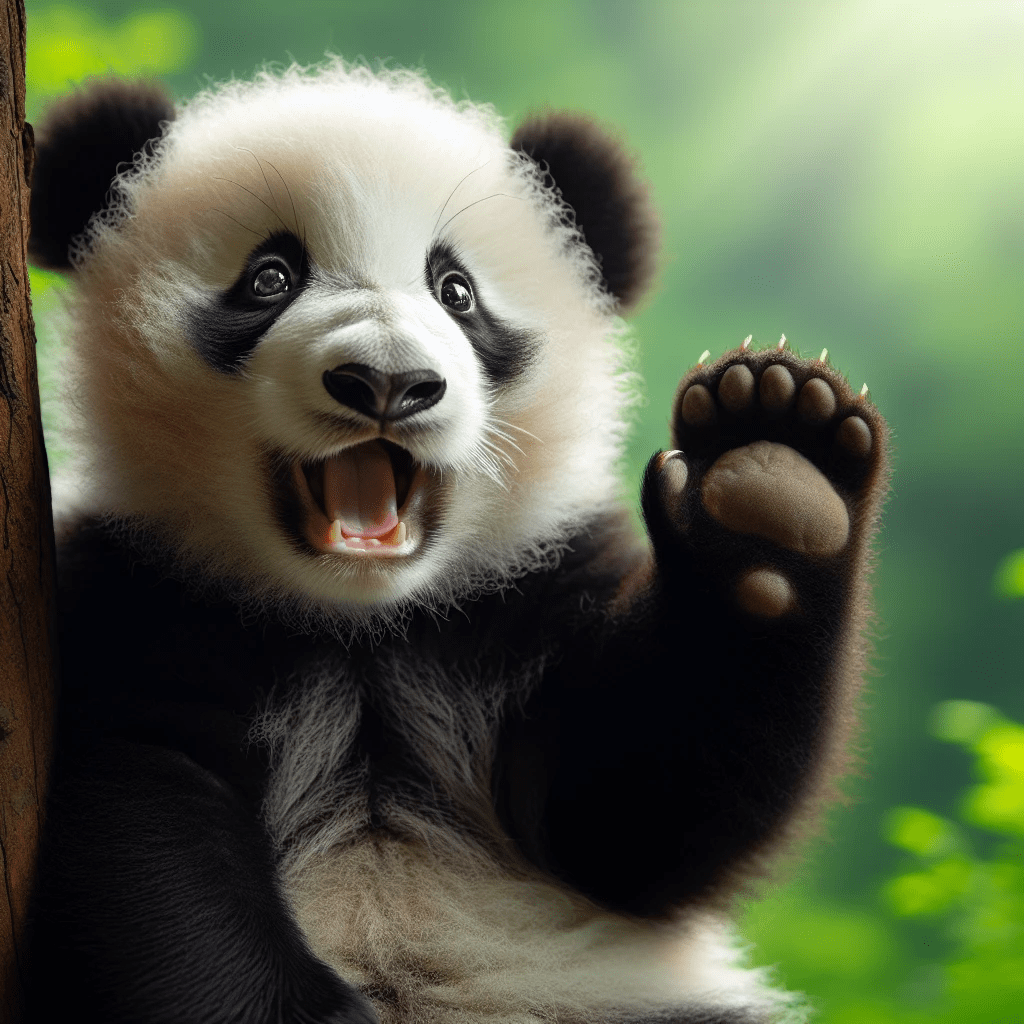} & \includegraphics[width=30mm,height=30mm]{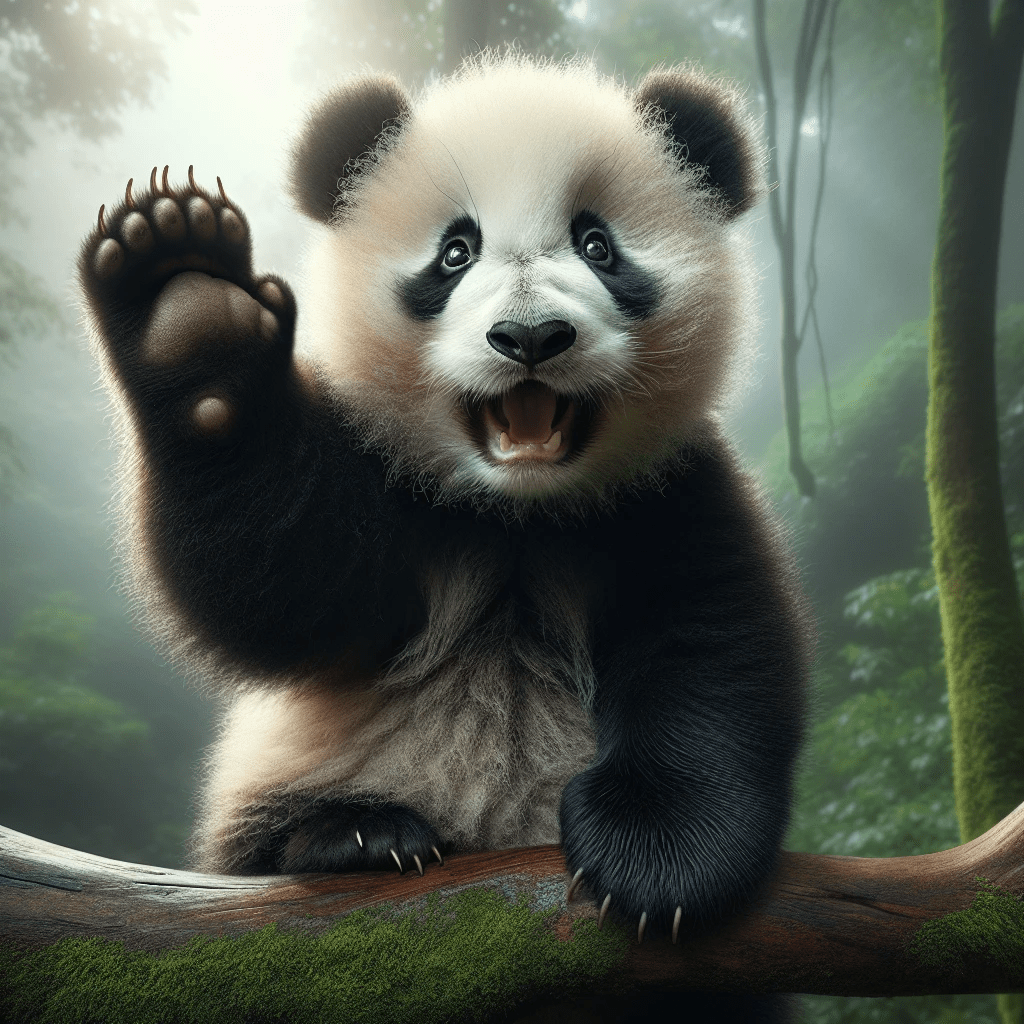} & Create an image of a juvenile giant panda with a striking black and white fur pattern, perched on a tree branch. The panda's mouth is agape as if mid-vocalization, and it is raising its left paw in a greeting gesture, showcasing its prominent claws. Its eyes are round and expressive, reflecting a sense of wonder. The background is a soft-focus portrayal of lush greenery, evoking a dense, misty forest atmosphere. The lighting is diffuse, with a subtle emphasis on the panda's facial features to highlight its endearing and playful demeanor. \\

\bottomrule[1.5pt]
\end{NiceTabular}
}
\caption{\textbf{Prompt inversion for natural images.} We show that our framework can also reverse engineer prompts for natural photos.}
\label{tab:real_img_inversion}
\end{table*}

\begin{table*}[h]
\centering
\scalebox{0.9}{
\begin{tabular}{c M{0.8\linewidth}}
\toprule[1.5pt]
\textbf{Score} & {\bf Meaning}
\\ 
\midrule
1 & {\bf Not Aligned.} The generated image shows a significant divergence from the user query. Key elements, attributes, or relationships differ notably, indicating a clear mismatch. \\ \midrule
2 & {\bf Mildly Similar.} The generated image shows a basic level of resemblance to the user query. It includes some of the requested elements or themes, but there are significant inaccuracies or omissions, making it only loosely related.\\ \midrule
3 & {\bf Moderately Similar.} The generated image is moderately aligned with the user query. Most of the key elements and attributes are present, but there may be some minor inaccuracies or missing details. \\ \midrule
4 & {\bf Highly Similar.} The generated image closely aligns with the user query, accurately representing most main objects, attributes, and their relations, with only minor discrepancies.\\ \midrule
5 & {\bf Perfect.} The generated image perfectly matches the user query in every aspect. All main objects, attributes, and their relations are exactly as requested, representing an ideal, precise match.\\
\bottomrule[1.5pt]
\end{tabular}
}
\caption{\textbf{Likert scale for human evaluation.} }
\label{tab:likert}
\end{table*}

\clearpage
\clearpage
{
    \small
    \bibliographystyle{ieeenat_fullname}
    \bibliography{main}
}

\end{document}